\documentclass[10pt,twocolumn,letterpaper]{article}

\usepackage{cvpr}      

\usepackage{graphicx}
\usepackage{amsmath}
\usepackage{amssymb}
\usepackage{multirow}
\usepackage{booktabs}

\usepackage[pagebackref,breaklinks,colorlinks]{hyperref}

\usepackage[capitalize]{cleveref}
\usepackage{algorithm}
\usepackage[noend]{algorithmic}
\crefname{section}{Sec.}{Secs.}
\Crefname{section}{Section}{Sections}
\Crefname{table}{Table}{Tables}
\crefname{table}{Tab.}{Tabs.}

\def\cvprPaperID{3314} 
\def\confName{CVPR}
\def\confYear{2023}

\begin{document}

\title{GDB: Gated convolutions-based Document Binarization}

\author{Zongyuan Yang\,\textsuperscript{\rm 1}, Yongping Xiong\,\textsuperscript{\rm 1}\,\thanks{Corresponding authors.}~, Guibin Wu\,\textsuperscript{\rm 1}\\
 \\
\textsuperscript{\rm 1}\,School of Computer Science, Beijing University of Posts and Telecommunications\\
{\tt\small
\{yangzongyuan0, ypxiong, wuguibin\}@bupt.edu.cn\
}
}
\maketitle

\begin{abstract}
Document binarization is a key pre-processing step for many document analysis tasks. However, existing methods can not extract stroke edges finely, mainly due to the fair-treatment nature of vanilla convolutions and the extraction of stroke edges without adequate supervision by boundary-related information. In this paper, we formulate text extraction as the learning of gating values and propose an end-to-end gated convolutions-based network (GDB) to solve the problem of imprecise stroke edge extraction. The gated convolutions are applied to selectively extract the features of strokes with different attention. Our proposed framework consists of two stages. Firstly, a coarse sub-network with an extra edge branch is trained to get more precise feature maps by feeding a priori mask and edge. Secondly, a refinement sub-network is cascaded to refine the output of the first stage by gated convolutions based on the sharp edge. For global information, GDB also contains a multi-scale operation to combine local and global features. We conduct comprehensive experiments on ten Document Image Binarization Contest (DIBCO) datasets from 2009 to 2019. Experimental results show that our proposed methods outperform the state-of-the-art methods in terms of all metrics on average and achieve top ranking on six benchmark datasets.
\end{abstract}

\section{Introduction}
\label{sec:intro}

Document binarization is the task of classifying each pixel in a document image into either foreground text or background. As an essential pre-processing step, its result significantly impacts the performance of many document analysis tasks, including optical character recognition (OCR) \cite{Bhunia_2021_CVPR}, document layout analysis \cite{Arroyo_2021_CVPR} and page segmentation \cite{Tran2016}, especially on heavily degraded document images such as the historical document images. These documents usually suffer from multiple degradations, including bleed-through, smear, broken holes, crease, faint ink and uneven strokes. The main challenge of document binarization is to extract fine stroke edges in such degraded document images. The different fonts, writing materials and character sizes in the documents make the extraction more challenging \cite{Kligler_2018_CVPR, Rowley-Brooke_2013_CVPR, yang2023novel}.

\begin{figure}[!htbp]
		\centering
		\begin{subfigure}{0.49\linewidth}
		      \includegraphics[width=\columnwidth]{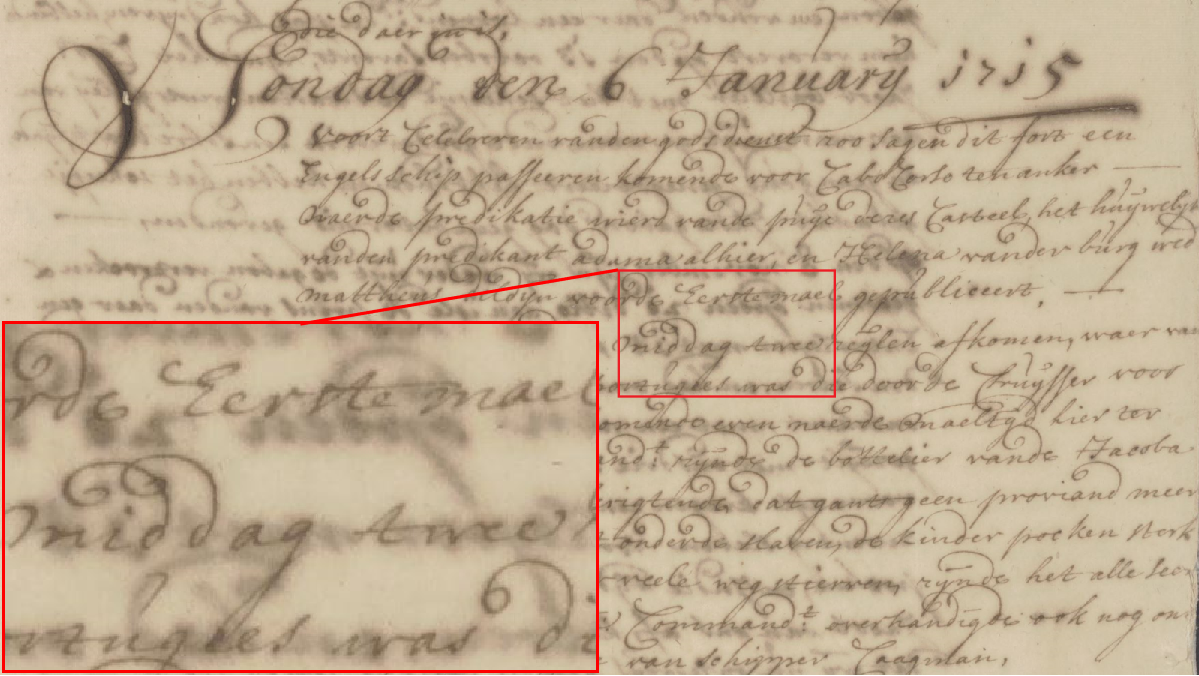}
		\caption{Original image}
		\end{subfigure}
		\begin{subfigure}{0.49\linewidth}
			\includegraphics[width=\columnwidth]{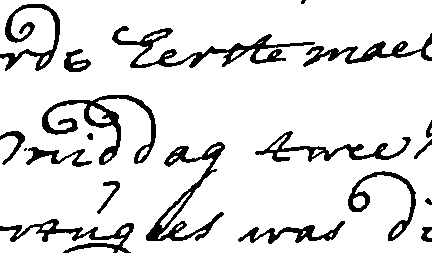}
		\caption{Ground-truth}
		\end{subfigure}
		\begin{subfigure}{0.49\linewidth}
			\includegraphics[width=\columnwidth]{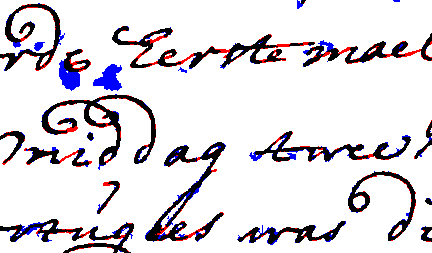}
		\caption{Jia \cite{Jia2018}}
		\end{subfigure}
		\begin{subfigure}{0.49\linewidth}
			\includegraphics[width=\columnwidth]{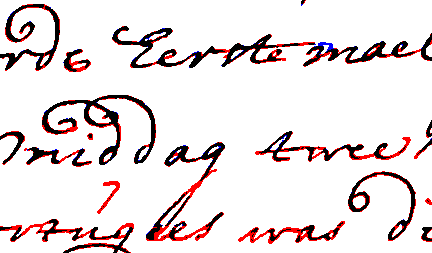}
		\caption{Suh \cite{Suh2022}}
		\end{subfigure}
		\begin{subfigure}{0.49\linewidth}
			\includegraphics[width=\columnwidth]{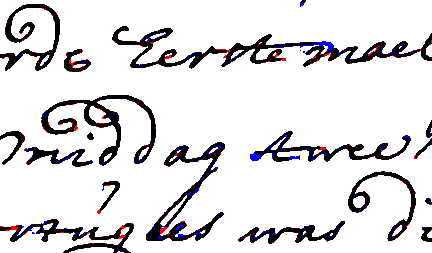}
		\caption{GDB}
		\end{subfigure}
		\begin{subfigure}{0.49\linewidth}
			\includegraphics[width=\columnwidth]{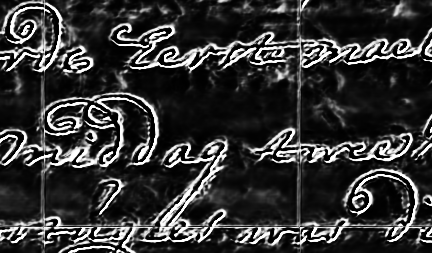}
		\caption{learned gating values}
		\label{fig:start}
		\end{subfigure}
		\caption{The effectiveness of gated convolutions. White pixels are appropriately classified background, whereas black pixels are correctly classified text. Text pixels classified as background are highlighted in \textcolor{red}{red}, whereas background pixels classified as text are highlighted in \textcolor{blue}{blue}. Same for the following Figures. Benefiting from the precise extraction of stroke edge features by the gating mechanism, GDB has sharper and more precise stroke edges. }
		\label{fig:DIBCO2013_9}
\end{figure}

Generally, the state-of-the-art methods can be divided into two categories \cite{Suh2022}. The first class of methods \cite{FarrahiMoghaddam2012, Lelore2013, Jia2018} are based on the adaptive threshold. They often utilize edge detection, background estimation, and histogram analysis as the basic steps. The second class of methods \cite{Souibgui2022, Zhao2019, Suh2022} are mainly based on fully convolutional networks. For document binarization, the stroke edge has been exploited as a powerful text indication to calculate the threshold \cite{Lelore2013, Jia2018}. Due to various types of degradation, stroke edges may not be accurately detected. In complex backgrounds, the noise with high gradient variation can easily be detected as stroke edges, whereas the edges of faint ink strokes or uneven pen strokes are often missed. In \cite{Lu2010, Ntirogiannis2014, Jia2018}, the extraction of stroke edges relies on the performance of background estimation, which leads to a large number of errors on low-quality backgrounds, such as the DIBCO'19 dataset \cite{Pratikakis2019}. 
The existing FCN-based methods \cite{Souibgui2022, Zhao2019, Suh2022, He2021} utilize vanilla convolutions for binarization, where the stroke edges are treated equally with the surrounding features within their receptive fields. This deprives the network of specific attention to stroke edges. Furthermore, all existing deep learning-based methods do not sufficiently supervise the extraction of stroke edges through the boundary-related information of the strokes.

Considering these problems, inspired by the characteristics of gated convolutions, we transform the extraction of stroke edges as the learning of gating values. We apply gated convolutions to selectively extract the features of strokes with different attention. Gated convolutions \cite{takikawa2019, Li2020, Wang2021b} have achieved promising performance in semantic segmentation.
During the propagation of features, gated convolutions distinguish between valid and invalid information by the learned gating values. Gated convolutions lead to a “boundary-sensitive” network by feeding the priori mask and edge map. As shown in \cref{fig:DIBCO2013_9}, the learned gating values (\cref{fig:start}) focus on the stroke edges, which allows the stroke edge features to be extracted and supervised efficiently during training. The gatings filter out noisy features except the stroke edges. Thus, GDB extracts stroke edges more accurately at the pixel level compared to the state-of-the-art methods \cite{Jia2018, Souibgui2022}.
 
In this paper, we propose an end-to-end gated convolutions-based network (GDB) for document binarization, which is designed with a coarse-to-refine generator and two discriminators. Firstly, we train a coarse sub-network with an extra edge branch to get more precise feature maps by feeding the priori mask and edge. To overcome the limitation of the receptive field, a multi-scale operation combining local and global features is processed in parallel. Secondly, we cascade a refinement sub-network to integrate the benefits of the coarse sub-network outputs at multiple scales by gated convolutions.
 
To demonstrate the effectiveness of our proposed methods, we have conducted comprehensive experiments on ten Document Image Binarization Contest (DIBCO) datasets from 2009 to 2019\cite{Gatos2009, Pratikakis2010, Pratikakis2011, Pratikakis2012, Pratikakis2013, Ntirogiannis2014a, Pratikakis2016, Pratikakis2017, Pratikakis2018, Pratikakis2019} based on four metrics, F-Measure (FM), pseudo-FMeasure (p-FM) \cite{Ntirogiannis2013}, Peak Signal to Noise Ratio (PSNR), and Distance Reciprocal Distortion Metric (DRD) \cite{Lu2004}. Experimental results show that our proposed methods outperform the state-of-the-art methods in terms of all metrics on average and achieve top ranking on six benchmark datasets.
In summary, the contributions of our paper are as follows:
\begin{itemize}
    \setlength{\itemsep}{0pt}
    \setlength{\parsep}{0pt}
    \setlength{\parskip}{0pt}
  \item We transform the extraction of text in degraded document images into the learning of gating values in gated convolutions. We demonstrate that learning of gating values can be weakly guided by feeding a priori mask and edge map, which improves the performance of stroke extraction.
  \item We propose a multi-branch gated convolutional generative adversarial network (GDB) for document binarization. By exploring the gated convolutions, our model generates refined stroke edges by adjusting mask and edge maps. An additional edge branch is proposed to focus on processing the edge information. The proposed method achieves better performance on average over the all (H-) DIBCO datasets compared to the state-of-the-art methods.
  \item We show that the learned gating values gate the propagation of different stroke features, contirbuting to the fine extraction of stroke edges and the suppression of background noises.
\end{itemize}

\section{Related work}\label{sec:relate}

\subsection{Gated convolutions}

Gated convolutions have been widely explored in semantic segmentation \cite{takikawa2019, Li2020, Wang2021b}, image generation \cite{van2016}, image inpainting \cite{Yu2019} and many other tasks \cite{Srivastava2015, dauphin2017, Cao2019}. For example, Takikawa \etal \cite{takikawa2019} propose a novel two-stream architecture with a new type of gate for semantic segmentation that use the higher-level features in the classical stream to gate the lower-level features in the shape stream. In \cite{Li2020}, Gated Full Fusion (GFF) is proposed for semantic segmentation that uses gating to selectively fuse features from different layers to make sufficient use of valid information. In \cite{Wang2021b}, Wang \etal propose a novel module Gated ScaleTransfer Operation (GSTO) that enhances the performance of extracting scale-aware multi-scale features in semantic segmentation. To solve the inability of vanilla convolution to differentiate between valid and invalid pixels, Yu \etal \cite{Yu2019} utilize gated convolutions for image inpainting with a free-form mask. Document binarization aims to use two shades of grey to differentiate between text and background, which resembles the goal of semantic segmentation. In our case, we utilize gated convolutions with noisy mask and edge as input to refine the extraction of stroke edges and to remove various types of noise from the background.

\subsection{Document binarization}

Generally, document binarization methods can be divided into two types \cite{He2021, Suh2022}: traditional threshold-based algorithms and deep-learning-based methods.

\begin{figure*}[!htbp]
		\centering
		\includegraphics[width=0.8\linewidth]{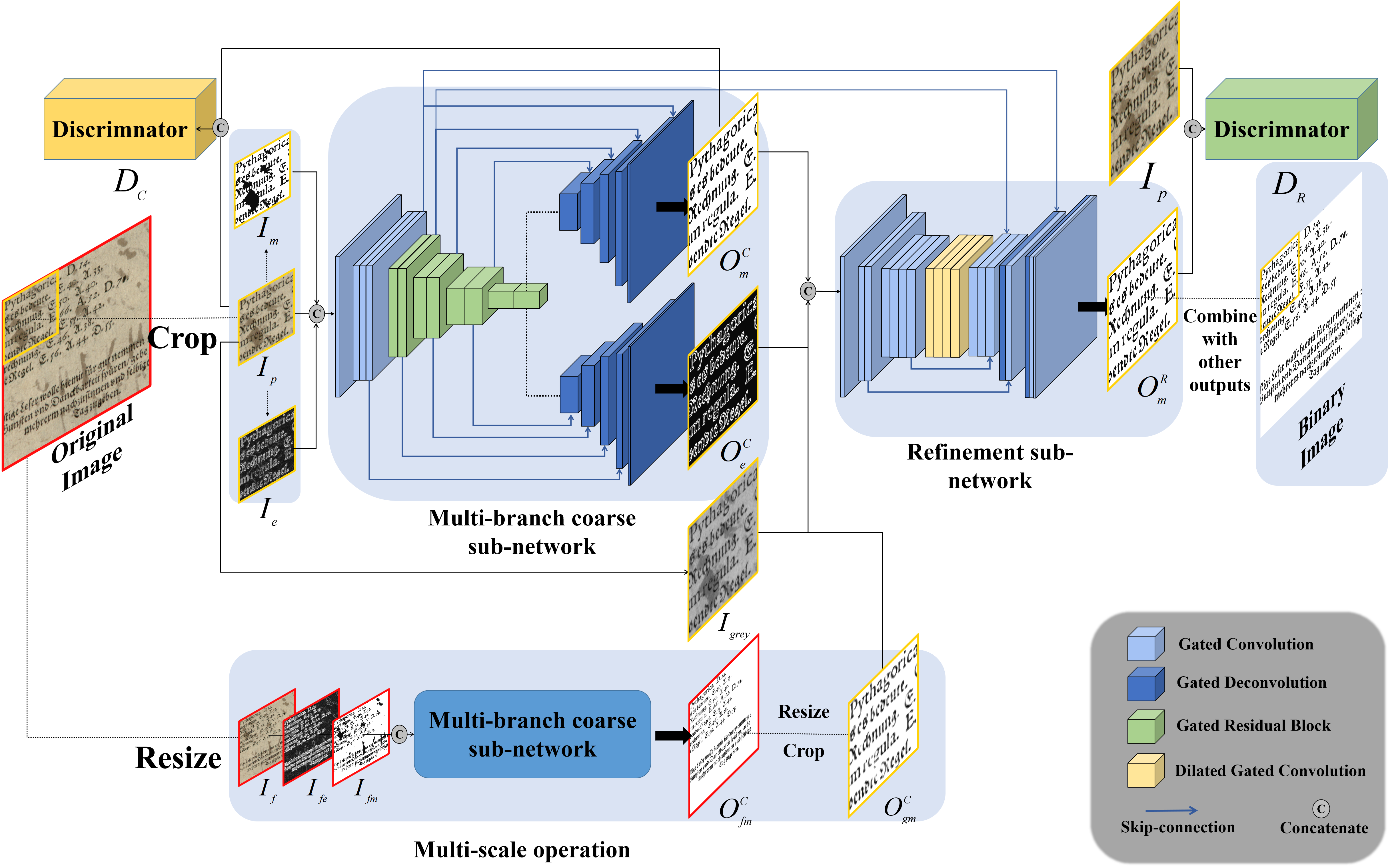}
		\caption{The architecture of the proposed GDB for document image binarization.}
		\label{fig:fig_overview}
\end{figure*}

\textbf{Traditional threshold-based algorithms} Otsu's method \cite{Otsu1979} computes the global threshold on the whole document images, while the local threshold methods, including Niblack \cite{niblack1985introduction} and Sauvola \cite{Sauvola2000}, compute the pixel-wise threshold based on statistical information of a local patch. In \cite{Lu2010}, the local threshold is calculated based on the detected text stroke edges by background estimation. AdOtsu \cite{FarrahiMoghaddam2012} introduces an adaptive form of Otsu's method to differentiate between text and background by the estimated background map. Howe \cite{Howe2013} introduces an automatic technique for tuning the parameters of the binarization algorithms. Jia \etal \cite{Jia2016, Jia2018} utilize the structural symmetric pixels on the background compensation image to compute the local threshold in the neighborhood. These traditional methods require certain empirical parameters, which limits their generalization and performance on low-quality document images.

\textbf{Deep-learning-based methods} Full Convolutional Networks (FCNs) have been widely explored in document binarization. Tensmeyer and Martinez \cite{Tensmeyer2017} apply a novel FCN architecture which combines multiple scales for document binarization. DeepOtsu \cite{He2019} iteratively removes background noise and extracts strokes with recurrent refinement and stack refinement. Zhao \etal \cite{Zhao2019} propose a two-stage cascaded conditional generative adversarial network based on Pix2Pix \cite{Isola2018}. He and Schomaker \cite{He2021} propose a cascaded T-shaped network termed CT-Net to learn both document enhancement and binarization tasks. In \cite{KhamekhemJemni2022}, a handwritten text recognizer is integrated with the architecture of cGANs in order to exploit text semantic information. In \cite{Suh2022}, Suh \etal proposes a two-stage colour-independent generative adversarial network based on EfficientNet \cite{tan2019efficientnet} for document image enhancement.

\section{Method}\label{sec:proposed}

The overall architecture of GDB is shown in \cref{fig:fig_overview}. GDB is designed with a coarse-to-refine generator and two discriminators. The generator contains a multi-branch coarse sub-network $G_{C}$ and a refinement sub-network $G_R$. Firstly, $G_C$ takes as input three images ($I_{p}$, $I_{m}$, $I_{e}$) where $I_{p}$ is the patch cropped from the original image, $I_{m}$ is the noisy mask generated by Otsu's method \cite{Otsu1979} and $I_{e}$ is the edge gradient map generated by Sobel edge detector \cite{sobel19683x3}. The two upsampling branches predict the mask $O_{m}^{C}$ and edge $O_{e}^{C}$, respectively. Additionally, a multi-scale operation that combines local and global features predicts the global mask $O_{gm}^{C}$ in parallel. Secondly, $G_R$ takes as input four images ($I_{grey}$, $O_{m}^{C}$, $O_{e}^{C}, O_{gm}^{C}$), where $I_{grey}$ is the grey scale of $I_{p}$, and then predicts the finer binarization results with sharp edges. The generated masks from the two stages are then fed into their corresponding discriminators and discriminated from the real ones.

\subsection{Gated Convolutional Layer}

Compared to vanilla convolutions, gated convolutions multiply a weight at different spatial locations of the feature maps in different channels based on the input mask (grey-scale map), which can distinguish foreground pixels from background pixels. This mechanism can dynamically update the mask during the training process and guide the effect of image generation with an extra edge channel (grey-scale map). Gated convolutions can be formulated as, 
\begin{gather}
  \alpha_{Gating}^{x,y} =W_{g} \otimes I  \notag \\
  O_{Features}^{x,y} =W_{f} \otimes I \\
  O^{x,y} = \sigma(\alpha_{Gating}^{x,y})\odot \phi(O_{Features}^{x,y}) \notag
\end{gather}
where $W_{g}$, $W_{f}$ are two different learnable convolution filters, $\sigma$ is the sigmoid activation function, $\phi$ can be any non-linear activation functions, such as ReLU, ELU and LeakyReLU. GDB is a fully gated convolutional network.

For degraded document images, their masks and edges can be easily obtained from traditional algorithms. Our goal is to utilize noisy masks and edges to weakly guide the feature extraction of coarse sub-network by gating values. In the second stage, the stroke edges can be finer extracted, and the background noise can be better suppressed by the input of cleaner masks and edges generated from stage one.

\begin{figure*}[!htbp]
		\centering
		\includegraphics[width=\linewidth]{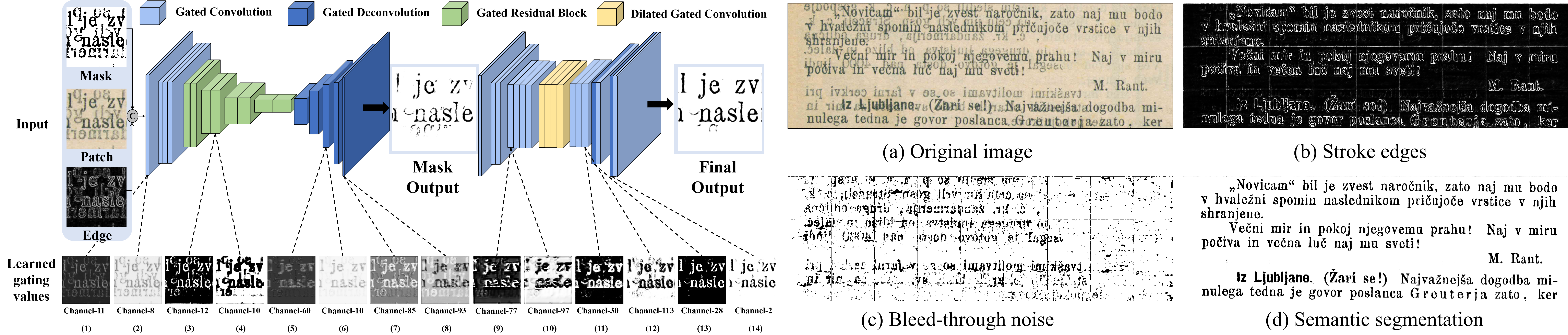}
		\caption{The visualization of learned gating values. (b), (c) and (d) are the learned gating values. The figure does not show the multi-scale operation, the edge branch, the connections between layers and the discriminators for simplicity.}
		\label{fig:lgv}
\end{figure*}

Each gating value is between zero and one, representing the degree of attention paid to the pixel values at different spatial locations in the feature map. \cref{fig:lgv} provides a visualization of learned gating values. Each saliency map representing the gating values can emphasize different areas of the feature map where the region of interest has higher gating values. For example, saliency map 7 concerns the text area, while saliency map 8 from the same layer focuses on the bleed-through noise. Ideally, each saliency map consists of only zeros and ones to distinguish between two distinct areas, but there is ambiguity in the learned gating values. For example, the bleed-through noise and the text have relatively high gating values in the saliency map 3. However, provided the background or text gating values are closer to one than the noise, the noise can be effectively suppressed, even if the learned gating values are ambiguous. This is because only features with gating values closer to one are retained when passing through the deep networks. As shown in \cref{fig:lgv}, the bleed-through noise is less significant than the text or the background in each saliency map. Therefore, the deeper the layer from which the learned gating values come, the less noise there is. 

We convert the learned gated values in the penultimate layer of the network into a grayscale map. As shown in \cref{fig:lgv}, the learned gating values gate different features. Gating (b) focuses on extracting stroke edges, which gates the propagation of stroke edge features. Gating (c) filters out the bleed-through noise by setting its gating values close to 0. Gating (d) can be interpreted as the semantic segmentations of text and background. It is worth noting that the gating values have not been supervised during training except for the priori information in the input (mask and edge).

\subsection{Multi-branch coarse sub-network}

Our multi-branch coarse sub-network is designed like an encoder-decoder FCN. For encoding, the input ($I_{p}$, $I_{m}$, $I_{e}$) is encoded by four gated convolutional layers and several gated residual blocks modified from residual blocks \cite{He2016}. For decoding, two branches share the same architecture with five up-sampling gated deconvolutional layers. The encoder and decoder are cascaded with four skip connections to improve the connectivity. The outputs $O_{m}^{C}$ and $O_{e}^{C}$ can be seen as an enhanced version of $I_{m}$ and $I_{e}$.

Local prediction with small patches has many limitations. When there is a large area of continuous noise in the document image, background pixels are sometimes misclassified as text pixels due to the limitation of the receptive field. It has been demonstrated that combining local and global features can improve the effect of noise suppression \cite{Vo2018, Zhao2019, He2021, Suh2022}. Inspired by this, we resize the full document image to 256 $\times$ 256. The mask $I_{fm}$ and the edge $I_{fe}$ are obtained in the same way through the resized full image $I_{f}$. As shown at the bottom of \cref{fig:fig_overview}, the multi-branch coarse sub-network similarly processes the input ($I_{f}$, $I_{fm}$, $I_{fe}$) in parallel. Next, the predicted mask $O_{fm}^{C}$ is resized to the same size as the original image and then cropped according to the given (x,y) coordinates to get the global mask patch $O_{gm}^{C}$ that includes the global information.

\subsection{Refinement sub-network}

Owing to the lack of multi-scale fusion and sufficient supervision, the coarse sub-network confronts the issue of the background noise not being eliminated, resulting in the predicted image including the remains of crude strokes or some piece of the noise in the input mask $I_{m}$ and edge $I_{e}$. Thus, we cascade a refinement sub-network to solve this problem. The task of our refinement network is to learn to integrate the benefits of stage 1 outputs at multiple scales. We utilize dilated convolutions to enlarge the receptive fields. 

The coarse sub-network and the refinement sub-network have similar inputs, except that the former contains an RGB patch while the latter is grey-scale. In both stages, the mask and edge are used as additional information to enhance the binarization performance. The closer the input mask and edge are to the real ones, the more precisely the network can predict the binarization results. As the refinement sub-network takes mask and edge cleaned by the coarse sub-network as input, its encoder can be less affected by noise. As a result, it can eliminate the noise and restore the strokes effectively.

\subsection{Discriminator}

Two discriminators $D_{C}$ and $D_{R}$ corresponding to the two stages share the same architecture, similar to SN-PatchGAN \cite{Yu2019}. 
Owing to the training instability of GANs, SN-GAN \cite{Miyato2018} applies the spectral normalization to the training process of GANs, so that the discriminator satisfies the Lipschitz constraint, limiting the drastic changes of its parameters. PatchGAN  \cite{Isola2018} only penalizes structure at the scale of patches. SN-PatchGAN combines these two technologies in one. 
We stack six convolutions to capture the features of Markovian patches \cite{Isola2018}. The discriminator input is the 4-channel concatenation of a degraded document image and the paired binary image.

\subsection{Loss functions}

Given the original image, the patch $I_{p}$, the mask $I_{m}$ and edge $I_{e}$, the coarse sub-network and the refinement sub-network respectively predict $O_{m}^{C}$, $O_{e}^{C}$, $O_{fm}^{C}$ and $O_{m}^{R}$, where $O_{m}^{R}$ is the final binarization result. The ground truth ($T_{f}$, $T_{f_{r}}$, $T_{m}$, $T_{e}$) where $T_{f}$ is the ground truth of the full binarization map (1 for text and 0 for background), $T_{f_{r}}$ is resized to 256*256 of $T_{f}$, $T_{m}$ is the patch cropped from $T_{f}$ and $T_{e}$ is the result of Sobel edge detection of $T_{m}$. In the multi-scale operation, the mask $O_{\hat{fm}}^{C}$ with the original size is also penalized. We train our GDB on the combination of dice loss, binary cross-entropy loss, $\mathcal{L}_{1}$ loss and adversarial loss. For the simplicity of the formula, the coarse sub-network inputs $I^{C}, I_{f}^{C}$, the outputs $O$ and the ground truth $T$ are set respectively to \{$I_{p}, I_{m}, I_{e}$\}, \{$I_{f}, I_{fm}, I_{fe}$\}, \{$O_{m}^{C}, O_{e}^{C}, O_{\hat{fm}}^{C}, O_{m}^{R}$\} and \{$T_{m}, T{e}, T_{f}, T_{m}$\}.

Due to the imbalance of training samples, we utilize dice loss to handle cases where texts only take up a small portion of the patch. We use the binary cross-entropy loss jointly for the training stability.

The dice loss  and the binary cross-entropy loss are defined as follows:
\begin{equation}
\mathcal{L}_{dice} = \sum_{i}^{4} \lambda_{i} \bigg [ 1- \frac{2 \sum_{j}^{N} O_{i}^{j} T_{i}^{j}}{\sum_{j}^{N} (O_{i}^{j})^{2}+\sum_{j}^{N} (T_{i}^{j})^{2}} \bigg ]
\end{equation}
\begin{equation}
\begin{aligned}
\mathcal{L}_{bce} & = - \frac{1}{N} \sum_{i}^{4} \lambda_{i} \bigg [ \sum_{j}^{N} T_{i}^{j} \log \left(O_{i}^{j}\right) \\ 
&+ \left(1 - T_{i}^{j}\right) \log \left(1 - O_{i}^{j}\right) \bigg ]
\end{aligned}
\end{equation}

where $N$ is the number of pixels. \{$\lambda_{i}$\} are set respectively to \{1,1,1,2\}. When the patch is pure background, we flip the text and background pixel values to prevent dice loss without passing the gradient back.

We further use $\mathcal{L}_{1}$ loss to reduce the generation ambiguity and force the pixel level consistency, which is defined as follows,
\begin{equation}\label{loss:L1}
    \mathcal{L}_{1} = \sum_{i}^{4} \lambda_{i} \| O_{i} - T_{i} \|_{1}
\end{equation}

For adversarial loss, We utilize the same hinge loss \cite{Yu2019} to penalize Markovian patches, which is defined as follows.
\begin{equation}\label{eq:g}
\mathcal{L}_{G}(z; \theta_{G})=-\mathbb{E}_{z \sim \mathbb{P}_{z}(z)}\left[D(G(z))\right]
\end{equation}
\begin{equation}\label{eq:d}
\begin{aligned}
\mathcal{L}_{D}(z, x; \theta_{D}) &= \mathbb{E}_{x \sim \mathbb{P}_{\text {data }}(x)}[\operatorname{ReLU}(1-D(x))] \\
&+\mathbb{E}_{z \sim \mathbb{P}_{z(z)}}[\operatorname{ReLU}(1+D(G(z)))]
\end{aligned}
\end{equation}
where $G$ is the generator, $D$ is the discriminator, z is the input of $G$ and x is the real image.
\begin{align}
    \label{loss:GC} \mathcal{L}_{G_{C}} &= \mathcal{L}_{G}(I^{C}; \theta_{G_{C}}) + \mathcal{L}_{G}(I_{f}^{C}; \theta_{G_{C}}) \\
    \label{loss:GR} \mathcal{L}_{G_{R}} &= \mathcal{L}_{G}((I^{C},I_{f}^{C}); (\theta_{G_{C}}, \theta_{G_{R}})) \\
    \mathcal{L}_{D_{C}} & = \mathcal{L}_{D}(I^{C}, T_{m};\theta_{D_{C}})  \notag \\
    \label{loss:DC} &+ \mathcal{L}_{D}(I_{f}^{C}, T_{fm};\theta_{D_{C}}) \\
    \label{loss:DR} \mathcal{L}_{D_{R}} &= \mathcal{L}_{D}((I^{C},I_{f}^{C}), T_{m};\theta_{D_{R}}) \\
    \label{loss:adv} \mathcal{L}_{adv}  &= \mathcal{L}_{G_{C}} + 2\mathcal{L}_{G_{R}}
\end{align}
where $G_{C}$ is the multi-branch coarse sub-network, $G_{R}$ is the refinement sub-network, $D_{C}$ and $D_{R}$ are the corresponding discriminators. The total loss of the generator can be written as,
\begin{equation}\label{loss:total}
    \mathcal{L}_{total} = \lambda_{d}\mathcal{L}_{dice} + \lambda_{b}\mathcal{L}_{bce} + \lambda_{\mathcal{L}_{1}}\mathcal{L}_{1} + \lambda_{a}\mathcal{L}_{adv}
\end{equation}
where $\lambda_{d}$, $\lambda_{b}$, $\lambda_{\mathcal{L}_{1}}$ and $\lambda_{a}$ are the weights to balance the dice, binary cross-entropy, $L1$ and adversarial loss, which are set to 1, 1, 10 and 0.1, respectively. Different from Zhao \etal \cite{Zhao2019} and Suh \etal \cite{Suh2022}, which require training two generators separately, our training process is end-to-end.

\section{Experiment}\label{sec:ex}

\subsection{Datasets and evaluation metrics}\label{metrics}

We evaluated our proposed method on ten public datasets from the document image binarization competitions: DIBCO'09 \cite{Gatos2009}, H-DIBCO'10 \cite{Pratikakis2010}, DIBCO'11 \cite{Pratikakis2011}, H-DIBCO'12 \cite{Pratikakis2012},DIBCO'13 \cite{Pratikakis2013}, H-DIBCO'14 \cite{Ntirogiannis2014a}, H-DIBCO'16 \cite{Pratikakis2016}, DIBCO'17 \cite{Pratikakis2017}, H-DIBCO'18 \cite{Pratikakis2018}, DIBCO'19 \cite{Pratikakis2019}. The training set also includes document images from 
the Bickley Diary dataset \cite{Deng2010}, Persian Heritage Image Binarization Dataset (PHIDB) \cite{Nafchi2013}, the Synchromedia Multispectral dataset (S-MS) \cite{Hedjam2015}. Inspired by \cite{Zhao2019, He2021}, all other datasets are utilized for training when evaluating a specific (H-)DIBCO dataset. 

We adopt four widely used evaluation metrics \cite{Zhao2019,Jia2018,He2021,Suh2022,Pratikakis2019} including: (1) F-Measure (FM); (2) pseudo-FMeasure (p-FM) \cite{Ntirogiannis2013}; (3) Peak Signal to Noise Ratio (PSNR) that measures similarity between two images; (4) Distance Reciprocal Distortion Metric (DRD) \cite{Lu2004} for evaluation of the visual distortion in binary images. We computed these metric scores fairly using the official evaluation tool of the competition.

\subsection{Ablation study}\label{sec:input}

\begin{figure}[!htbp]
		\centering
		\begin{subfigure}{0.23\columnwidth}
			\begin{minipage}[t]{\columnwidth}
		      \centering
              \includegraphics[width=\columnwidth]{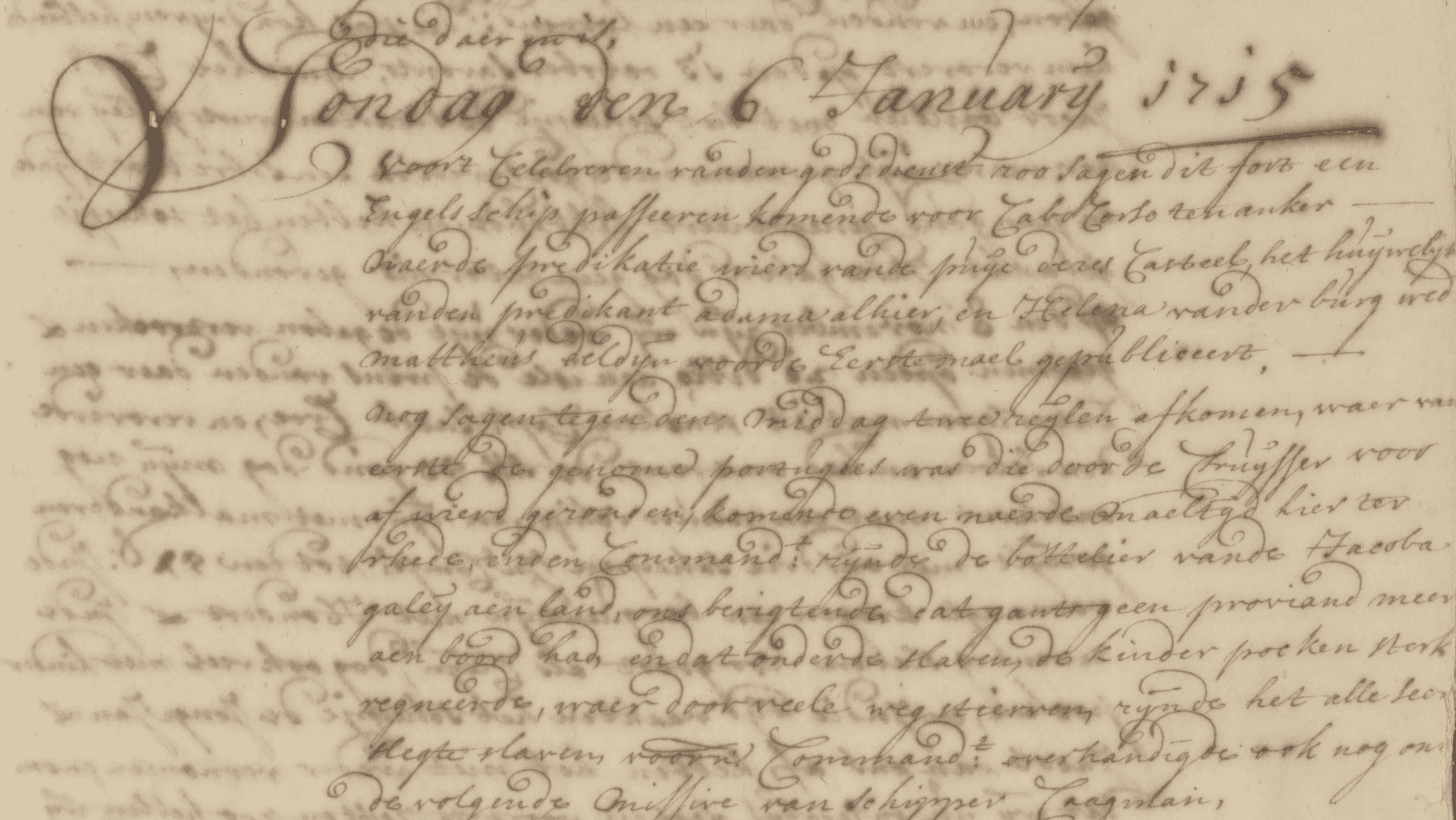}
		      \includegraphics[width=\columnwidth]{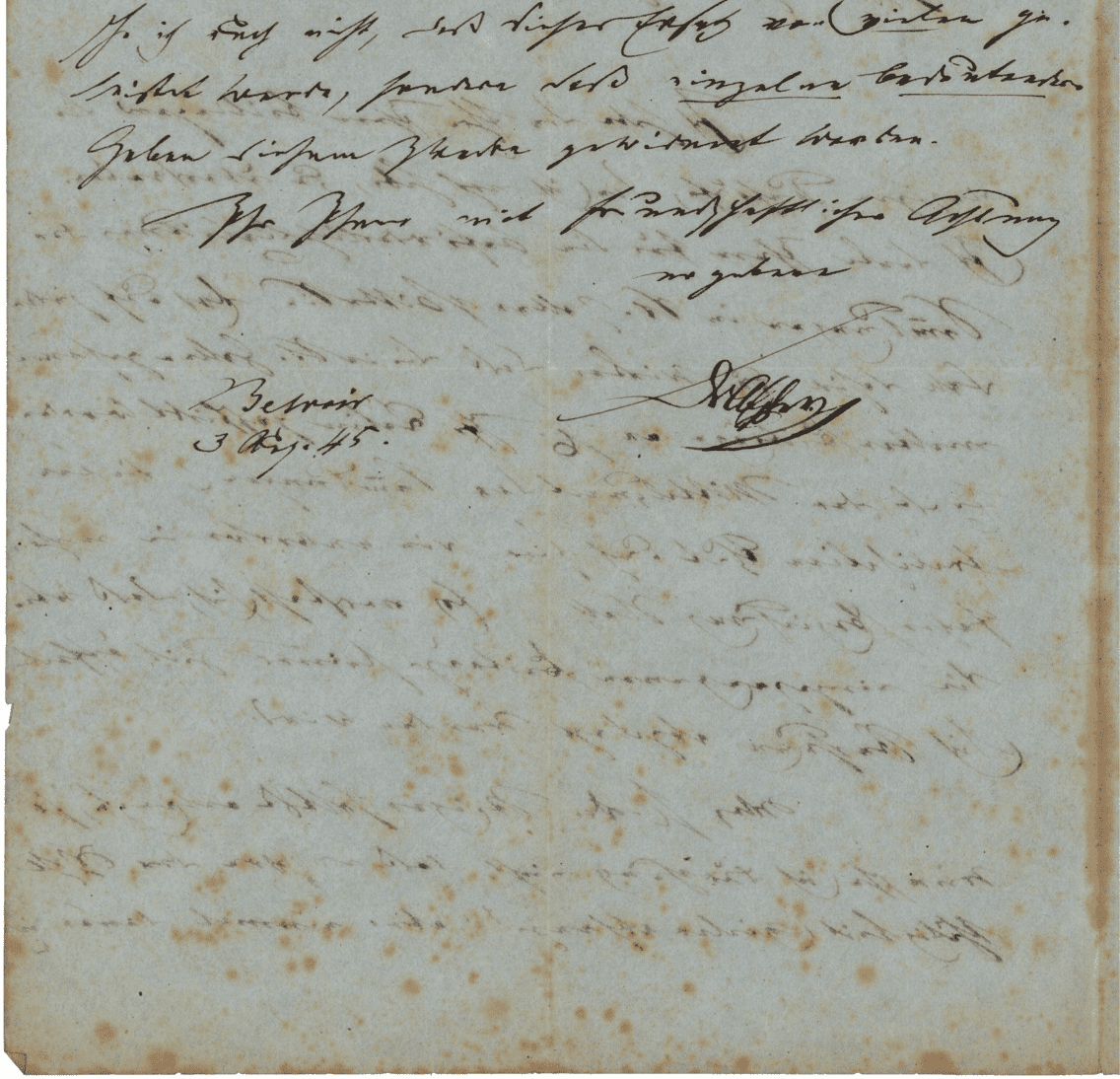}
            \end{minipage}
            \caption{}
		\end{subfigure}
		\begin{subfigure}{0.23\columnwidth}
			\begin{minipage}[t]{\columnwidth}
		      \centering
              \includegraphics[width=\columnwidth]{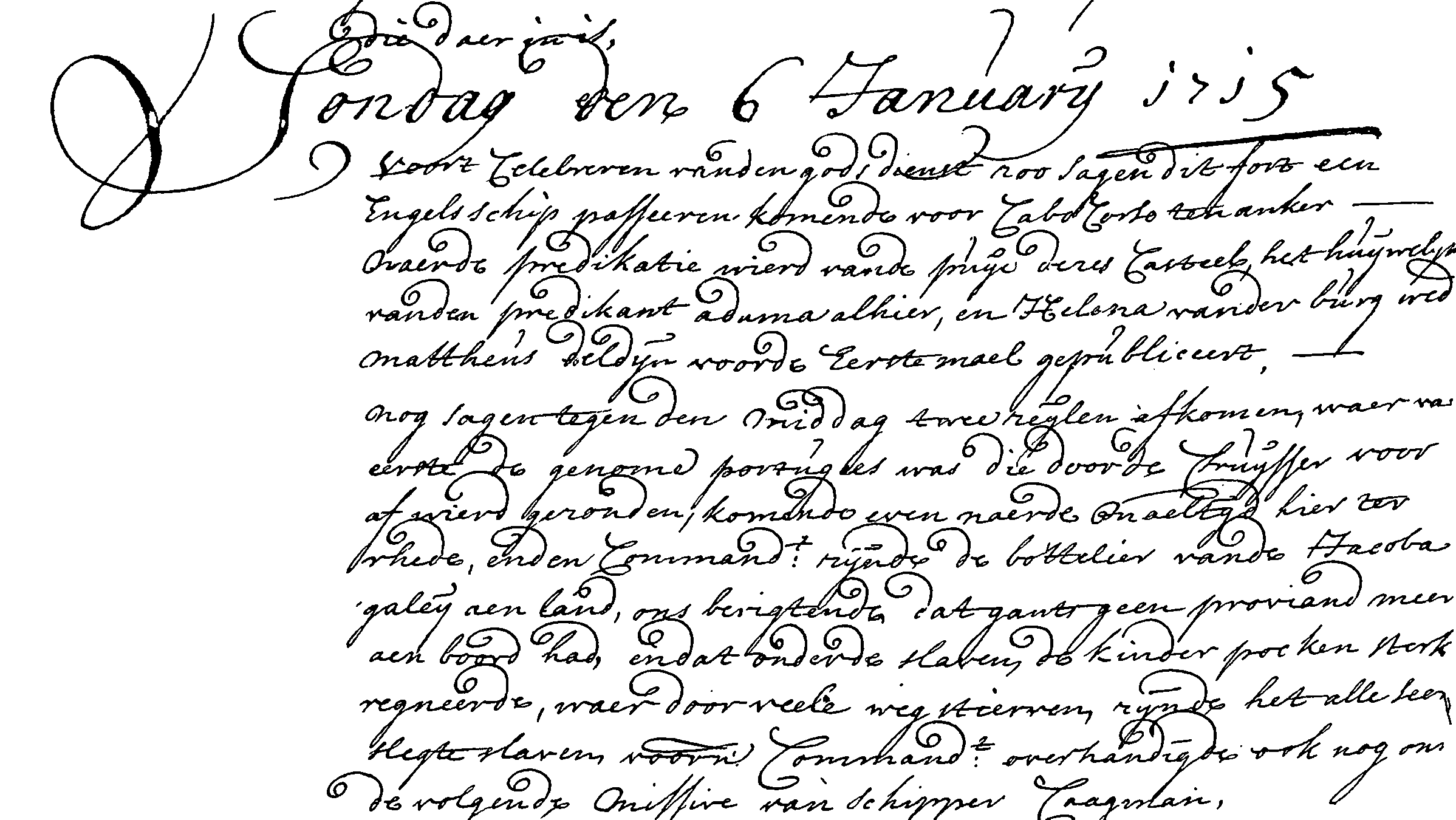}
		      \includegraphics[width=\columnwidth]{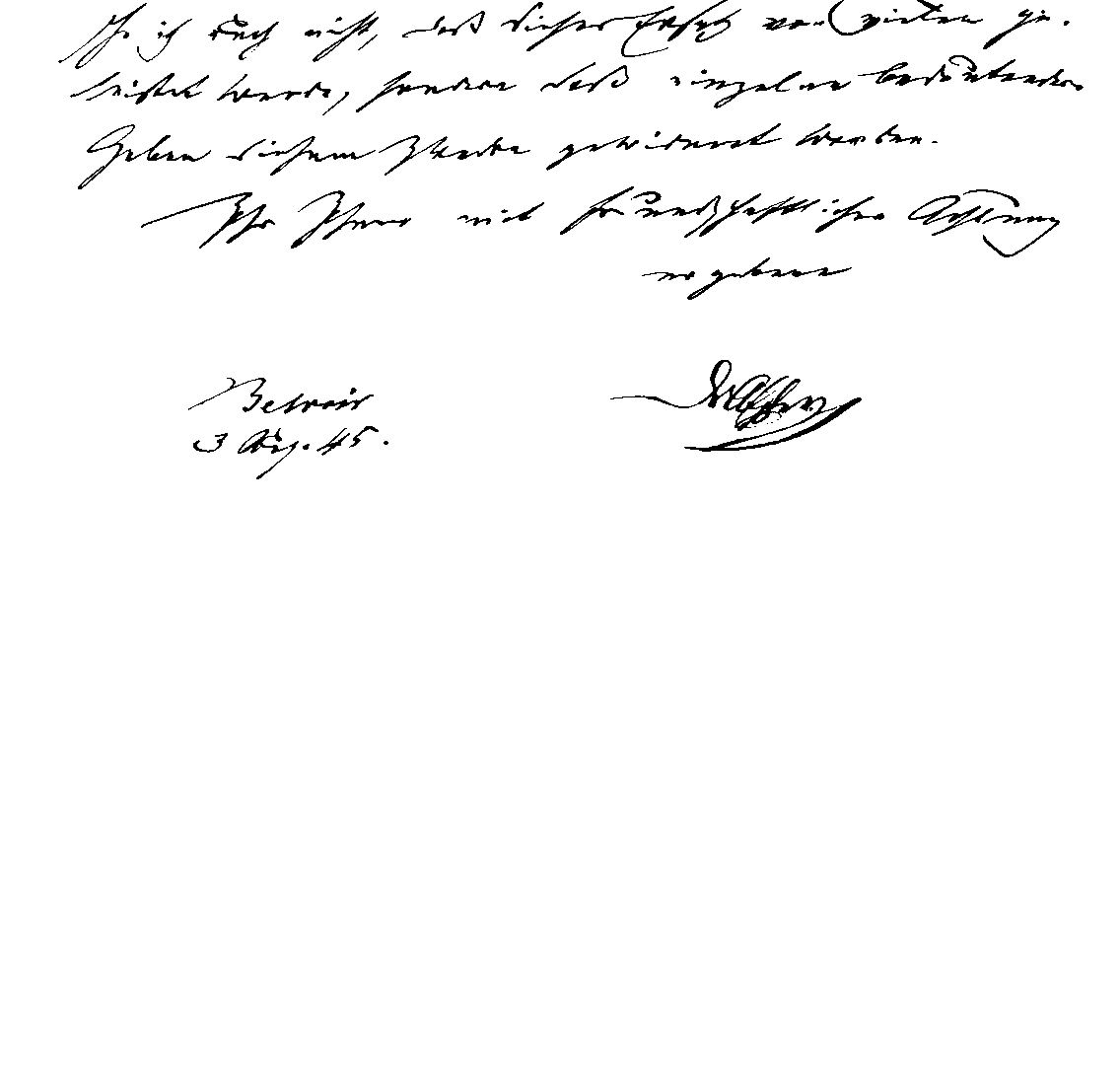}
            \end{minipage}
            \caption{}
            \label{sub:b}
		\end{subfigure}
		\begin{subfigure}{0.23\columnwidth}
			\begin{minipage}[t]{\columnwidth}
		      \centering
              \includegraphics[width=\columnwidth]{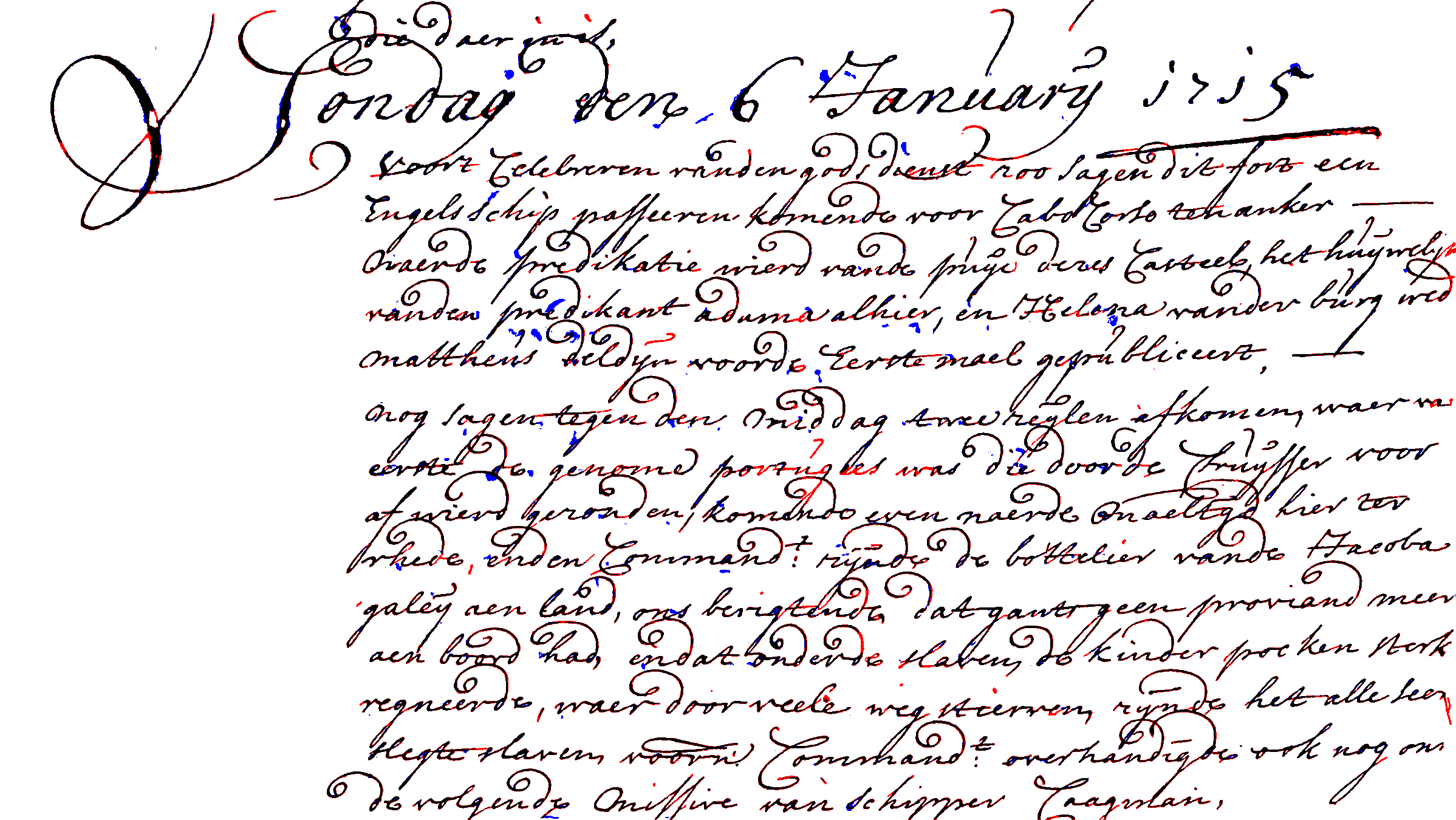}
		      \includegraphics[width=\columnwidth]{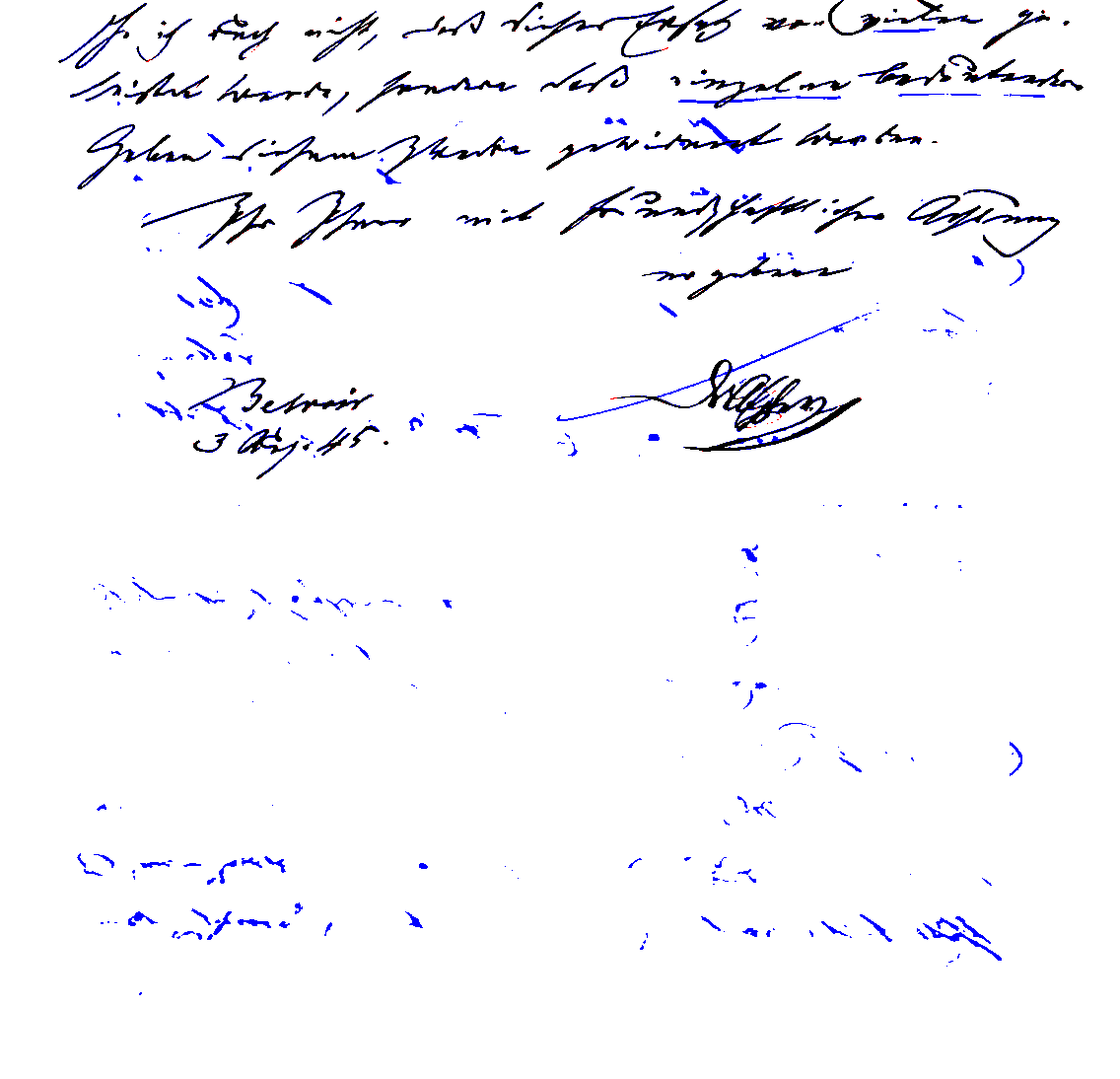}
            \end{minipage}
            \caption{}
            \label{sub:c}
		\end{subfigure}
		\begin{subfigure}{0.23\columnwidth}
        	\begin{minipage}[t]{\columnwidth}
		      \centering
              \includegraphics[width=\columnwidth]{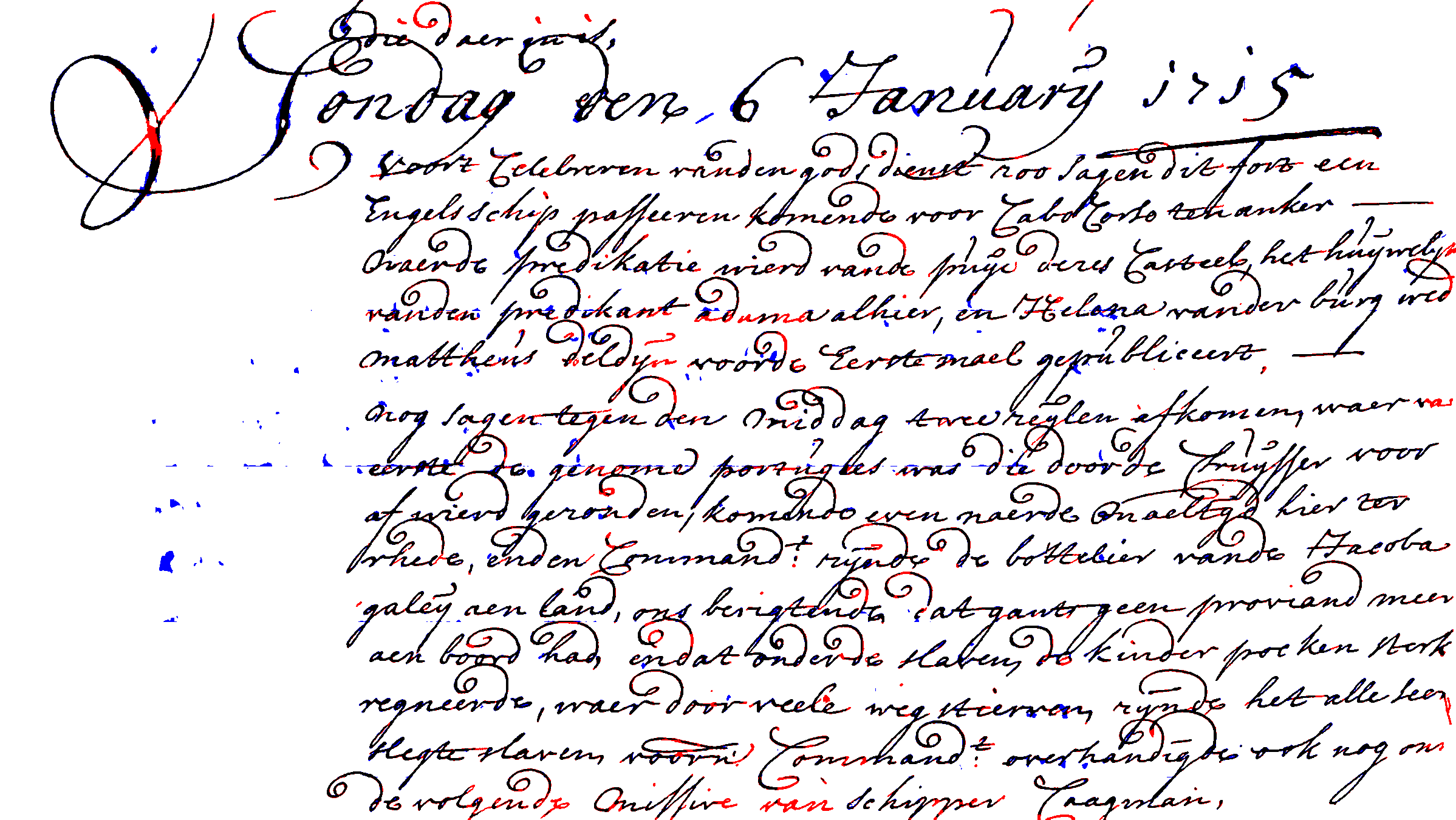}
		      \includegraphics[width=\columnwidth]{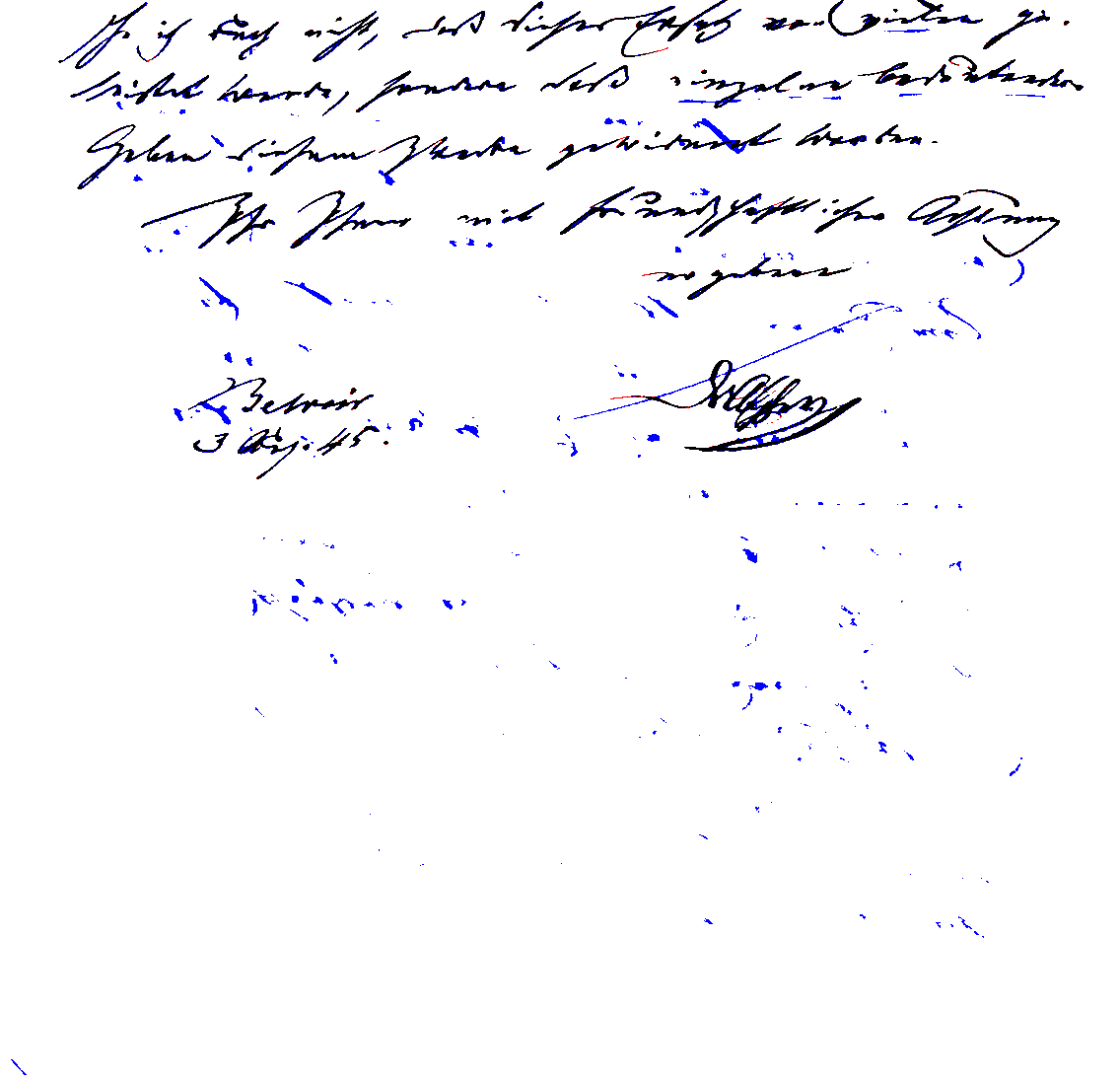}
            \end{minipage}
            \caption{}
            \label{sub:d}
		\end{subfigure}
        \hskip 0.5\columnwidth
        \begin{subfigure}{0.23\columnwidth}
        	\begin{minipage}[t]{\columnwidth}
		      \centering
              \includegraphics[width=\columnwidth]{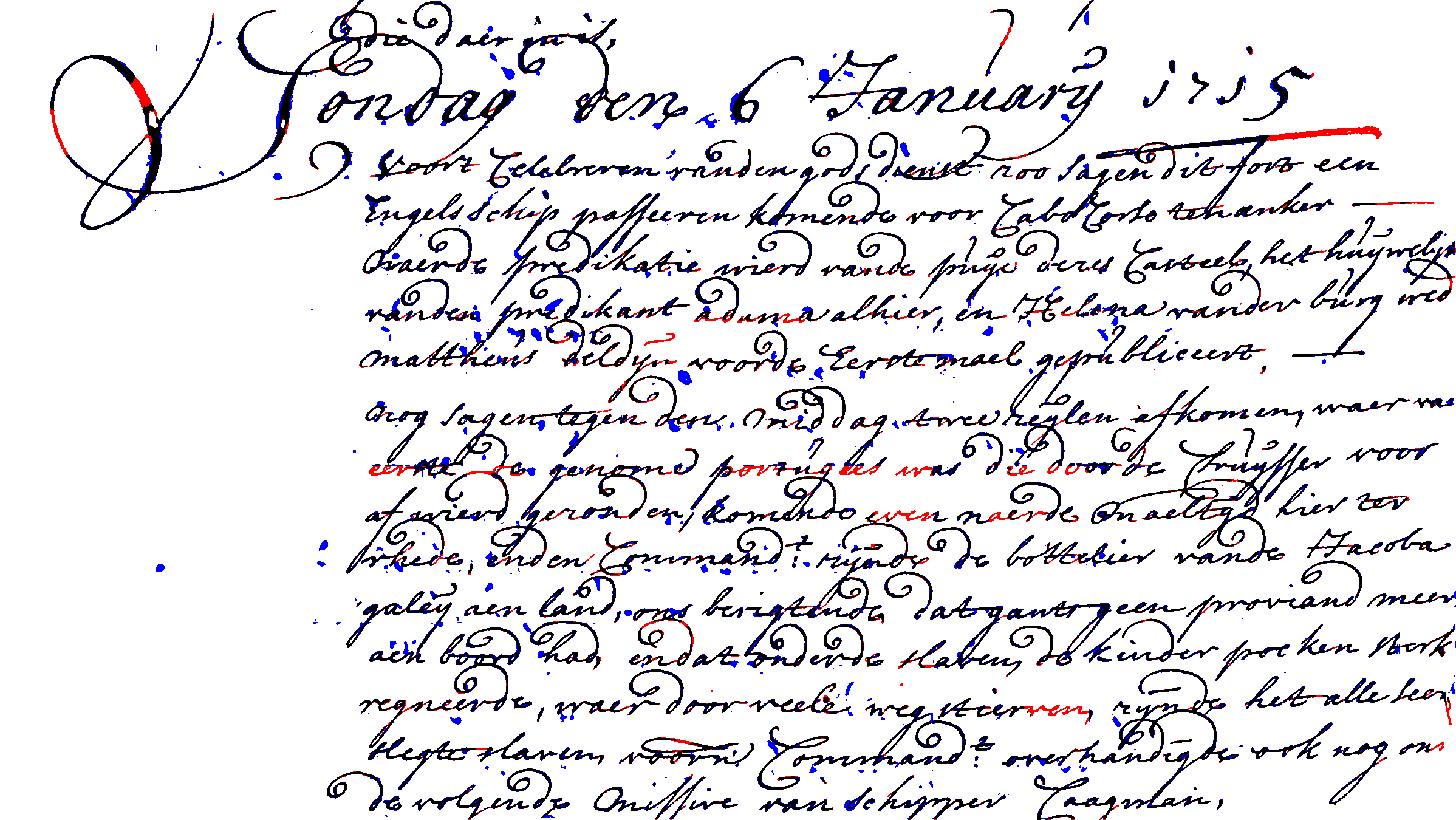}
		      \includegraphics[width=\columnwidth]{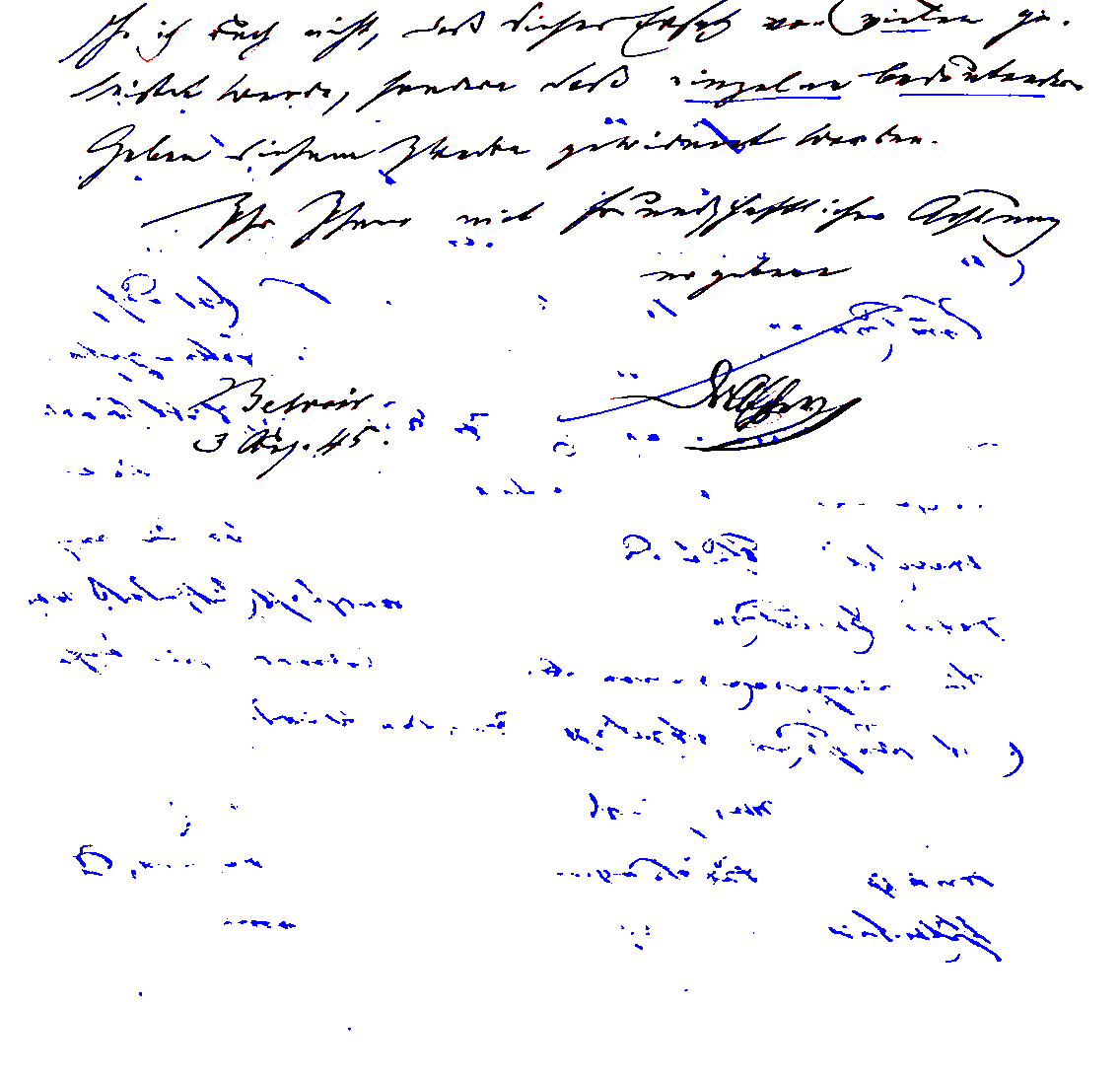}
            \end{minipage}
            \caption{}
            \label{sub:e}
		\end{subfigure}
		\begin{subfigure}{0.23\columnwidth}
        	\begin{minipage}[t]{\columnwidth}
		      \centering
              \includegraphics[width=\columnwidth]{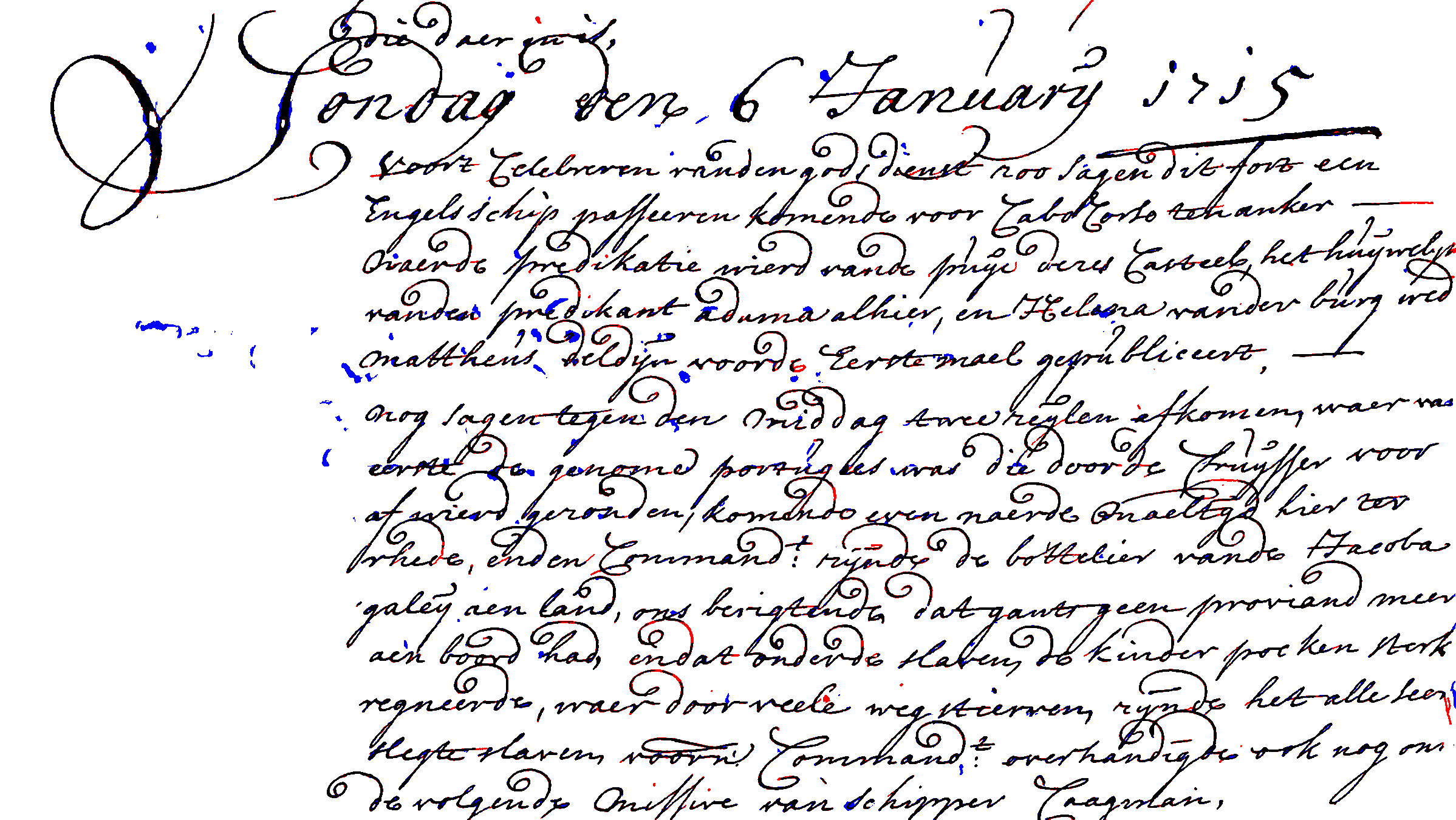}
		      \includegraphics[width=\columnwidth]{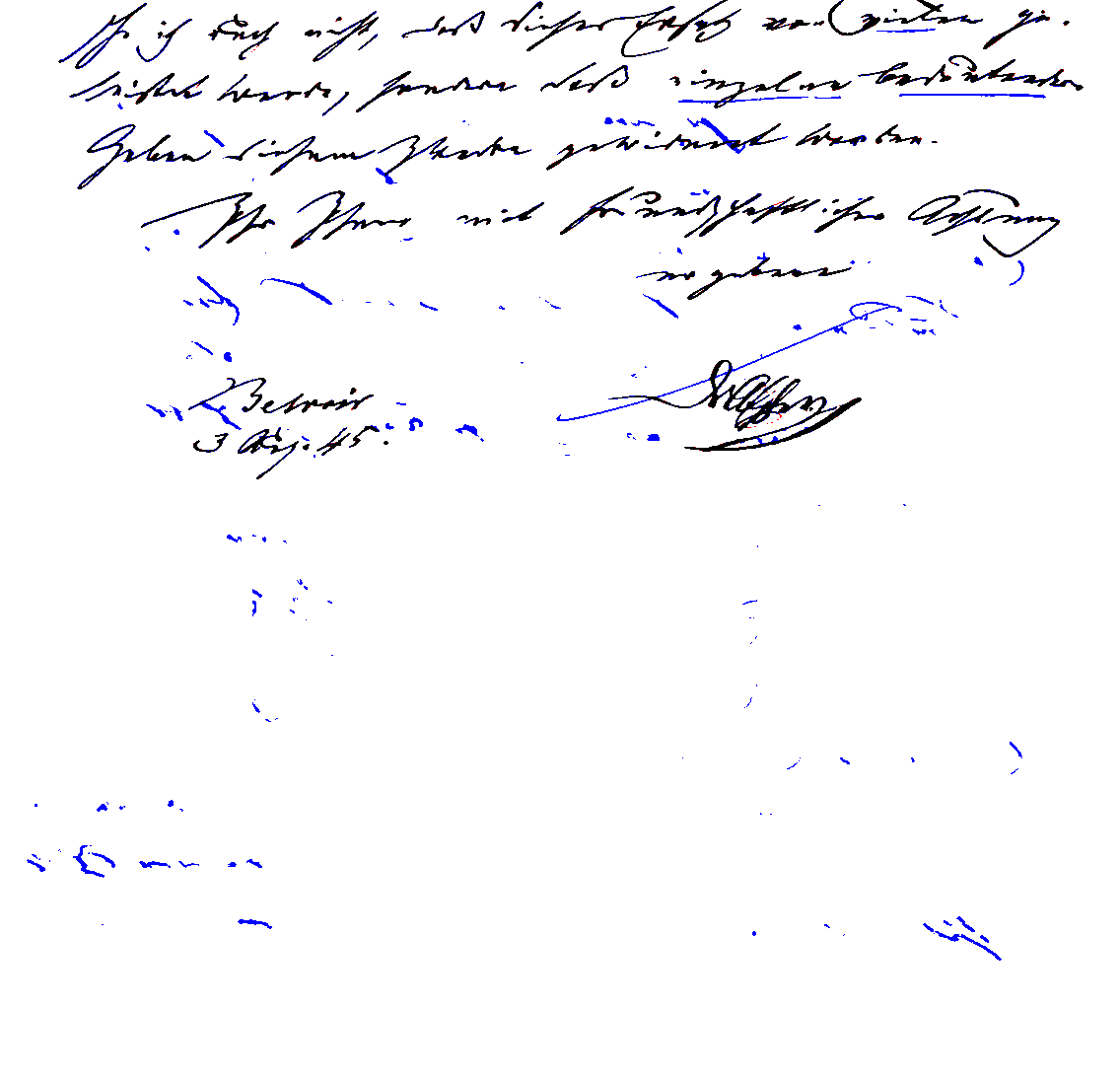}
            \end{minipage}
            \caption{}
            \label{sub:edge}
		\end{subfigure}
		\begin{subfigure}{0.23\columnwidth}
			\begin{minipage}[t]{\columnwidth}
		      \centering
              \includegraphics[width=\columnwidth]{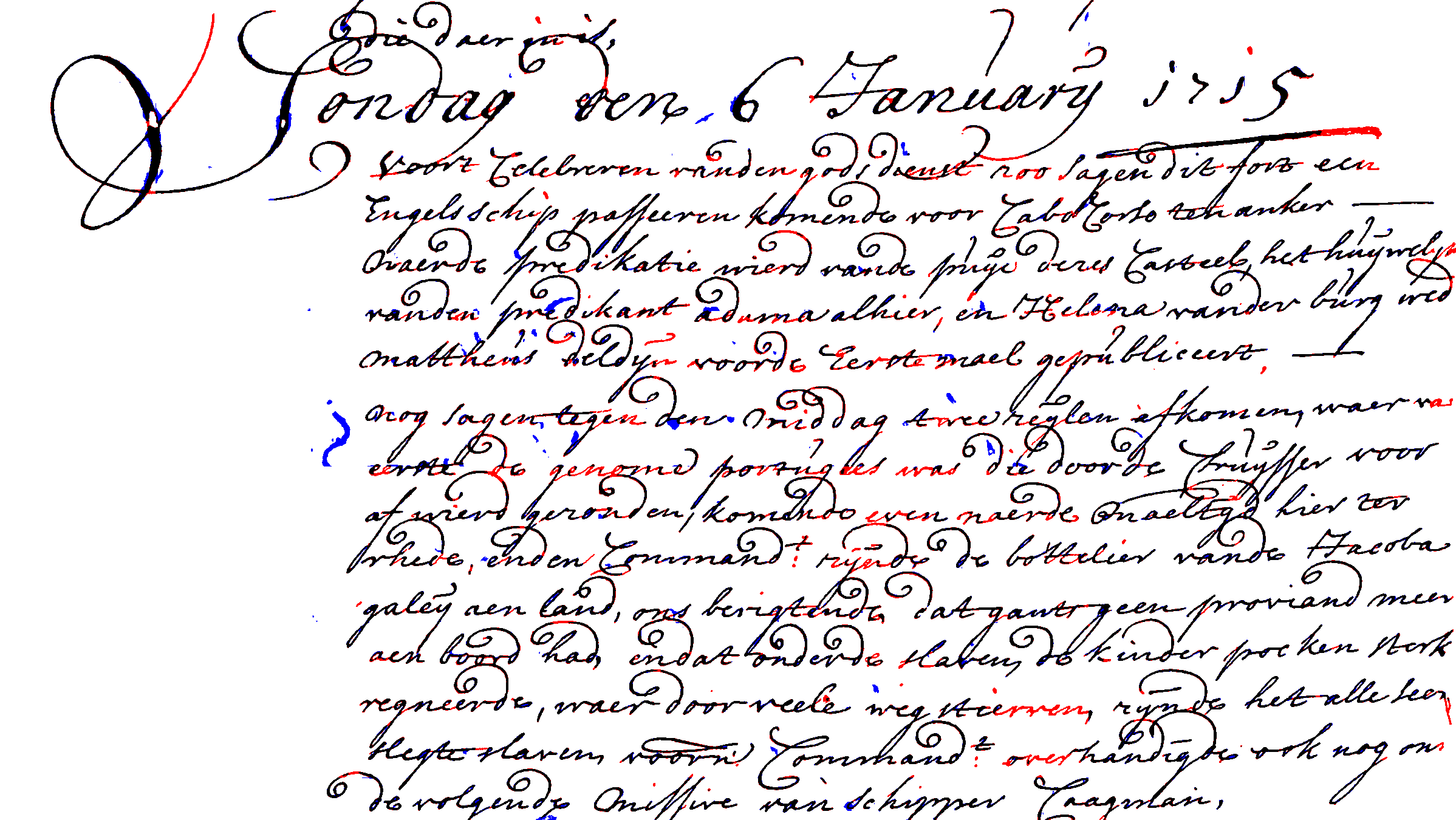}
		      \includegraphics[width=\columnwidth]{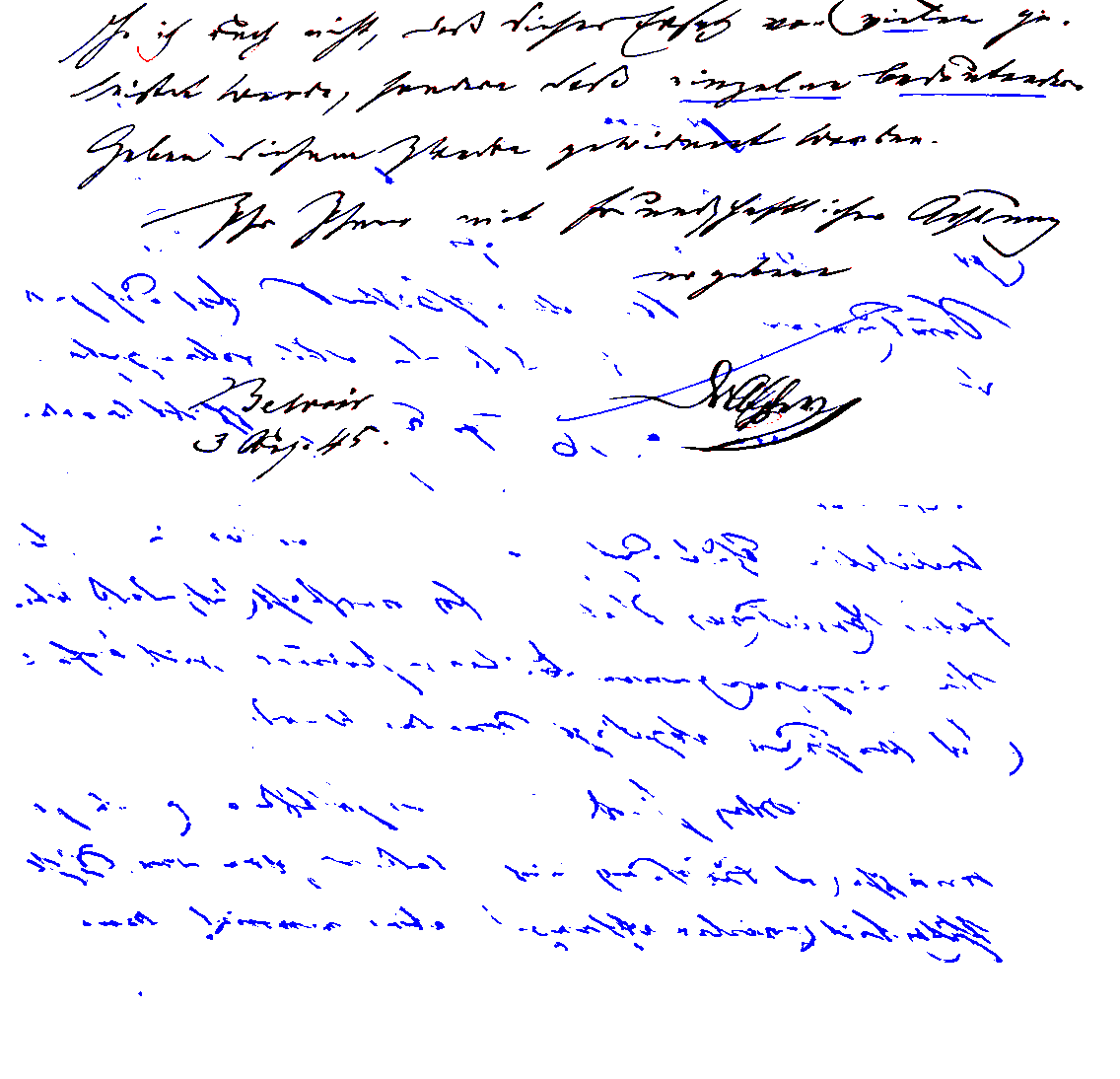}
            \end{minipage}
            \caption{}
            \label{sub:f}
		\end{subfigure}
		\begin{subfigure}{0.23\columnwidth}
			\begin{minipage}[t]{\columnwidth}
		      \centering
			  \includegraphics[width=\columnwidth]{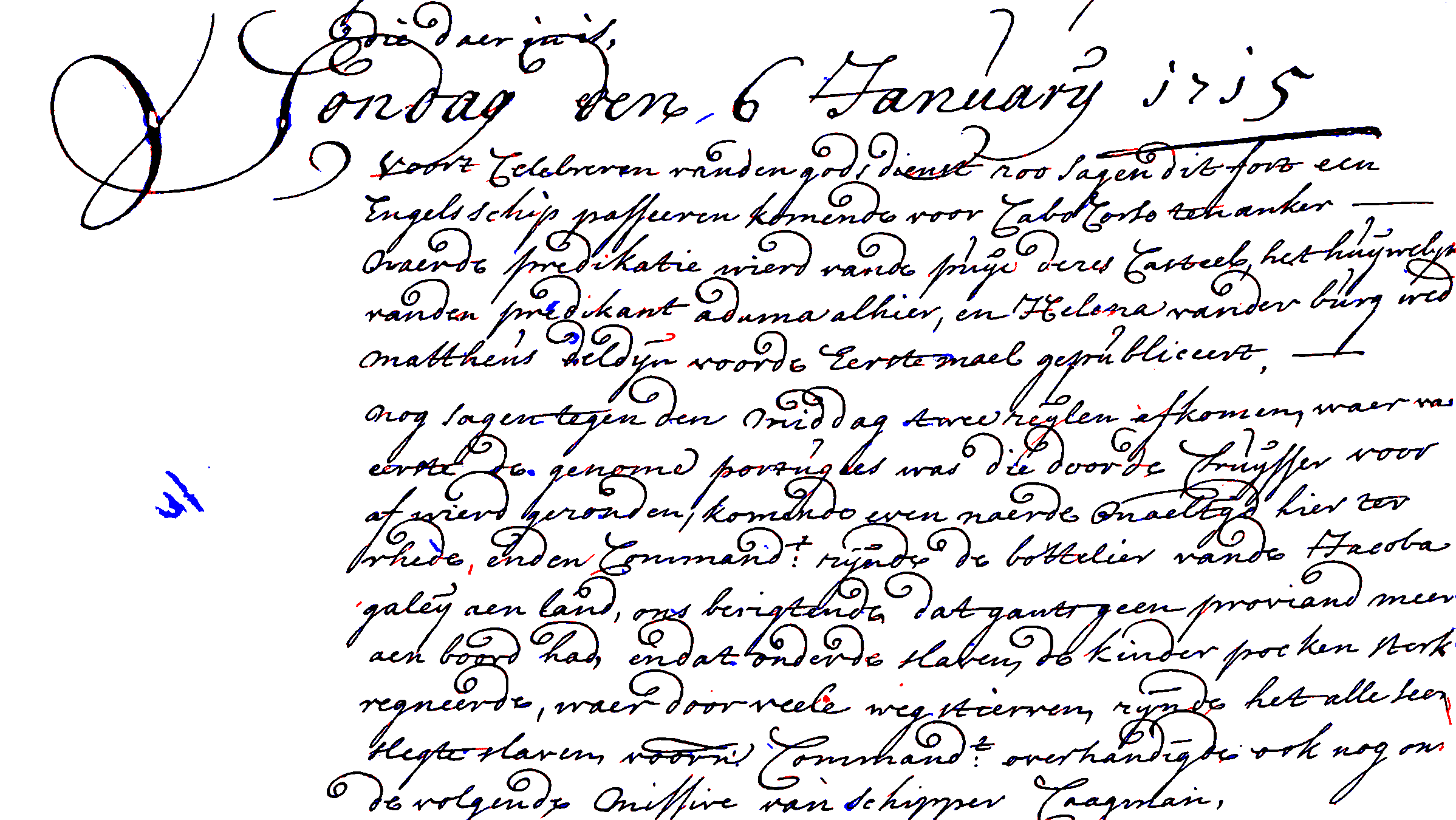}
              \includegraphics[width=\columnwidth]{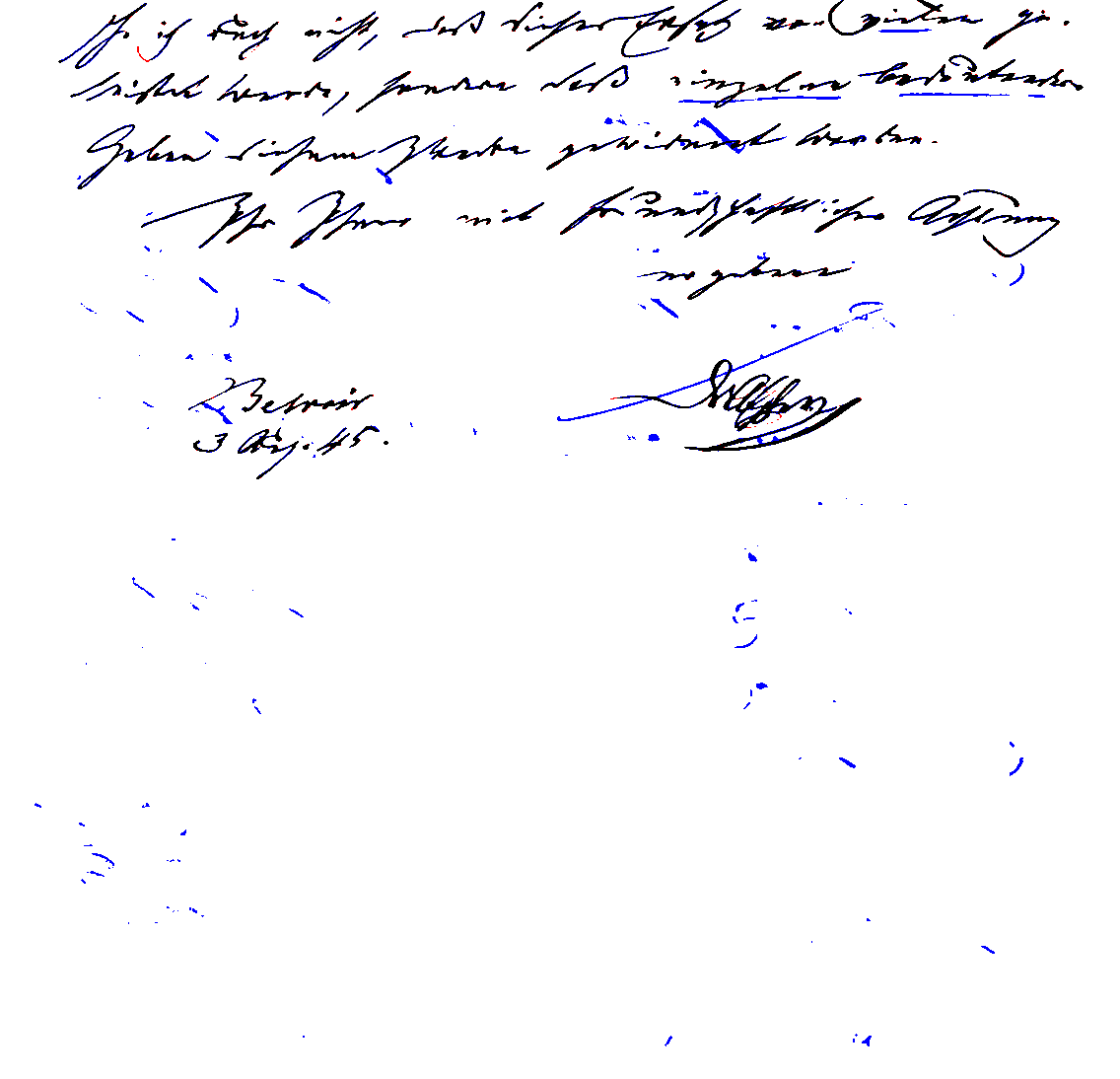}
            \end{minipage}
            \caption{}
            \label{sub:g}
		\end{subfigure}
		\caption{Binarization results of two example images in DIBCO'17 and DIBCO'19. (a) Original image, (b) Ground-truth, (c) RN w (MO + ME + EB), (d) RN w (GC + MO + EB), (e) CN w (GC + MO + ME + EB), (f) RN w (GC + MO + ME), (g) RN w (GC + ME + EB), (h) RN w (GC + MO + ME + EB). MO: Multi-scale operation. ME: Input mask and edge. GC: Gated convolution. EB: Edge branch. RN: Refinement sub-network. CN: Coarse sub-network.}
		\label{fig:abl}
\end{figure}

In this section, we conduct the ablation experiments on all (H-)DIBCO datasets to verify the benefits of different components of GDB: Gated convolution (GC), multi-scale operation (MO), refinement sub-network (RN), edge branch (EB) and input mask and edge (ME). \Cref{table:ablation} shows the average results for the 136 documents in the ten (H-)DIBCO datasets. \Cref{fig:abl} shows several qualitative examples.

\begin{table}[!htbp]
\centering
\renewcommand{\tablename}{Table}
\caption{Ablation study and average results on (H-)DIBCO datasets from 2009 to 2019. MO: Multi-scale operation. ME: Input mask and edge. GC: Gated convolution. RN: Refinement sub-network. CN: Coarse sub-network. EB: Edge branch. IO: Iterative operation. The best score is shown in \textbf{bold}.}
\label{table:ablation}
\resizebox{\columnwidth}{15mm}{
\begin{tabular}{ccccc}
\hline
Method                                   & FM             & p-FM           & PSNR           & DRD           \\ \hline
RN w (MO + ME + EB)            & 87.73          & 91.86          & 19.23          & 4.25          \\
RN w (GC + MO + EB)            & 87.99          & 90.32          & 19.64          & 4.11          \\
CN w (GC + MO + ME + EB)       & 89.78          & 91.21          & 19.08          & 4.50          \\
RN w (GC + MO + ME)            & 88.96          & 91.18          & 20.10          & 3.47          \\
RN w (GC + ME + EB)            & 90.89          & 92.84          & 20.03          & 3.33          \\
RN w (GC + MO + ME + EB)       & \textbf{91.16} & \textbf{93.06} & \textbf{20.20} & \textbf{3.24} \\
*RN w (GC + MO + ME + EB + IO) & 91.21          & 93.07          & 20.22          & 3.21          \\ \hline
Note: * results do not participate in rankings. \\
\end{tabular}}
\end{table}

\textbf{Gated convolutions} GDB with GC performs better on all four metrics than without GC. It indicates that replacing the vanilla convolutional layers with the gated convolutional layers can extract text features and suppress the background noise more effectively. GDB without GC (\cref{sub:c}) extracts the edge of the text incompletely, whereas GDB with GC (\cref{sub:g}) extracts the stroke edges more accurately and obtains binarization results of higher quality.

\textbf{Multi-scale operation} The performance of GDB can be enhanced by combining the local and global predicted results. In \cref{sub:g}, the bleed-through noise of large independent background areas is effectively suppressed by applying multi-scale feature fusion. The reason is that in a local image patch, pixels are more likely to be classified as text the darker they are compared to their surroundings (\cref{sub:f}), while the pixels of bleed-through noise are easily classified as background rather than text.

\textbf{Refinement sub-network} It is beneficial to cascade a refinement sub-network after the coarse sub-network for document binarization according to the results in the third and penultimate rows of \cref{table:ablation}. The refinement sub-network leads in all four metrics more than the coarse sub-network. The refinement sub-network (\cref{sub:g}) provides finer pen strokes and less noise overall than the coarse sub-network (\cref{sub:e}).

\textbf{Edge branch} The edge branch can supervise the extraction of stroke edges and avoid incorrect, excessive extraction of stroke edges. GDB with EB classifies stroke edges more accurately at the pixel level.

\textbf{Input mask and edge} On the one hand, this operation adds priori information to our model. On the other hand, the mask and edge can effectively guide the gated convolutions to change the gating values. It can be said that the input mask and edge are complementary to gated convolutions. \Cref{fig:abl} shows some qualitative results.

We can notice that the input and output of the coarse sub-network have the same pattern, except for the original patch. We utilize the output of our coarse sub-network as input for iterative operation (IO) similarly to \cite{He2019}. As shown in the last two rows of \cref{table:ablation}, GDB achieves better performance in all four metrics after one round of iteration. This indicates that the input mask and edge guide feature extraction by gated convolutions. However, the effect of binarization does not improve after two or more iterative operations because no iterative training is performed.

In the next section, we compare GDB(RN) with (GC + ME + EB) and GDB(RN) with (GC + MO + ME + EB) to other state-of-the-art methods, representing the local and global prediction results, respectively.

\subsection{Comparison with state-of-the-art methods and best competition system}

\begin{table*}[!htbp]
\centering
\renewcommand{\tablename}{Table}
\caption{Evaluation results for different binarization methods over all ten (H-)DIBCO datasets(best values are highlighted in \textcolor{red}{red}, second best are highlighted in \textcolor{blue}{blue}. MO: Multi-scale operation.}
\label{table:sota}
\resizebox{0.9\textwidth}{78mm}{
\begin{tabular}{ccccccccccc}
\hline
\multicolumn{2}{c}{Competition}         & Otsu \cite{Otsu1979}  & Sauvola \cite{Sauvola2000} & \begin{tabular}[c]{@{}c@{}}Best\\ Competition\\ System\end{tabular} & Howe \cite{Howe2013} & Jia\cite{Jia2018} & cGANs \cite{Zhao2019} & Suh \cite{Suh2022} & \begin{tabular}[c]{@{}c@{}} Ours\\w/o MO\end{tabular} & \begin{tabular}[c]{@{}c@{}} Ours\\w MO\end{tabular} \\ \hline
\multirow{4}{*}{DIBCO'09}	&FM &78.6 &85.37 &91.24 &94.04 &93.17 &\textcolor{blue}{94.1} &93.28 &93.56 &\textcolor{red}{94.79} \\
&p-FM &80.53 &89.08 &- &95.06 &\textcolor{blue}{95.32} &95.26 &94.94 &94.88 &\textcolor{red}{96.36} \\
&PSNR &15.31 &16.37 &18.66 &20.43 &19.44 &20.3 &19.74 &\textcolor{blue}{20.44} &\textcolor{red}{20.68} \\
&DRD &22.57 &7.08 &- &2.1 &2.29 &\textcolor{blue}{1.82} &2.79 &2.48 &\textcolor{red}{1.7} \\ \hline
\multirow{4}{*}{H-DIBCO'10}	&FM &85.43 &75.18 &91.50 &93.59 &91.75 &94.03 &93.92 &\textcolor{red}{95.45} &\textcolor{blue}{95.19} \\
&p-FM &90.64 &84.08 &- &94.81 &95.36 &95.39 &96.53 &\textcolor{red}{97.43} &\textcolor{blue}{96.62} \\
&PSNR &17.52 &15.94 &19.78 &21.08 &19.84 &21.12 &21.18 &\textcolor{red}{22.31} &\textcolor{blue}{21.98} \\
&DRD &4.05 &7.22 &- &1.72 &2.17 &1.58 &1.50 &\textcolor{red}{1.21} &\textcolor{blue}{1.32} \\ \hline
\multirow{4}{*}{DIBCO'11}	&FM &82.1 &82.14 &88.74 &90.79 &91.65 &\textcolor{blue}{93.81} &93.44 &\textcolor{red}{94.64} &93.44 \\
&p-FM &85.96 &87.7 &- &92.28 &95.56 &95.7 &\textcolor{blue}{96.18} &\textcolor{red}{96.74} &95.82 \\
&PSNR &15.72 &15.65 &17.97 &19.01 &18.88 &\textcolor{blue}{20.26} &19.97 &\textcolor{red}{20.74} &20.1 \\
&DRD &8.95 &8.5 &5.36 &4.46 &2.66 &\textcolor{blue}{1.81} &1.93 &\textcolor{red}{1.66} &2.25 \\ \hline
\multirow{4}{*}{H-DIBCO'12}	&FM &75.07 &81.56 &92.85 &93.73 &92.96 &94.96 &94.5 &\textcolor{blue}{95.39} &\textcolor{red}{95.8} \\
&p-FM &78.14 &87.35 &- &94.24 &95.76 &96.15 &\textcolor{red}{97.36} &96.63 &\textcolor{blue}{97.03} \\
&PSNR &15.03 &16.88 &21.80 &21.85 &20.43 &21.91 &21.78 &\textcolor{blue}{22.31} &\textcolor{red}{22.62} \\
&DRD &26.46 &6.46 &2.66 &2.1 &2.3 &\textcolor{blue}{1.55} &1.73 &1.59 &\textcolor{red}{1.32} \\ \hline
\multirow{4}{*}{DIBCO'13}	&FM &80.04 &82.71 &92.70 &91.34 &93.28 &\textcolor{blue}{95.28} &94.75 &\textcolor{red}{95.86} &95.19 \\
&p-FM &83.43 &87.74 &94.19 &91.79 &96.58 &96.47 &\textcolor{blue}{97.2} &\textcolor{red}{97.22} &96.37 \\
&PSNR &16.63 &17.02 &21.29 &21.29 &20.76 &22.23 &21.79 &\textcolor{red}{22.89} &\textcolor{blue}{22.58} \\
&DRD &10.98 &7.64 &3.10 &3.18 &2.01 &\textcolor{blue}{1.39} &1.66 &\textcolor{red}{1.27} &1.78 \\ \hline
\multirow{4}{*}{H-DIBCO'14}	&FM &91.62 &84.7 &\textcolor{blue}{96.88} &96.49 &94.89 &96.41 &96.19 &\textcolor{red}{97.58} &96.66 \\
&p-FM &95.69 &87.88 &97.65 &97.38 &\textcolor{blue}{97.68} &97.55 &97.13 &\textcolor{red}{98.27} &97.26 \\
&PSNR &18.72 &17.81 &22.66 &22.24 &20.53 &22.12 &21.77 &\textcolor{red}{23.74} &\textcolor{blue}{23.13} \\
&DRD &2.65 &4.77 &\textcolor{blue}{0.90} &1.08 &1.5 &1.07 &1.14 &\textcolor{red}{0.72} &1.21 \\ \hline
\multirow{4}{*}{H-DIBCO'16}	&FM &86.59 &84.64 &88.72 &87.47 &90.01 &\textcolor{red}{91.66} &\textcolor{blue}{91.11} &89.94 &90.41 \\
&p-FM &89.92 &88.39 &91.84 &92.28 &93.72 &94.58 &\textcolor{red}{95.22} &94.6 &\textcolor{blue}{94.7} \\
&PSNR &17.79 &17.09 &18.45 &18.05 &19.0 &\textcolor{red}{19.64} &\textcolor{blue}{19.34} &18.9 &19.0 \\
&DRD &5.58 &6.27 &3.86 &5.35 &4.03 &\textcolor{red}{2.82} &\textcolor{blue}{3.25} &3.55 &3.34 \\ \hline
\multirow{4}{*}{DIBCO'17}	&FM &77.73 &77.11 &91.04 &90.1 &85.66 &90.73 &90.95 &\textcolor{blue}{91.33} &\textcolor{red}{94.32} \\
&p-FM &77.89 &84.1 &92.86 &90.95 &88.3 &92.58 &\textcolor{blue}{94.65} &93.84 &\textcolor{red}{96.58} \\
&PSNR &13.85 &14.25 &18.28 &\textcolor{blue}{18.52} &16.4 &17.83 &18.4 &18.34 &\textcolor{red}{20.04} \\
&DRD &15.54 &8.85 &3.40 &5.12 &7.67 &3.58 &\textcolor{blue}{2.93} &3.24 &\textcolor{red}{1.79} \\ \hline
\multirow{4}{*}{H-DIBCO'18}	&FM &51.45 &67.81 &88.34 &80.84 &76.05 &87.73 &\textcolor{red}{91.86} &90.78 &\textcolor{blue}{91.09} \\
&p-FM &53.05 &74.08 &90.24 &82.85 &80.36 &90.6 &\textcolor{red}{96.25} &94.25 &\textcolor{blue}{94.57} \\
&PSNR &9.74 &13.78 &19.11 &16.67 &16.9 &18.37 &\textcolor{red}{20.03} &19.68 &\textcolor{blue}{19.92} \\
&DRD &59.07 &17.69 &4.92 &11.96 &8.13 &4.58 &\textcolor{red}{2.6} &3.14 &\textcolor{blue}{3.07} \\ \hline
\multirow{4}{*}{DIBCO'19} & FM   & 47.83  & 51.73   & 72.88                 & 48.20 & 55.87      & 62.33 & 70.64      & \textcolor{red}{73.88}       & \textcolor{blue}{73.51}     \\
& p-FM & 45.59  & 55.15   & 72.15 & 48.37 & 56.28      & 62.89 & 70.80      & \textcolor{red}{74.96}       & \textcolor{blue}{74.81}     \\
& PSNR & 9.08   & 13.72   & 14.48 & 11.38 & 11.34      & 12.43 & 14.72      & \textcolor{blue}{14.80}       & \textcolor{red}{14.96}     \\
& DRD  & 109.46 & 13.83   & 16.24 & 36.03 & 43.81      & 17.65 & \textcolor{red}{9.75}       & \textcolor{blue}{10.41}       & 10.75     \\ \hline
\multirow{4}{*}{\begin{tabular}[c]{@{}l@{}}The Mean Values\\ \quad of 136 images\end{tabular}} & FM   & 74.22  & 75.98   & 88.59  & 84.68 & 84.92      & 88.64 & 89.92      & \textcolor{blue}{90.89}       & \textcolor{red}{91.16}     \\
& p-FM & 76.27  & 81.25   & - & 85.84 & 87.73      & 90.18 & 92.40      & \textcolor{blue}{92.84}       & \textcolor{red}{93.06}     \\
& PSNR & 14.54  & 15.65   & 18.93 & 18.63 & 17.88      & 19.17 & 19.53      & \textcolor{blue}{20.03}       & \textcolor{red}{20.20}     \\
& DRD  & 30.36  & 9.06    & -   & 8.80  & 9.69       & 4.53  & \textcolor{blue}{3.29}       & 3.33        & \textcolor{red}{3.24} \\ \hline
\multirow{4}{*}{\begin{tabular}[c]{@{}l@{}}The Mean Values\\ \quad of 10 DIBCO\end{tabular}} & FM   & 75.65 & 	77.30 &	89.49 &	86.66 &	86.53 &	90.10 	&91.06 &	\textcolor{blue}{91.84} & \textcolor{red}{92.04}   \\
& p-FM & 78.08 &	82.56 &	-	&88.00& 	89.49 &	91.72 &	93.63 	&\textcolor{blue}{93.88} &	\textcolor{red}{94.01} \\
& PSNR & 14.94 &	15.85 &	19.25 	&19.05 &	18.35 	&19.62 &	19.87 &	\textcolor{blue}{20.41 	}&\textcolor{red}{20.50} \\
& DRD  & 26.53 &	8.83 &	-	&7.31 	&7.66 &	3.78 &	\textcolor{blue}{2.93} 	&\textcolor{blue}{2.93} &	\textcolor{red}{2.85}  \\ \hline
\end{tabular}}
\end{table*}

\begin{figure}[!htbp]
		\centering
		\begin{subfigure}{0.3\columnwidth}
            \begin{minipage}[t]{\columnwidth}
		      \centering
		      \includegraphics[width=\columnwidth]{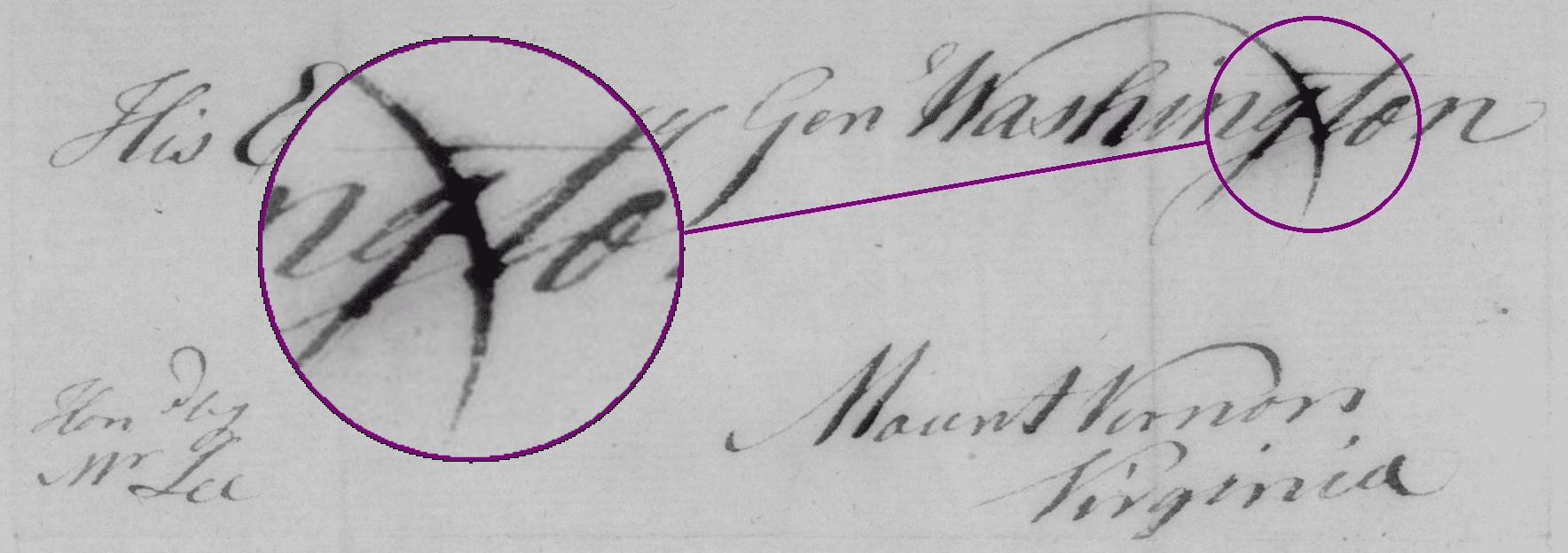}

              \includegraphics[width=\columnwidth]{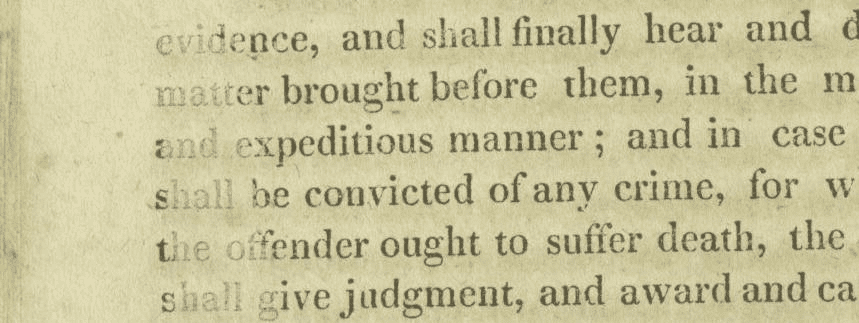}
              
              \includegraphics[width=\columnwidth]{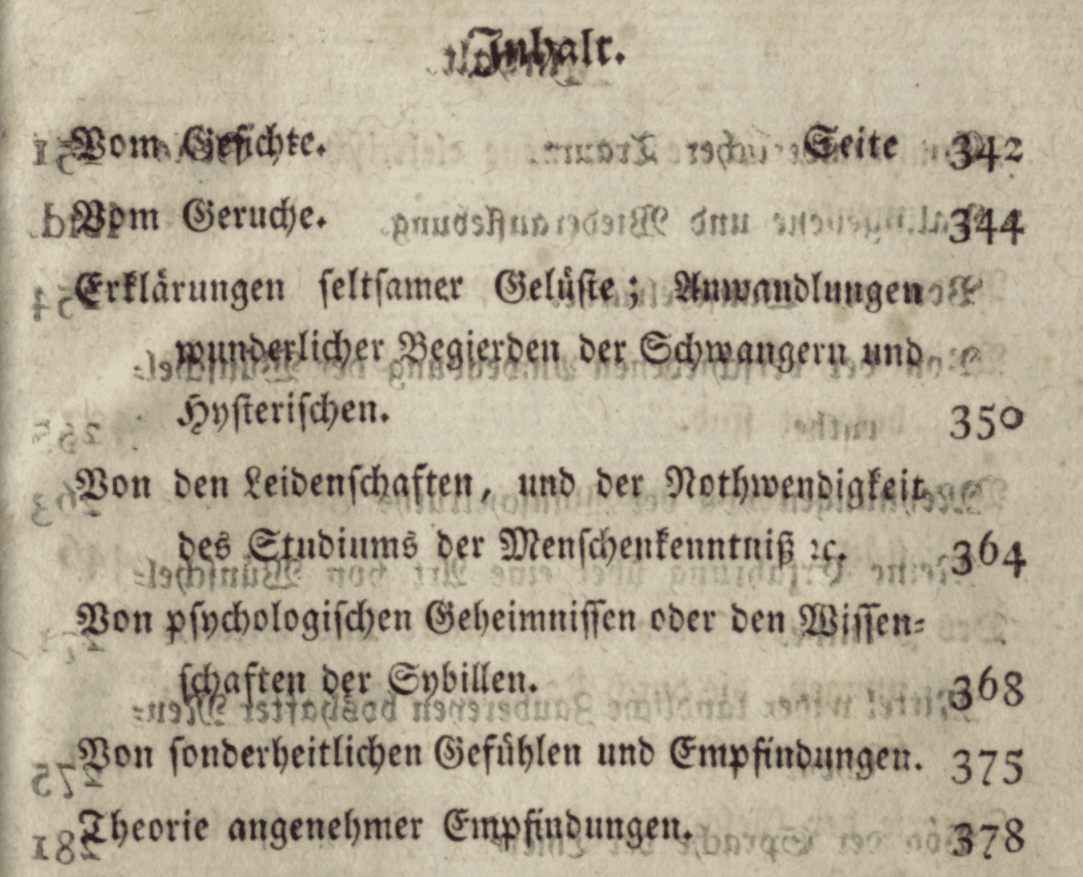}
            \end{minipage}
            \caption{Original image}
		\end{subfigure}
		\begin{subfigure}{0.3\columnwidth}
			\begin{minipage}[t]{\columnwidth}
		      \centering
		      \includegraphics[width=\columnwidth]{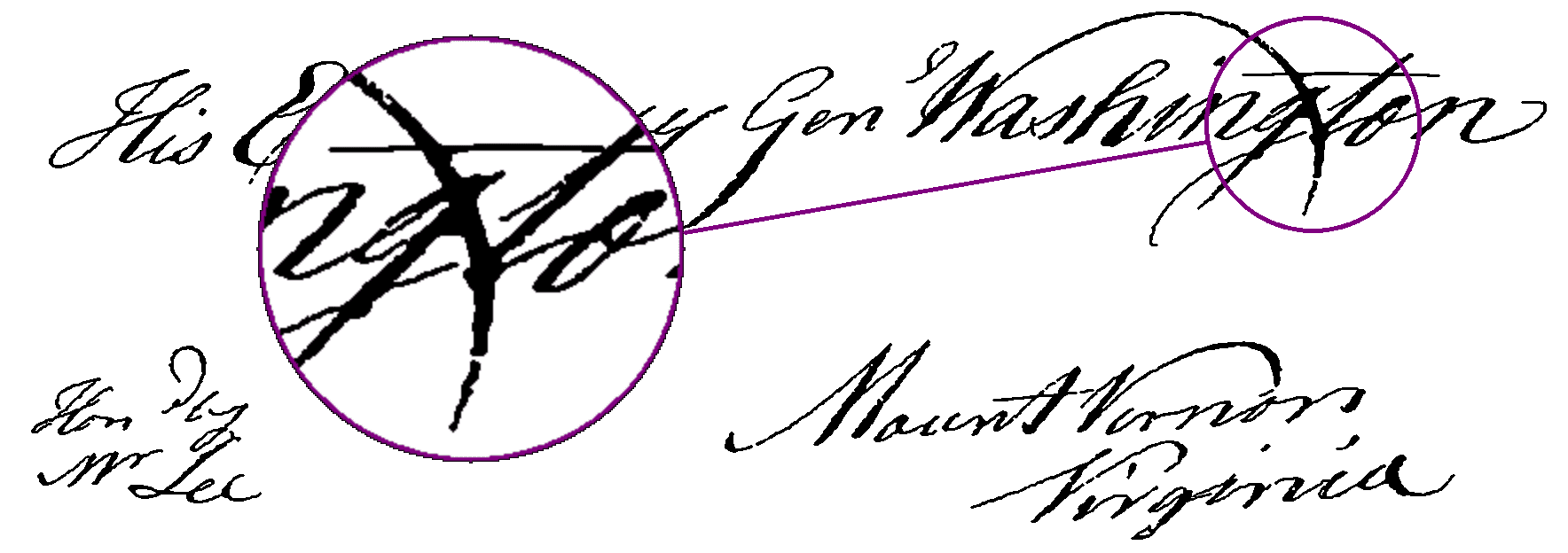}

              \includegraphics[width=\columnwidth]{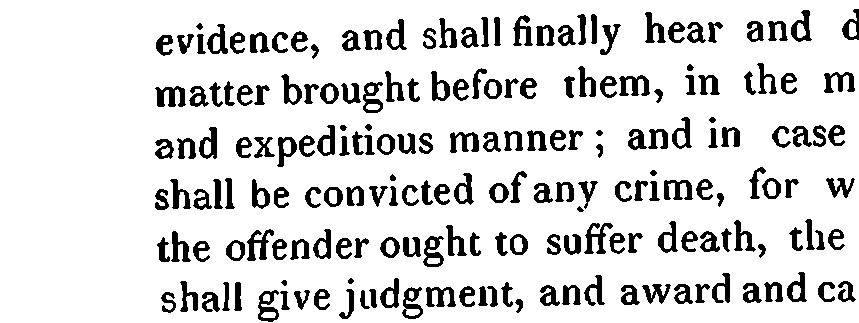}
              
              \includegraphics[width=\columnwidth]{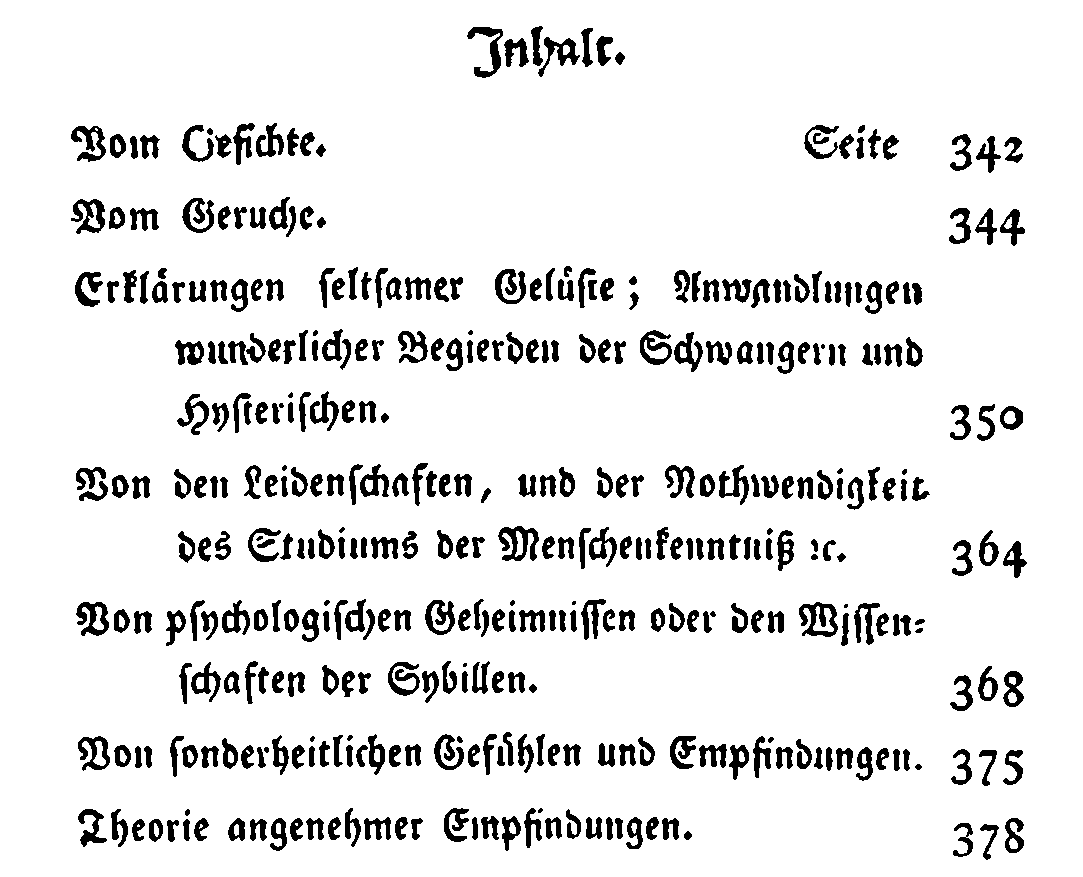}
            \end{minipage}
            \caption{Ground-truth}
		\end{subfigure}
		\begin{subfigure}{0.3\columnwidth}
			\begin{minipage}[t]{\columnwidth}
		      \centering
		      \includegraphics[width=\columnwidth]{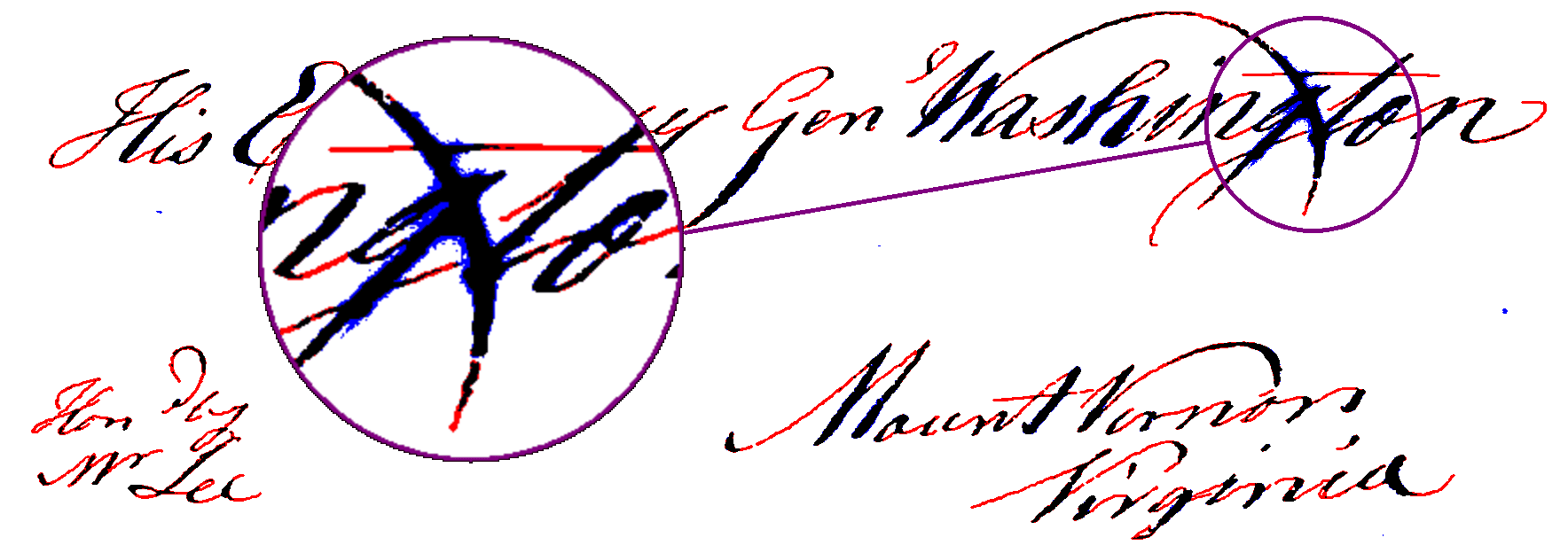}

              \includegraphics[width=\columnwidth]{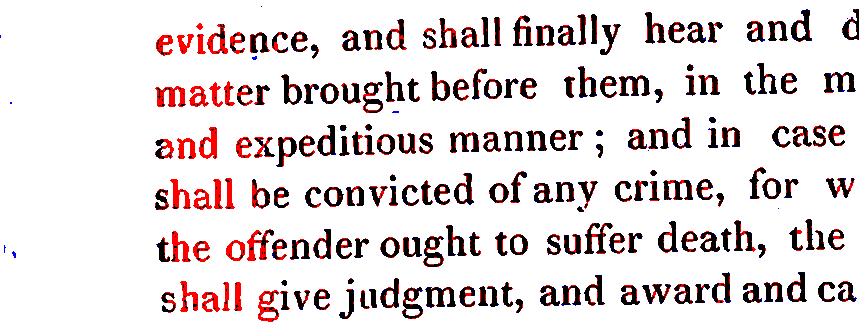}
              
              \includegraphics[width=\columnwidth]{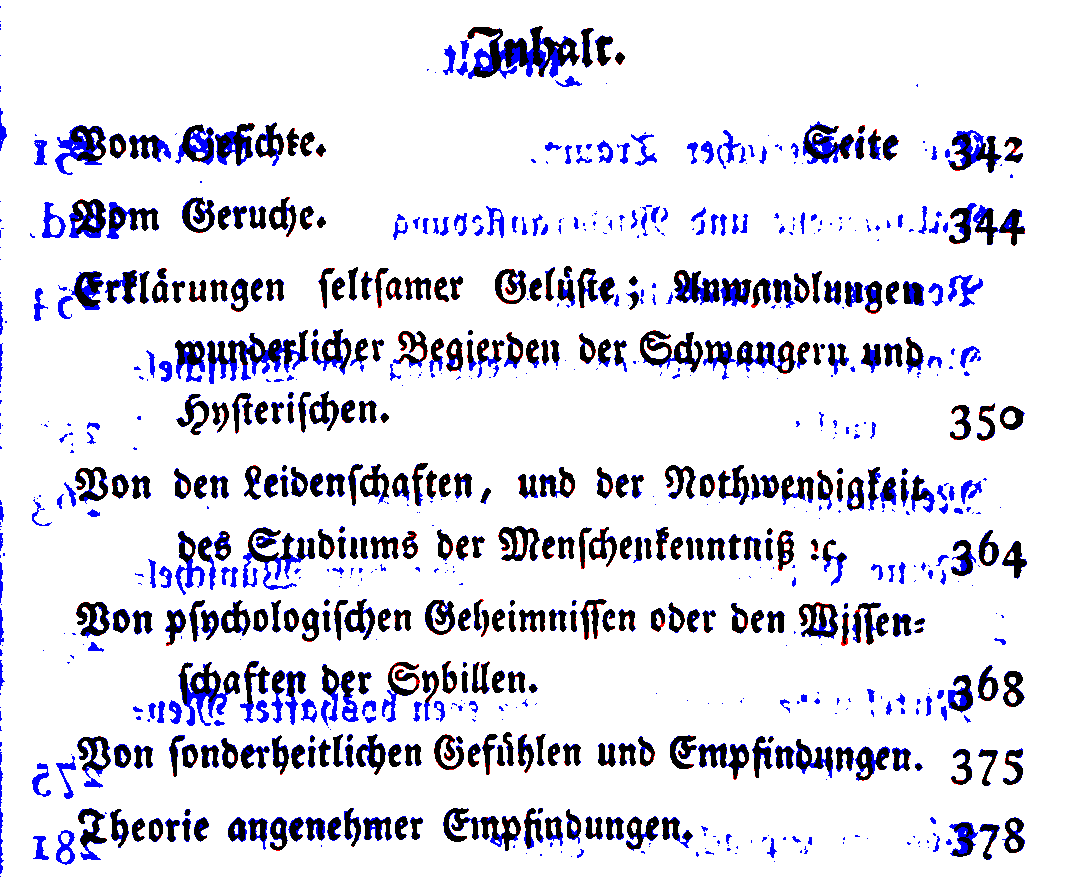}
            \end{minipage}
            \caption{Otsu \cite{Otsu1979}}
		\end{subfigure}
		\hskip 0.5\columnwidth
		\begin{subfigure}{0.3\columnwidth}
			\begin{minipage}[t]{\columnwidth}
		      \centering
		      \includegraphics[width=\columnwidth]{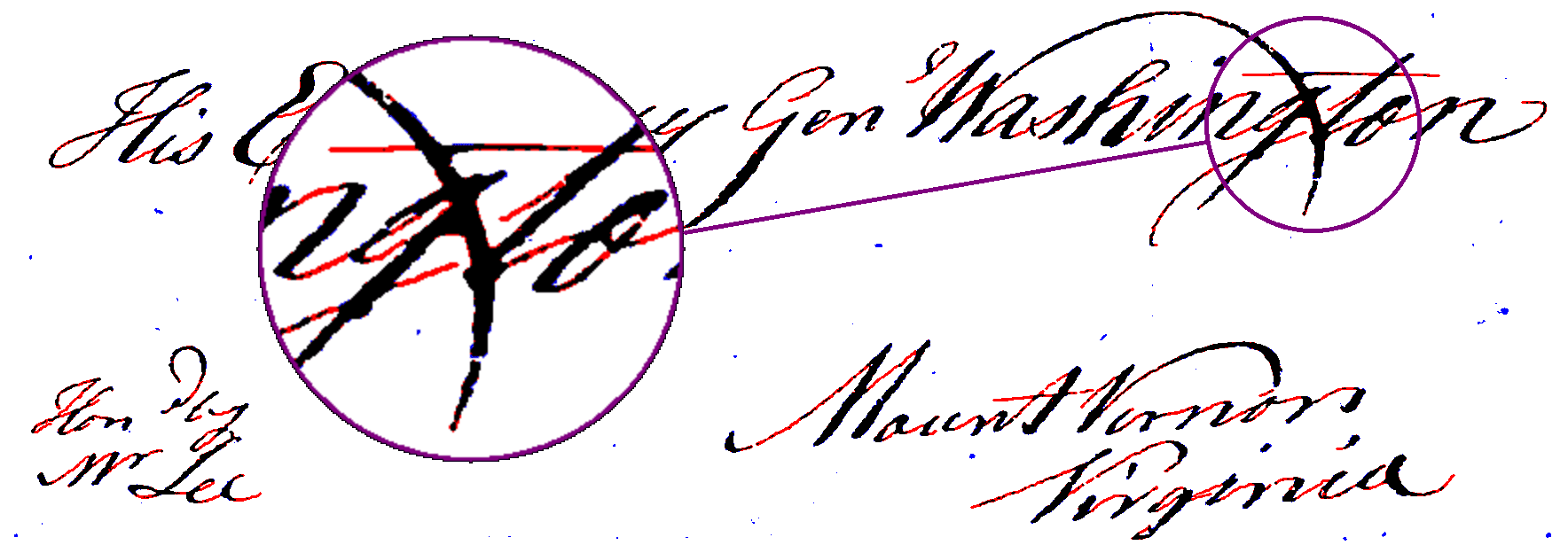}

              \includegraphics[width=\columnwidth]{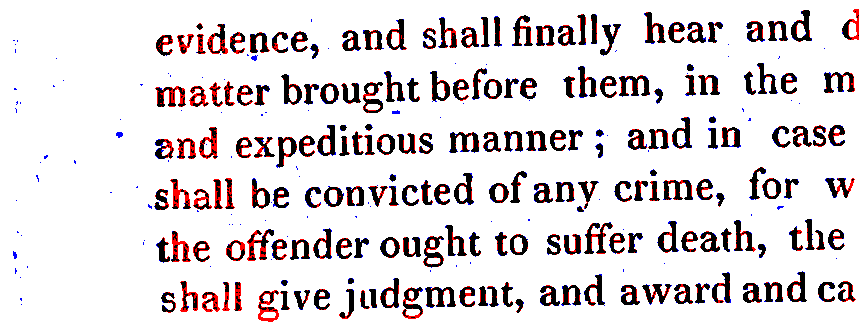}
              
              \includegraphics[width=\columnwidth]{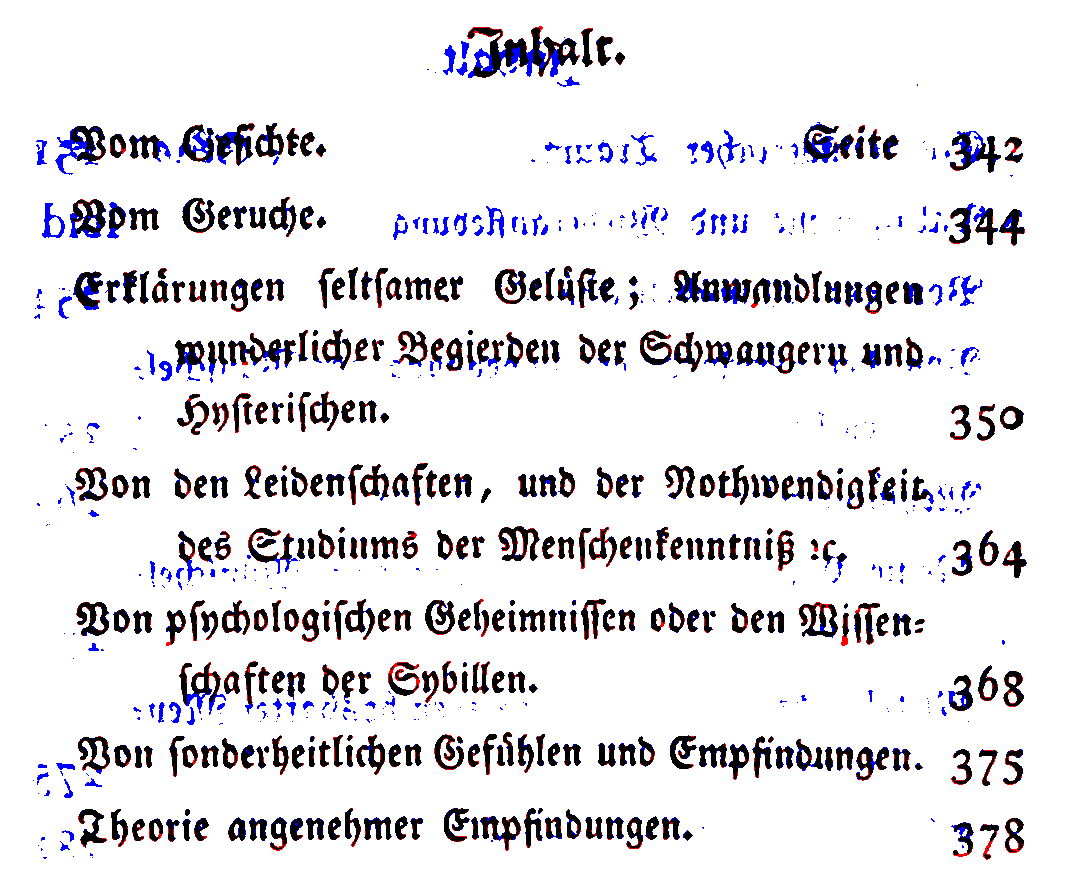}
            \end{minipage}
            \caption{Sauvola \cite{Sauvola2000}}
		\end{subfigure}
		\begin{subfigure}{0.3\columnwidth}
			\begin{minipage}[t]{\columnwidth}
		      \centering
		      \includegraphics[width=\columnwidth]{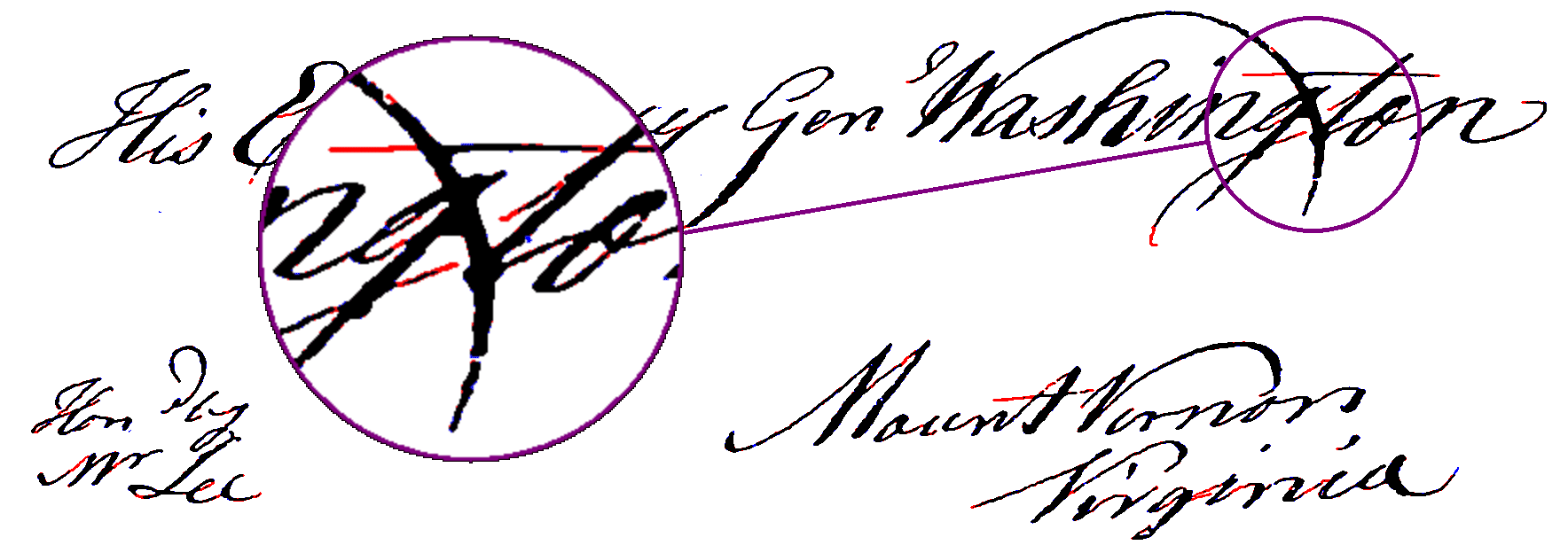}

              \includegraphics[width=\columnwidth]{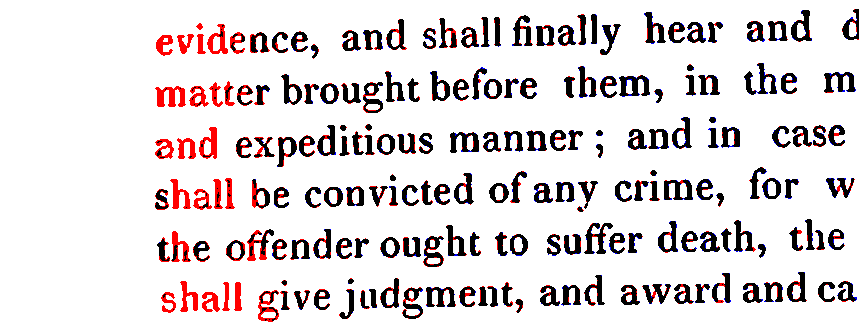}
              
              \includegraphics[width=\columnwidth]{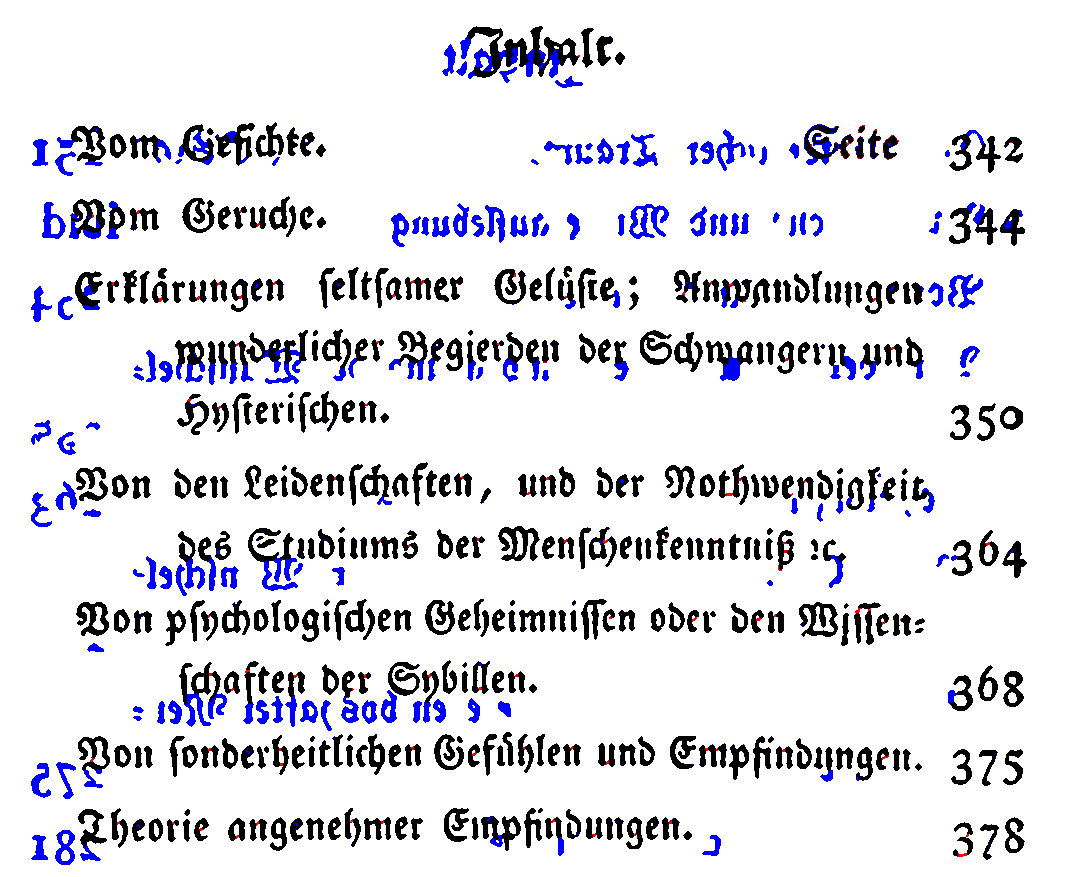}
            \end{minipage}
            \caption{Howe \cite{Howe2013}}
		\end{subfigure}
		\begin{subfigure}{0.3\columnwidth}
			\begin{minipage}[t]{\columnwidth}
		      \centering
		      \includegraphics[width=\columnwidth]{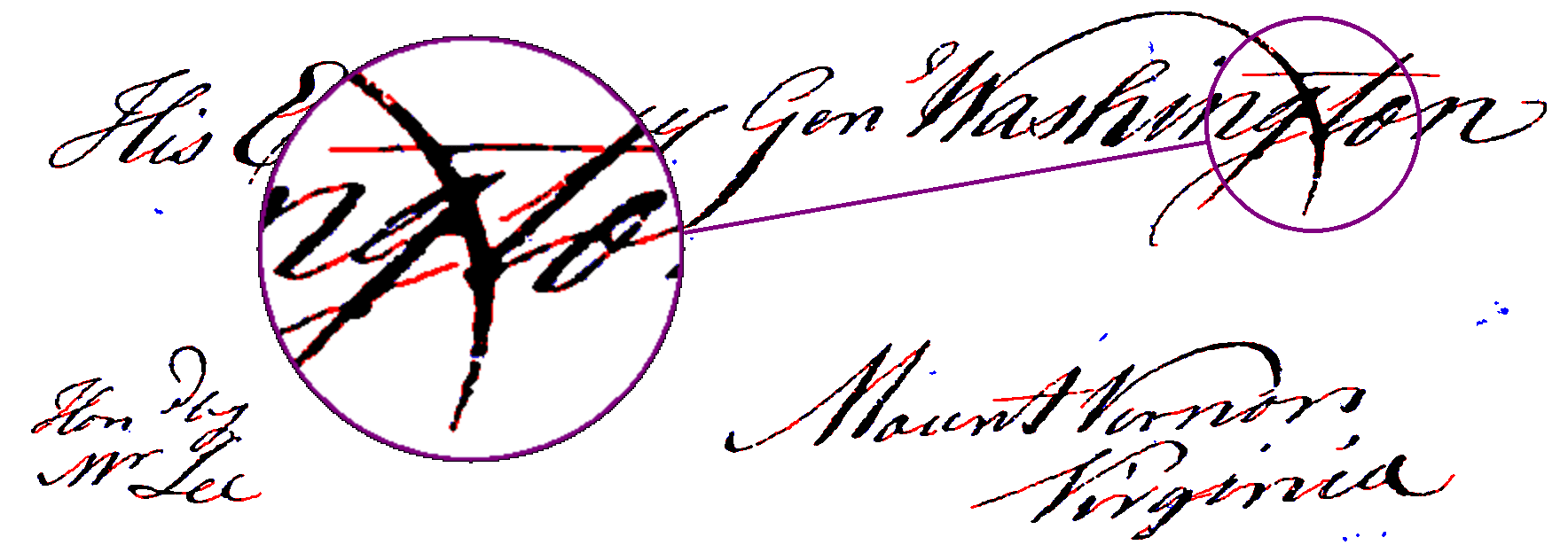}

              \includegraphics[width=\columnwidth]{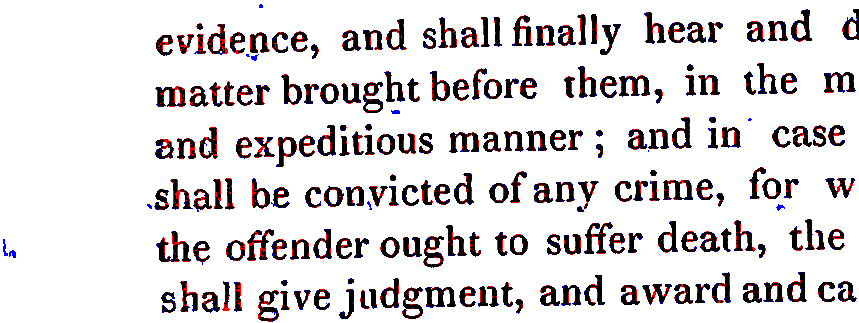}
              
              \includegraphics[width=\columnwidth]{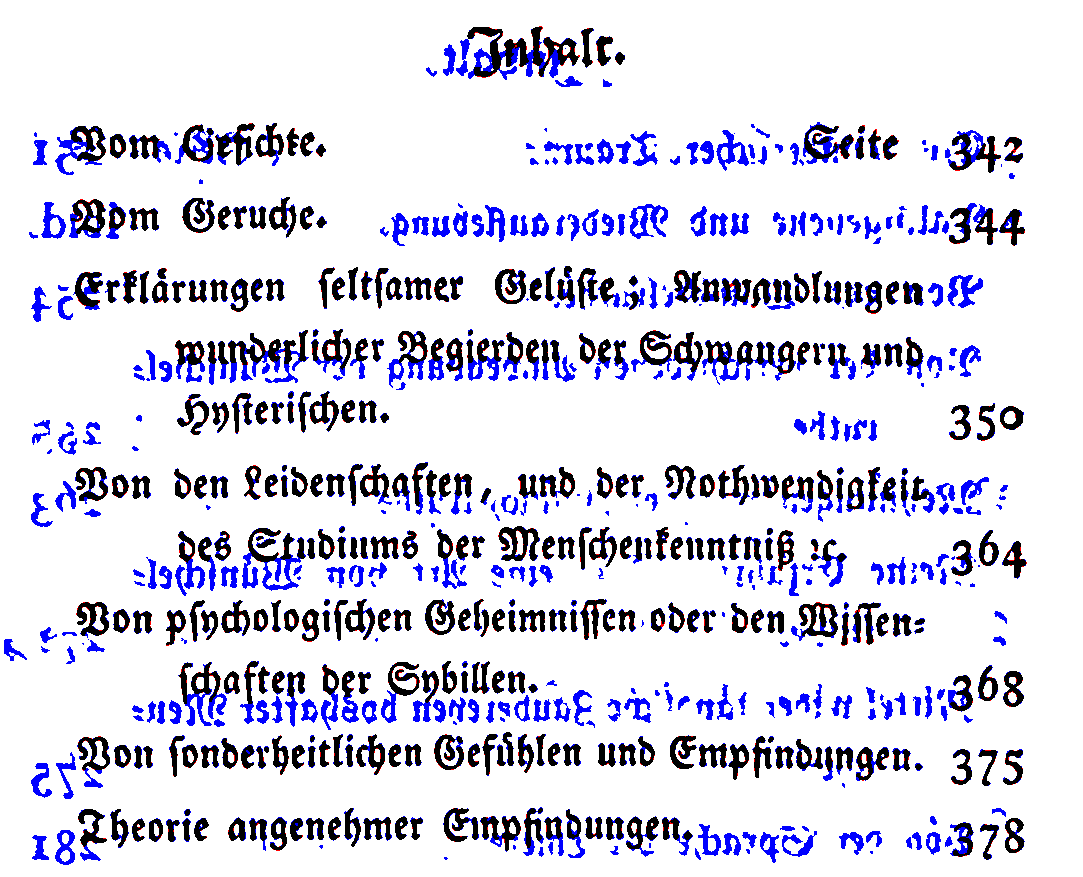}
            \end{minipage}
            \caption{Jia \cite{Jia2018}}
		\end{subfigure}
        \hskip 0.5\columnwidth
        \begin{subfigure}{0.3\columnwidth}
			\begin{minipage}[t]{\columnwidth}
		      \centering
		      \includegraphics[width=\columnwidth]{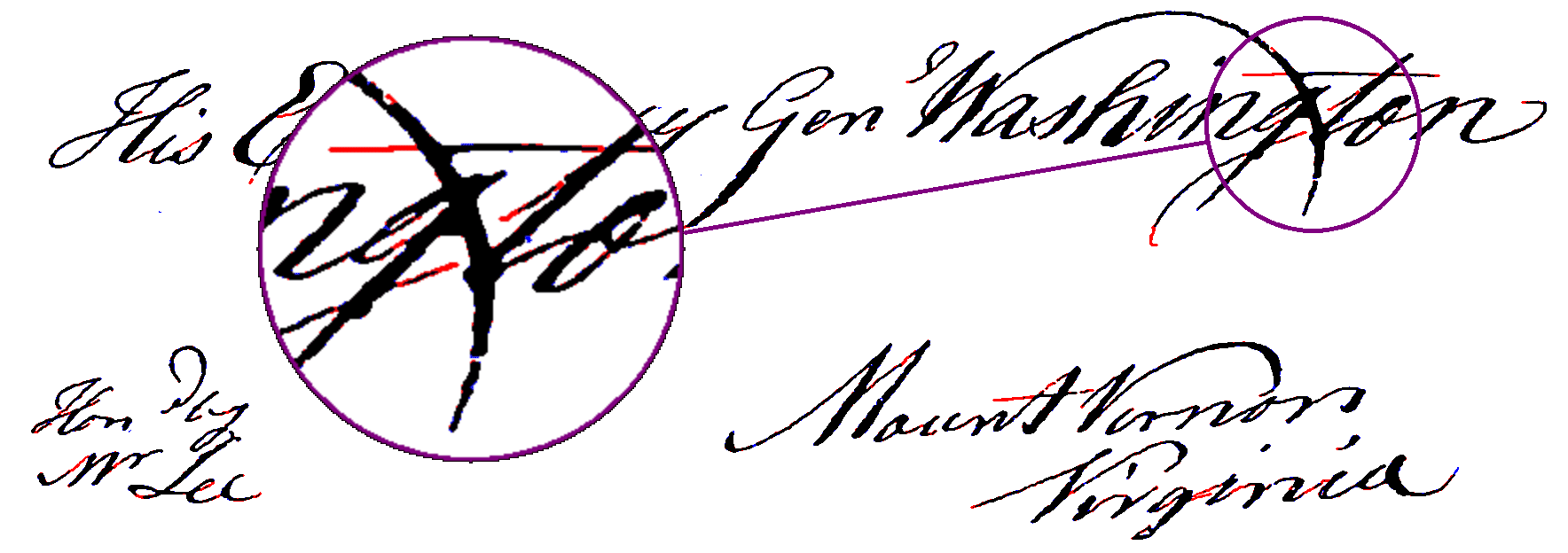}

              \includegraphics[width=\columnwidth]{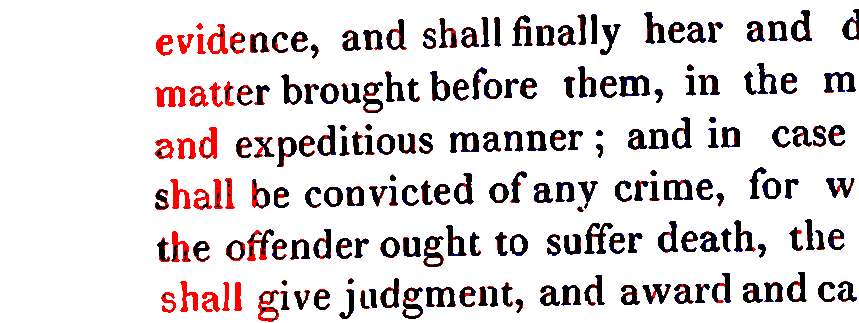}
              
              \includegraphics[width=\columnwidth]{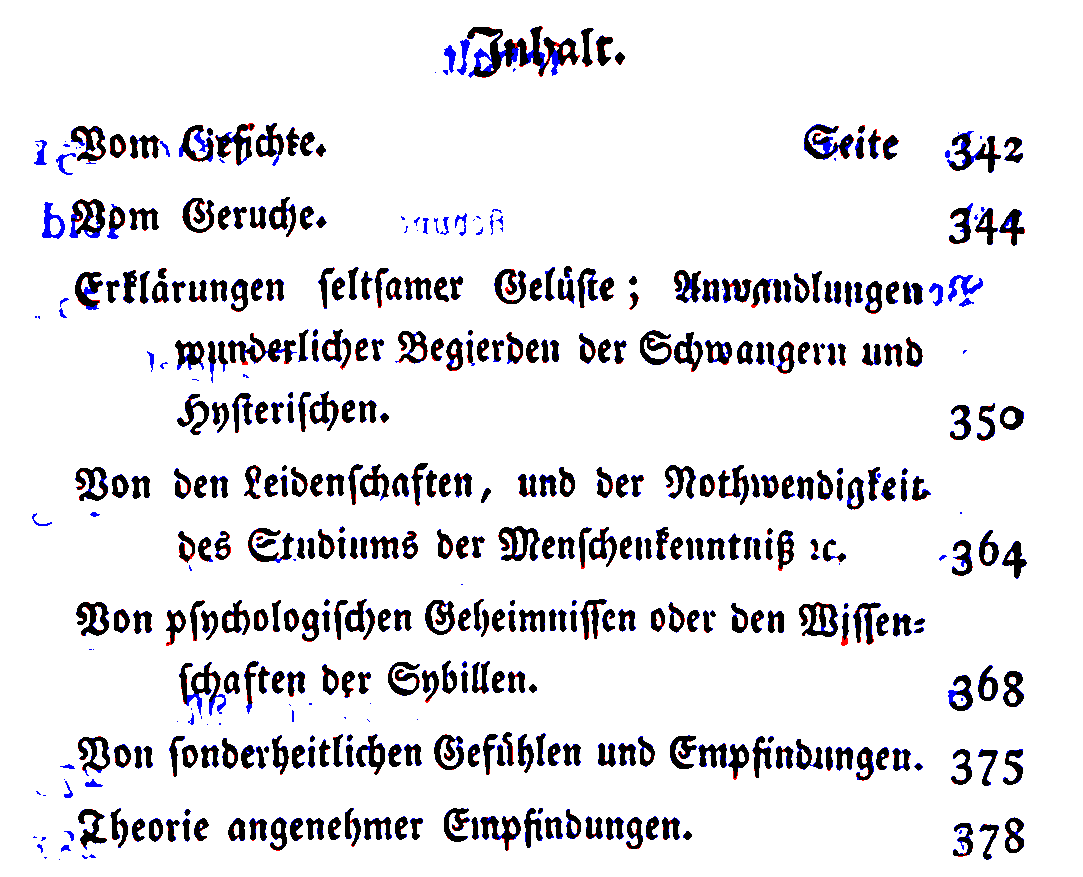}
            \end{minipage}
            \caption{Suh \cite{Suh2022}}
		\end{subfigure}
		\begin{subfigure}{0.3\columnwidth}
			\begin{minipage}[t]{\columnwidth}
		      \centering
		      \includegraphics[width=\columnwidth]{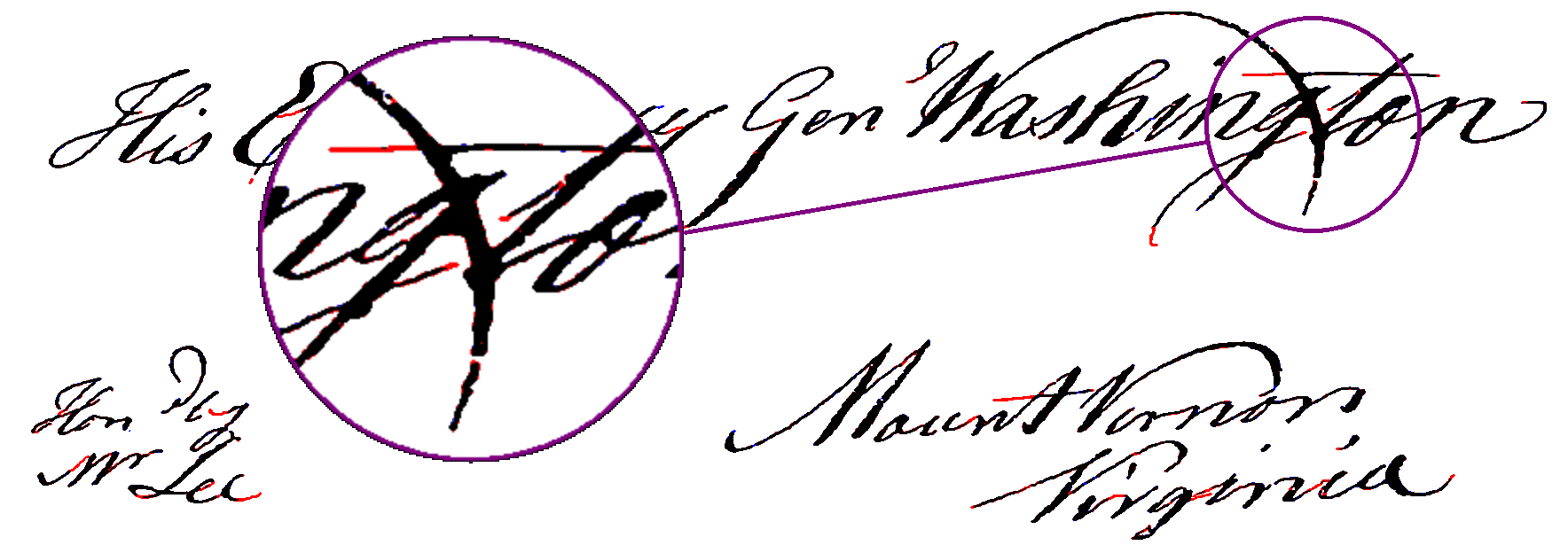}

              \includegraphics[width=\columnwidth]{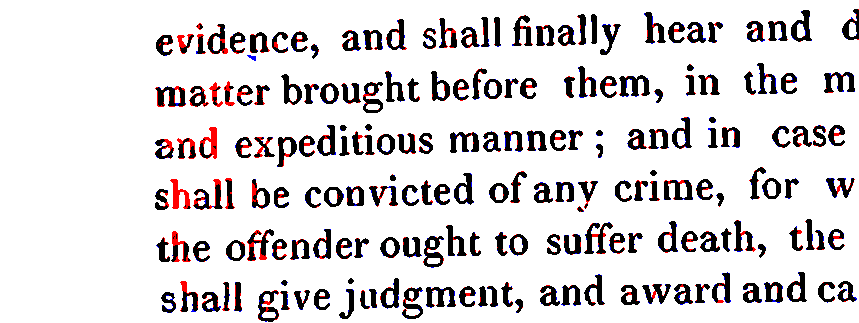}
              
              \includegraphics[width=\columnwidth]{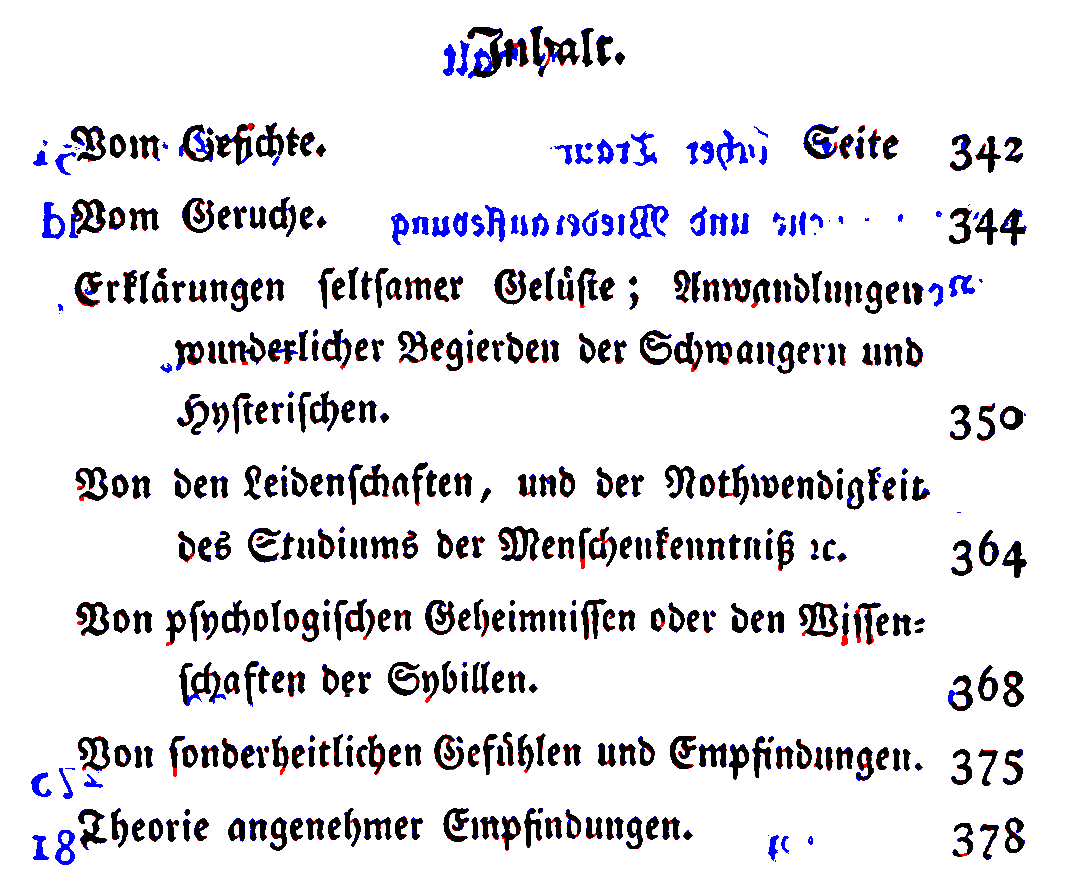}
            \end{minipage}
            \caption{Ours w/o MO}
		\end{subfigure}
		\begin{subfigure}{0.3\columnwidth}
			\begin{minipage}[t]{\columnwidth}
		      \centering
		      \includegraphics[width=\columnwidth]{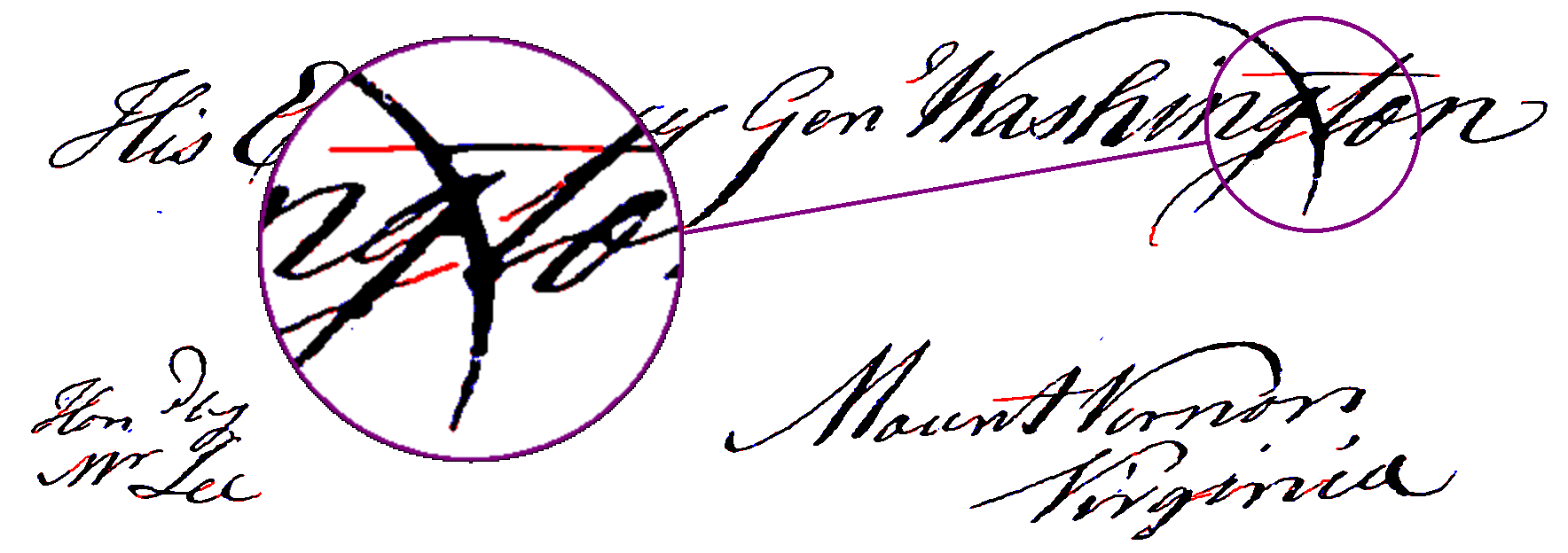}

              \includegraphics[width=\columnwidth]{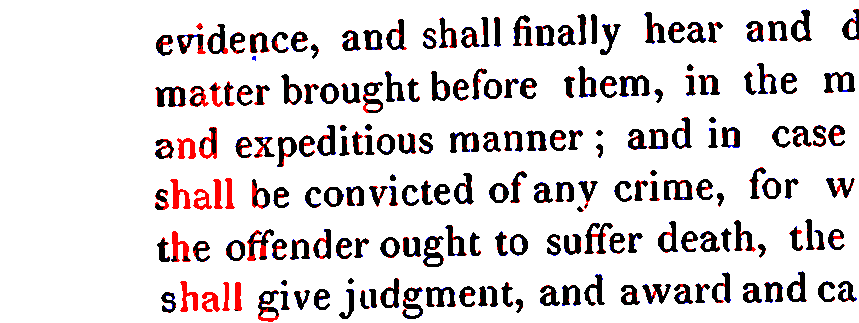}
              
              \includegraphics[width=\columnwidth]{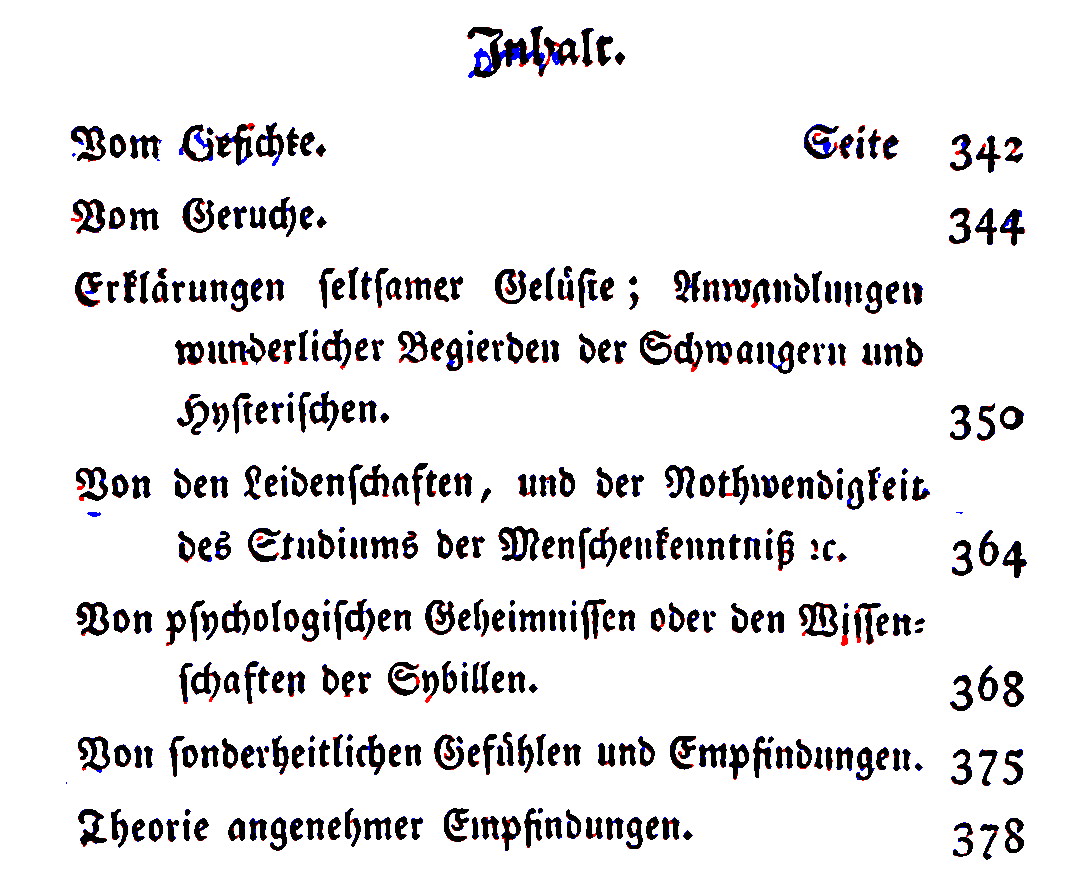}
            \end{minipage}
            \caption{Ours w MO}
		\end{subfigure}
		\caption{Binarization results of three representative document images. From top to bottom, the images are from H-DIBCO'10, DIBCO'11, and DIBCO'17. MO: Multi-scale operation.}
		\label{fig:DIBCO2010-2011}
\end{figure}

The proposed methods are evaluated on the ten (H-)DIBCO datasets and compared with traditional binarization methods including Otsu \cite{Otsu1979}, Sauvola \cite{Sauvola2000}, Howe \cite{Howe2013}, Jia \etal \cite{Jia2018}, reproducible deep-learning-based methods including cGANs \cite{Zhao2019}, and Suh \etal \cite{Suh2022} and best score per metric for any submission on each year, named best competition system. For a fair comparison, we use the publicly available source codes provided by the authors.

\Cref{table:sota} shows the quantitative evaluation results on the ten (H-)DIBCO datasets. On DIBCO'09, DIBCO'17, and the average of all ten datasets, GDB with MO achieves the best results than other state-of-the-art methods in terms of all four metrics. On H-DIBCO'10, DIBCO'11, DIBCO'13, and H-DIBCO'14, GDB without MO provides the best scores. On DIBCO'19, GDB without MO performs the best in terms of FM and p-FM but ranks second in PSNR and DRD. Each benchmark dataset focuses differently on different noises. The average results demonstrate the robustness of GDB against various noises. The coarse sub-network misclassifies some faint stroke edges as background based on the edge maps, which leads to relatively mediocre performance of GDB on H-DIBCO'16 and H-DIBCO'18.

\Cref{fig:DIBCO2010-2011} shows the binarization results of three representative document images containing three common degradations of uneven strokes, faint ink, and bleed-through. From top to bottom, the images are from H-DIBCO'10, DIBCO'11, and DIBCO'17. Due to the uneven intensity of text, the low-intensity text is easily recognized as background in a global field of view. Thus, GDB without MO can retain more text and extract strokes more finely than GDB with MO, especially at the end of the strokes. Howe \cite{Howe2013}, and Suh \etal \cite{Suh2022} also do not sufficiently keep the low-intensity text. Jia \cite{Jia2018} does an excellent job of preserving large areas of low-intensity text but needs to delineate the stroke edges sharply. Suh \etal \cite{Suh2022} can effectively remove the text-like background noise but still keeps a small amount of high-intensity bleed-through noise. GDB with MO shows the best performance in eliminating bleed-through noise.

\begin{figure}
		\centering
		\begin{subfigure}{0.3\columnwidth}
			\includegraphics[width=\columnwidth]{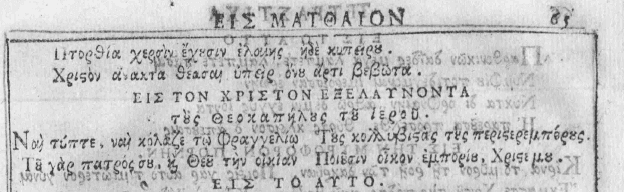}
			
		\includegraphics[width=\columnwidth]{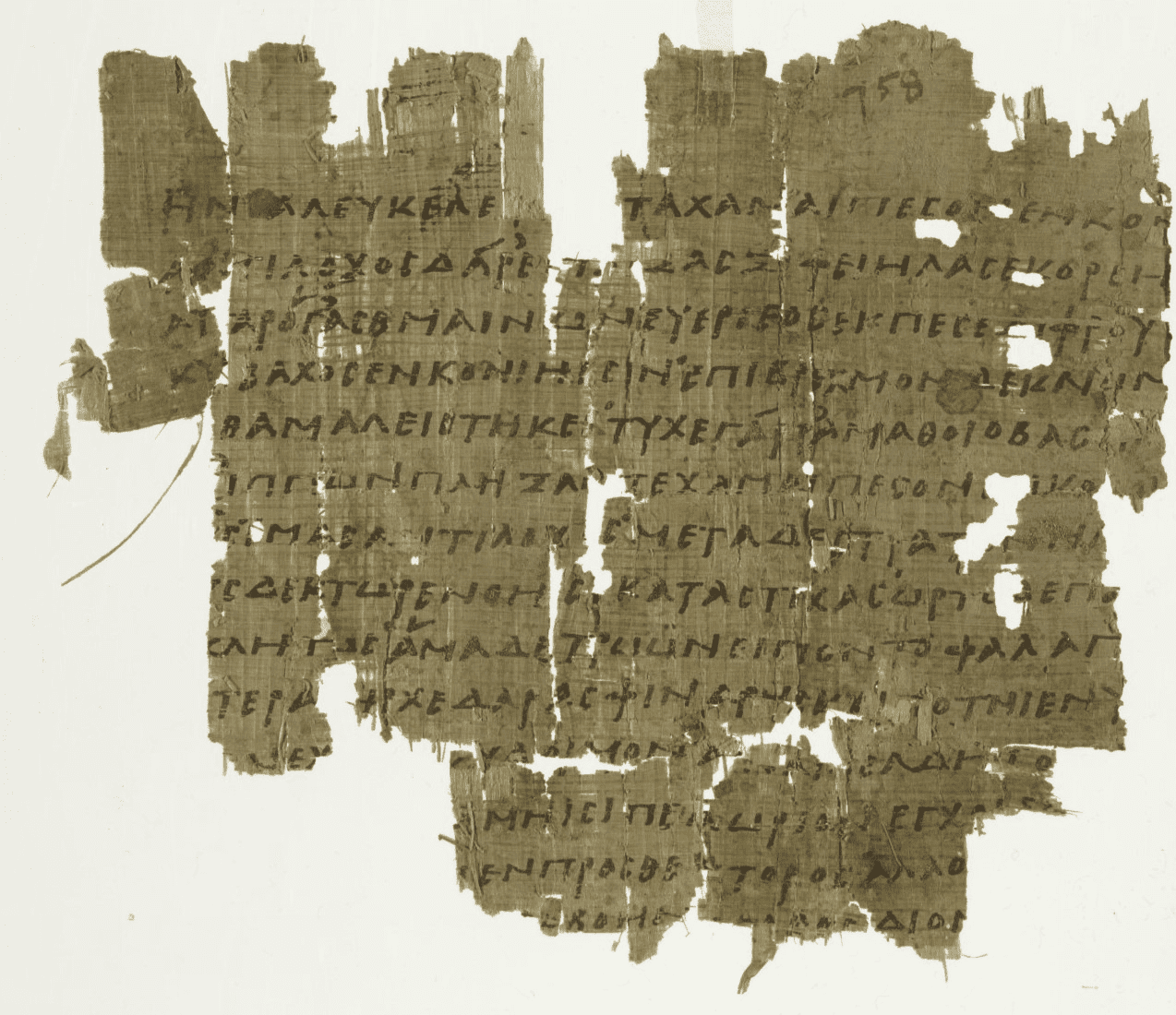}
			\caption{Original image}
		\end{subfigure}
		\begin{subfigure}{0.3\columnwidth}
			\includegraphics[width=\columnwidth]{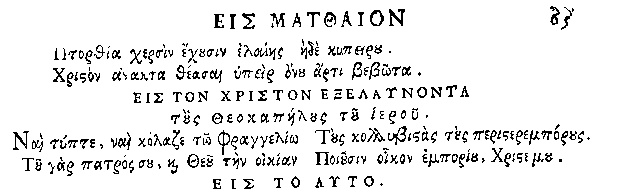}
			
    	\includegraphics[width=\columnwidth]{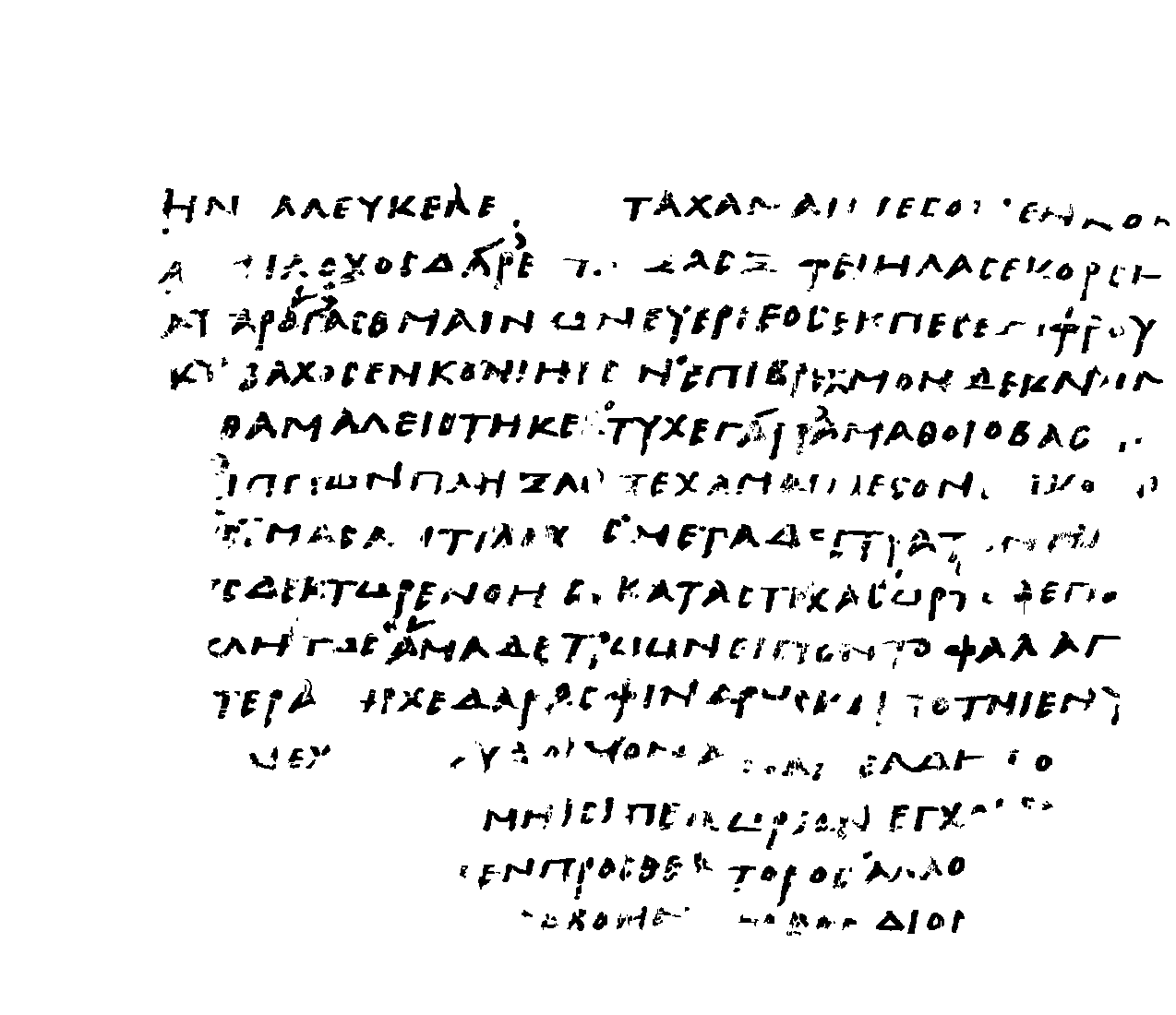}
			\caption{Ground-truth}
		\end{subfigure}
		\begin{subfigure}{0.3\columnwidth}
			\includegraphics[width=\columnwidth]{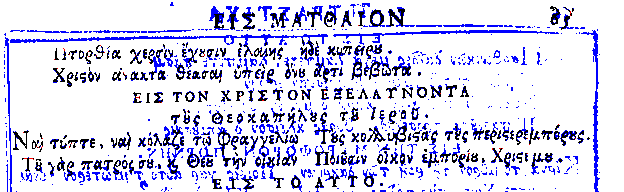}
			
		\includegraphics[width=\columnwidth]{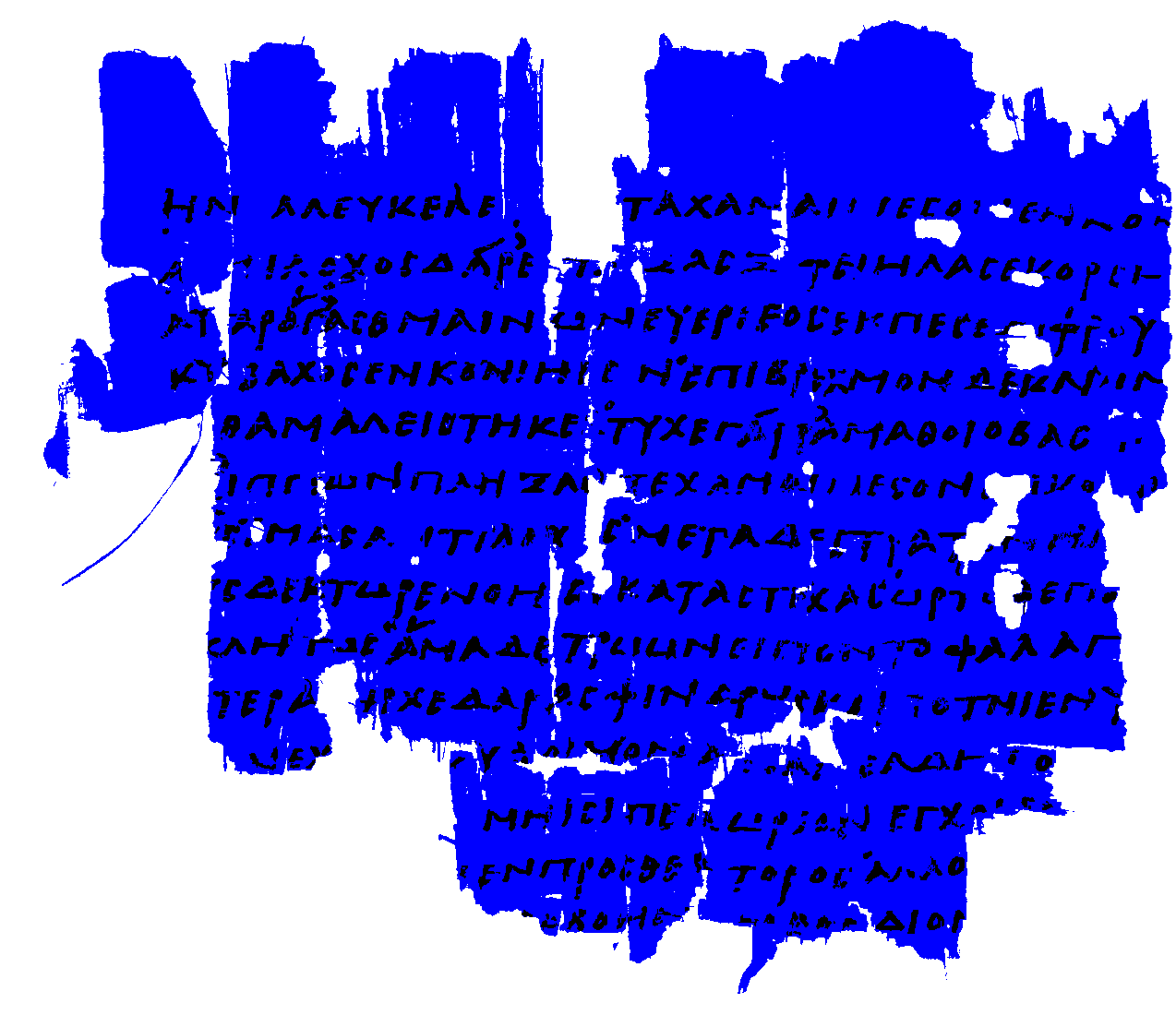}
			\caption{Otsu \cite{Otsu1979}}
		\end{subfigure}
		\hskip 0.5\columnwidth
        \begin{subfigure}{0.3\columnwidth}
			\includegraphics[width=\columnwidth]{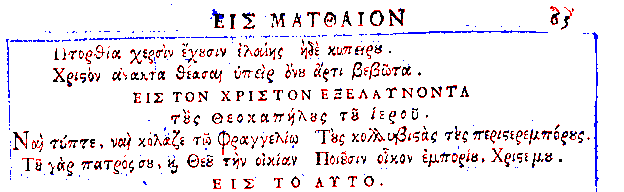}
			\includegraphics[width=\columnwidth]{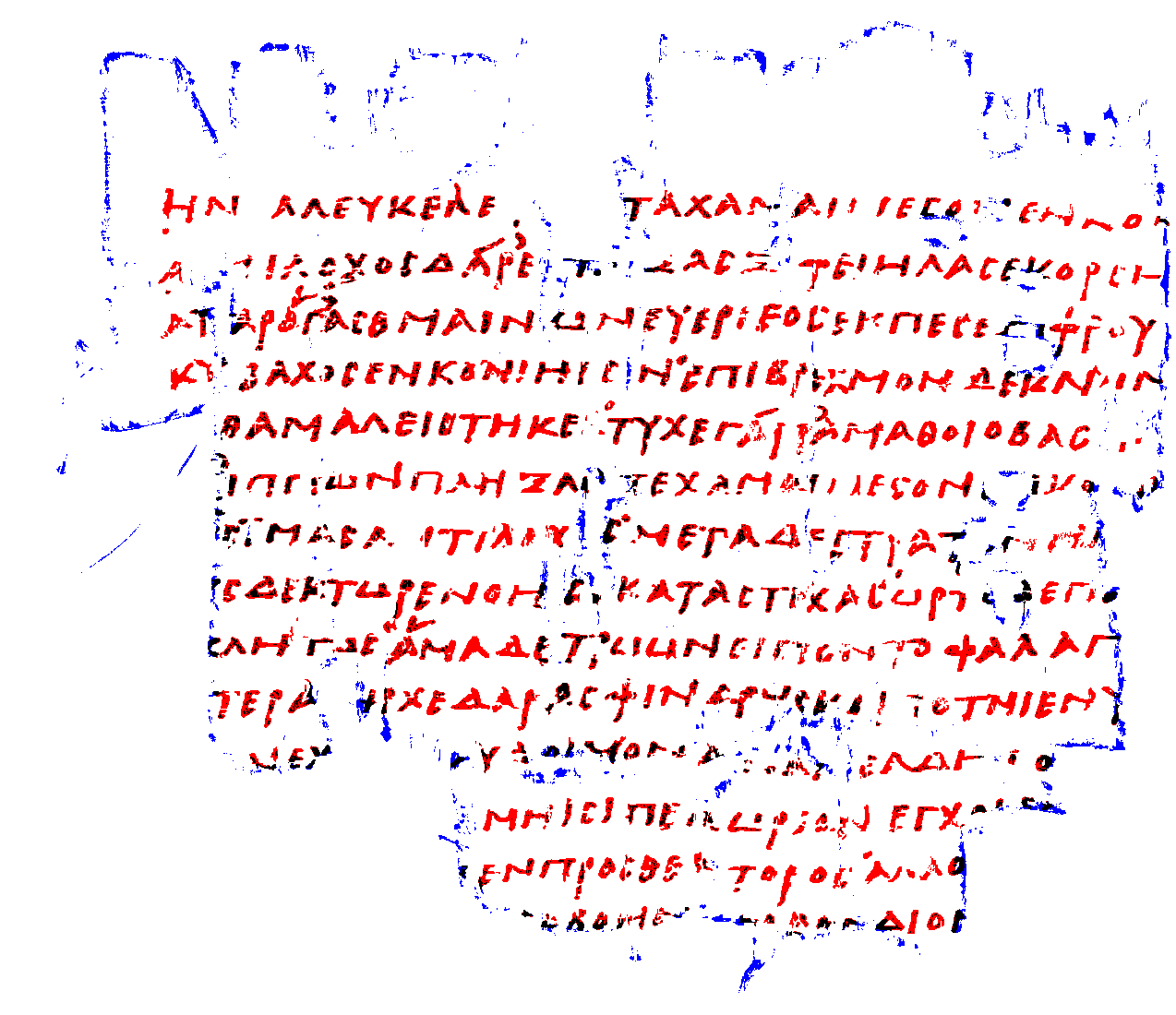}
			\caption{Sauvola \cite{Sauvola2000}}
		\end{subfigure}
        \begin{subfigure}{0.3\columnwidth}
			\includegraphics[width=\columnwidth]{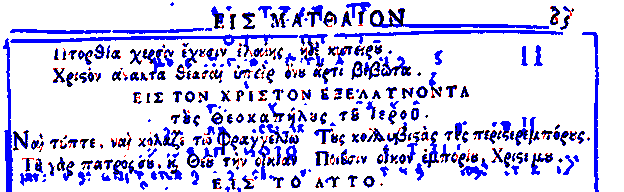}
			\includegraphics[width=\columnwidth]{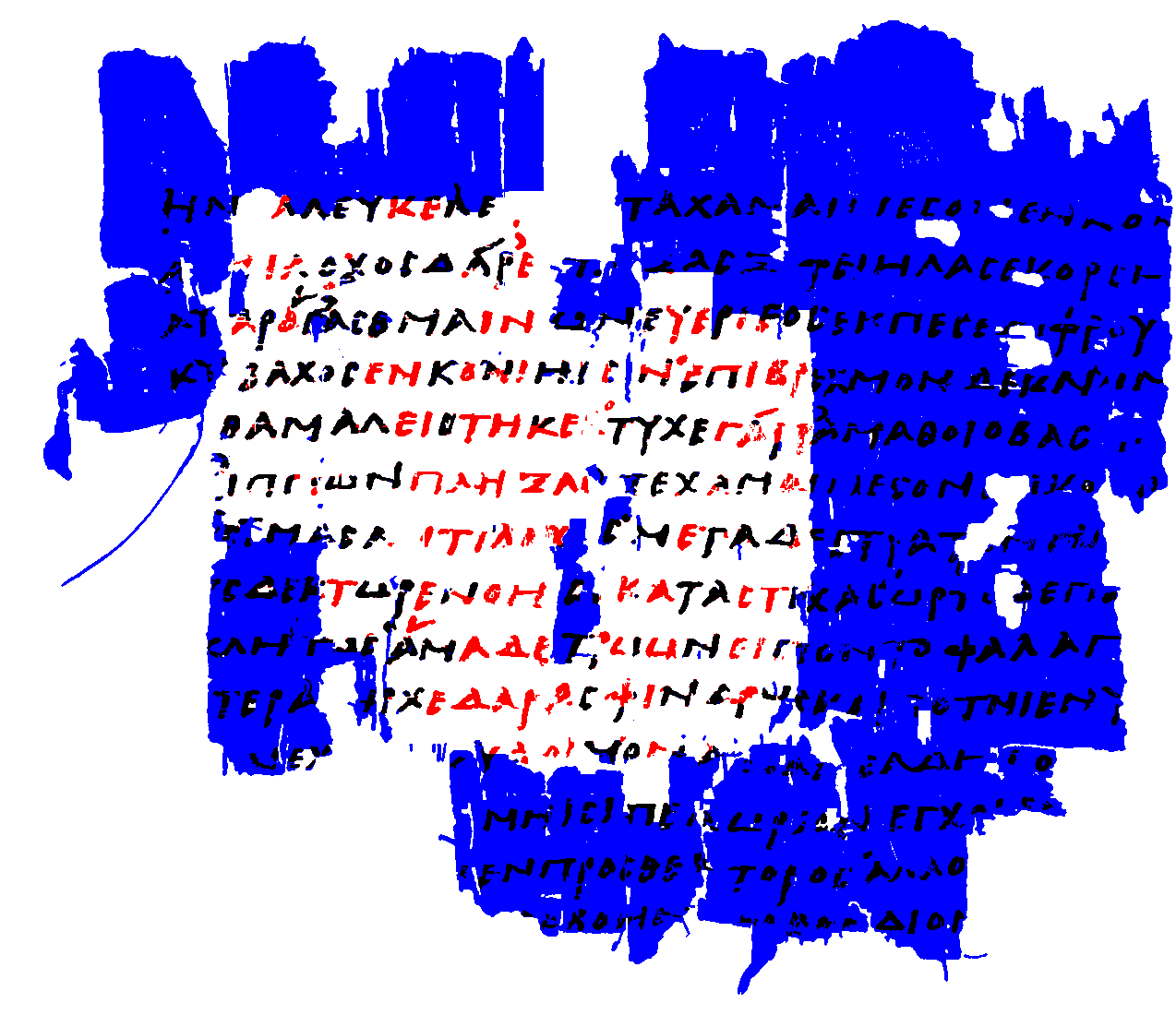}
			\caption{Howe \cite{Howe2013}}
		\end{subfigure}
		\begin{subfigure}{0.3\columnwidth}
            \label{fig:2019_9_f}
			\includegraphics[width=\columnwidth]{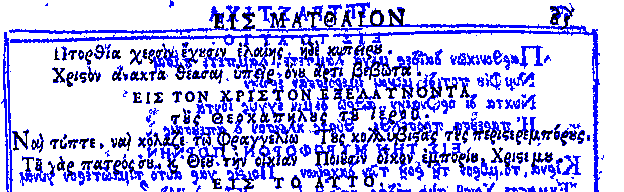}
			\includegraphics[width=\columnwidth]{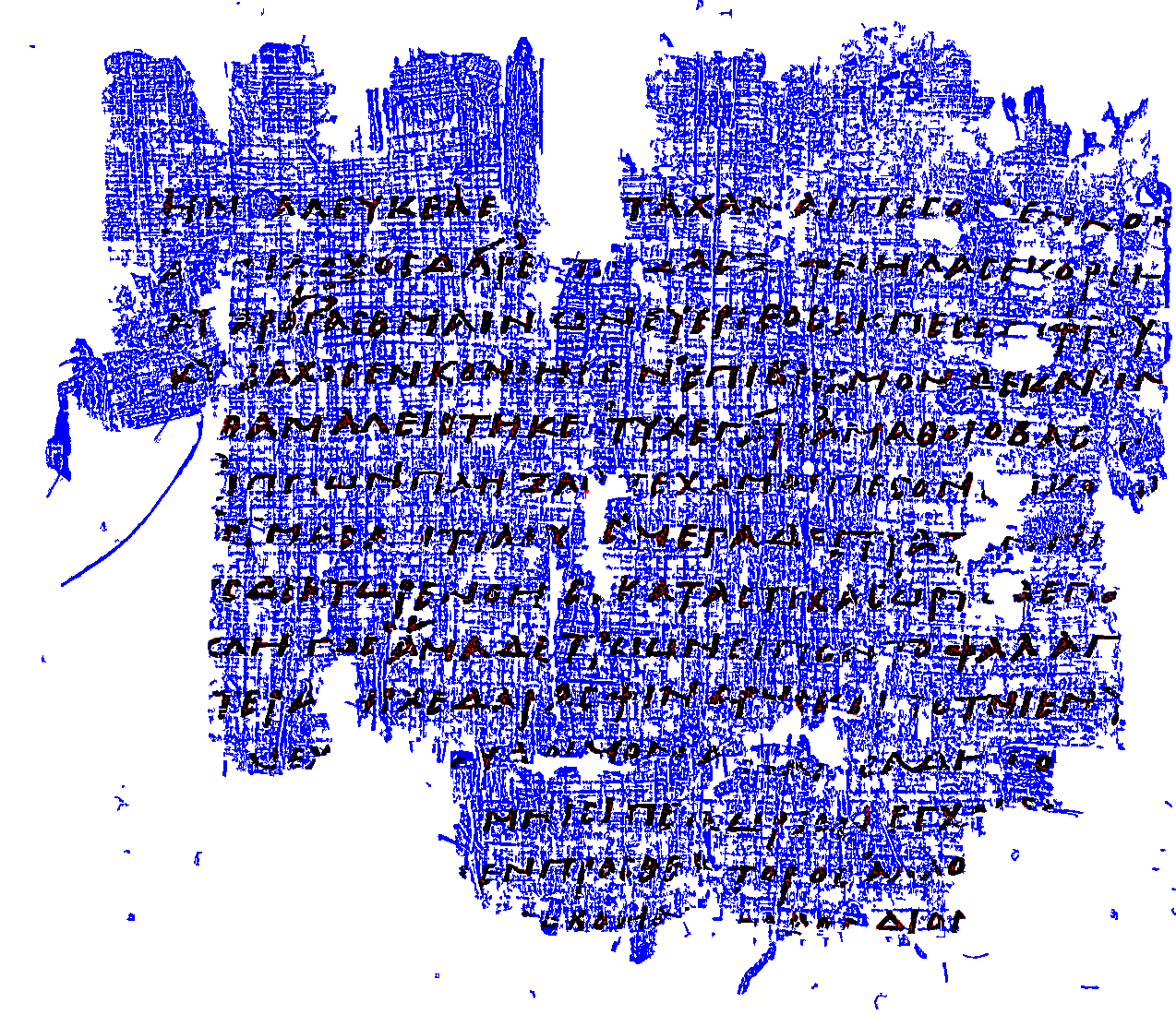}
			\caption{Jia \cite{Jia2018}}
		\end{subfigure}
        \hskip 0.5\columnwidth
        \begin{subfigure}{0.3\columnwidth}
			\includegraphics[width=\columnwidth]{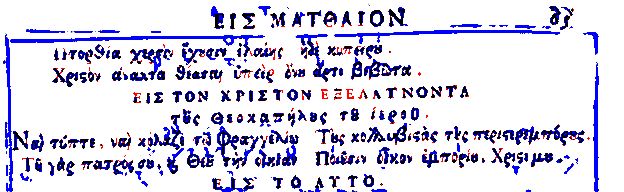}
			\includegraphics[width=\columnwidth]{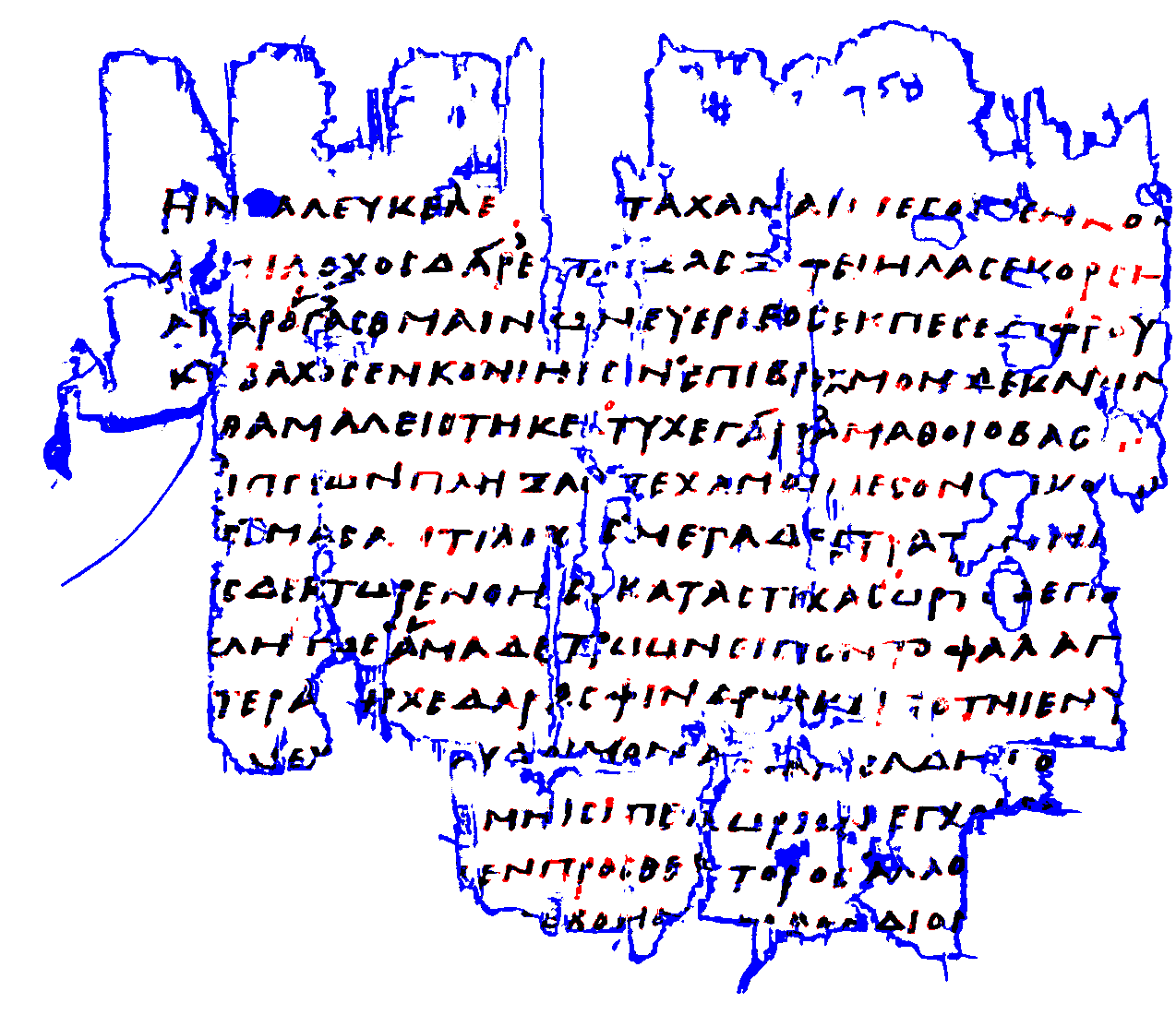}
			\caption{cGANs \cite{Zhao2019}}
		\end{subfigure}
        \begin{subfigure}{0.3\columnwidth}
			\includegraphics[width=\columnwidth]{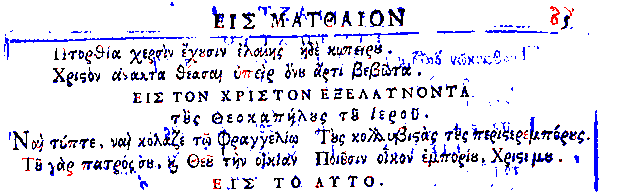}
			\includegraphics[width=\columnwidth]{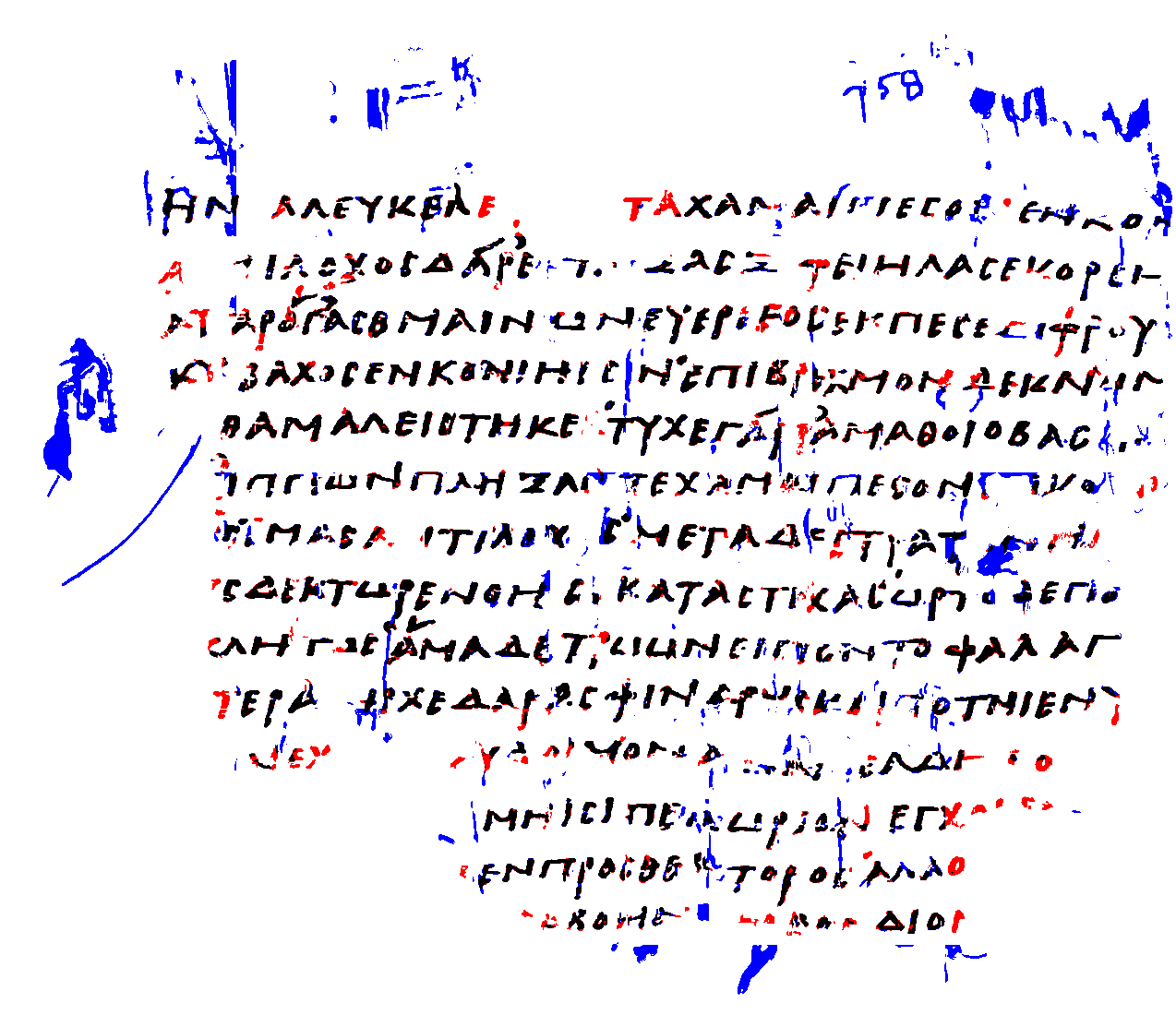}
			\caption{Ours w/o MO}
		\end{subfigure}
        \begin{subfigure}{0.3\columnwidth}
			\includegraphics[width=\columnwidth]{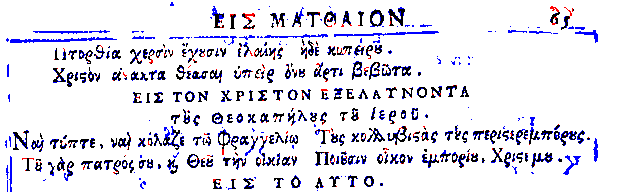}
			\includegraphics[width=\columnwidth]{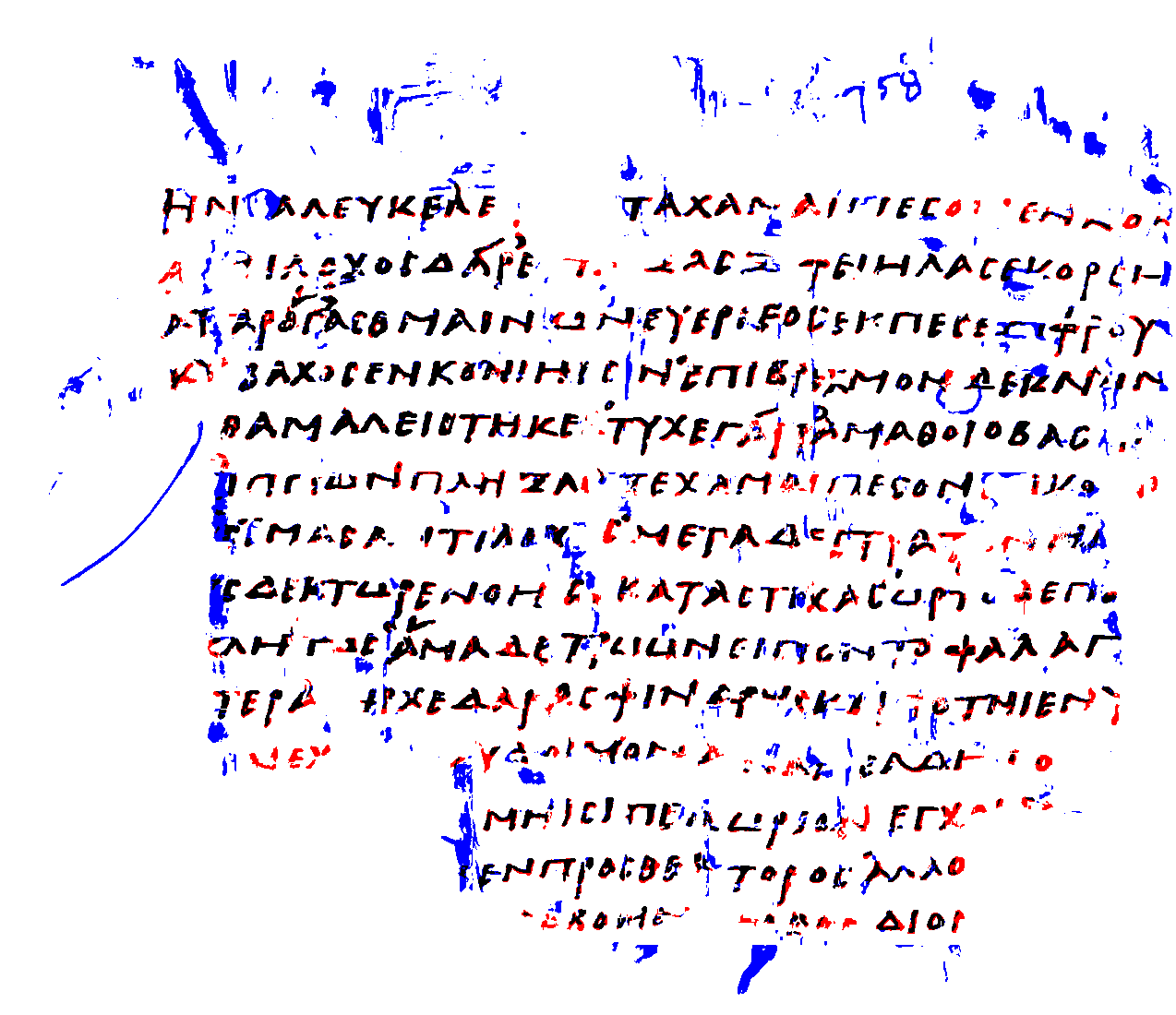}
			\caption{Ours w MO}
		\end{subfigure}
		\caption{Binarization results on DIBCO'19. MO: Multi-scale operation.}
		\label{fig:DIBCO2019}
\end{figure}

\Cref{fig:DIBCO2019} shows the binarization results on DIBCO’19. DIBCO'19 is a chanllenging dataset in which handwritten papyri images suffer from severe degradations, including broken holes, misplaced fibres, noisy margins, uneven distribution and text smudges. The traditional methods \cite{Otsu1979, Sauvola2000, Howe2013, Jia2018} do not effectively remove alien lines from the background and do not accurately extract the strokes from the document. The cGANs \cite{Zhao2019} removes the bleed-through noise better than the traditional methods but still suffers from the loss of text and the inclusion of alien lines. GDB provides relatively clean binarization results, where large areas of bleed-through noise in the background is effectively suppressed. For the papyri images, the performances of traditional methods are unacceptable. A large number of pixels are misclassified. Deep learning methods cannot binarize the papyri images well because the papyri images only appear in DIBCO'19. (the top two submissions in DIBCO'19 are traditional methods). Text is efficiently classified by cGANs \cite{Zhao2019}, but the torn margin of the papyri and the edge of the broken holes are also preserved. Compared to cGANs \cite{Zhao2019}, the GDB suppresses edge noise better.

\section{Conclusions}\label{sec:con}

In this paper, we propose a document binarization method termed GDB. Our proposed method focuses on the extraction of sharp stoke edges. By applying gated convolutions, We transform text extraction into the learning of gating values. We demonstrate that the learning process of gating values can be weakly guided by feeding a priori mask and edge map. We show that the learned gating values have different regions of interest, which can gate the propagation of stroke edge features and effectively suppress noisy features. We add an edge branch to supervise the extraction of stroke edges. We evaluated the GDB over ten (H-)DIBCO benchmark datasets. The experimental results show that the GDB achieves state-of-art performance benefitting from the different components of GDB.

{\small
\bibliographystyle{ieee_fullname}
\bibliography{egbib}
}

\end{document}